%% file: _main.tex
\input{_constants}
\arxiv

\pdfoutput=1
\documentclass[10pt,twocolumn,letterpaper]{article}
\input{cvpr_header}
\begin{document}
\title{\paperTitle}
\author{\authorBlock}
\maketitle

\input{contents/00_abstract}

\input{contents/01_intro}
\input{contents/02_related}
\input{contents/03_method}
\input{contents/04_experiments}
\input{contents/10_conclusion}

{\small
\bibliographystyle{ieeenat_fullname}
\bibliography{_references}
}

\ifarxiv \clearpage \appendix \input{contents/12_appendix} \fi

\end{document}

%% file: _constants.tex
\def\paperID{12147}
\def\confName{CVPR}
\def\confYear{2024}

\def\paperTitle{
HarmonyView: Harmonizing Consistency and Diversity in One-Image-to-3D
}

\def\authorBlock{
    Sangmin Woo\textsuperscript{\rm 1}\thanks{Equal contribution} \qquad
    Byeongjun Park\textsuperscript{\rm 1}\footnotemark[1] \qquad
    Hyojun Go\textsuperscript{\rm 2} \qquad
    Jin-Young Kim\textsuperscript{\rm 2} \qquad
    Changick Kim\textsuperscript{\rm 1} \vspace{0.1cm} \\
    \textsuperscript{\rm 1}KAIST \qquad
    \textsuperscript{\rm 2}Twelve Labs \\
    {\tt\small \textsuperscript{\rm 1}\{smwoo95, pbj3810, changick\}@kaist.ac.kr} \qquad
    {\tt\small \textsuperscript{\rm 2}\{william, jeremy\}@twelvelabs.io} \vspace{0.1cm} \\
    {\normalsize {\href{https://byeongjun-park.github.io/HarmonyView/}{https://byeongjun-park.github.io/HarmonyView/}} }
}

\newif\ifreview 
\newif\ifarxiv \newcommand{\arxiv}{\arxivtrue}
\newif\ifcamera 
\newif\ifrebuttal 

%% file: cvpr_header.tex
\ifreview \usepackage[review]{cvpr} \fi
\ifarxiv \usepackage[pagenumbers]{cvpr} \fi
\ifrebuttal \usepackage[rebuttal]{cvpr} \fi
\ifcamera \usepackage{cvpr} \fi

\input{_macros}  

\usepackage{xr-hyper}

\makeatletter
\newcommand*{\addFileDependency}[1]{
  \typeout{(#1)}
  \@addtofilelist{#1}
  \IfFileExists{#1}{}{\typeout{No file #1.}}
}

\makeatother

\definecolor{cvprblue}{rgb}{0.21,0.49,0.74}
\usepackage[pagebackref,breaklinks,colorlinks,citecolor=cvprblue]{hyperref}
\usepackage[capitalize]{cleveref}
\crefname{section}{Sec.}{Secs.}
\crefname{table}{Table}{Tables}
\crefname{figure}{Fig.}{Figs.}

%% file: _macros.tex
\usepackage{graphicx}	
\usepackage{amsmath}	
\usepackage{amssymb}	
\usepackage{booktabs}
\usepackage{cuted}
\usepackage{times}
\usepackage{microtype}
\usepackage{epsfig}
\usepackage[table,xcdraw,dvipsnames]{xcolor}
\usepackage{caption}
\usepackage{float}
\usepackage{placeins}
\usepackage{color, colortbl}
\usepackage{stfloats}
\usepackage{enumitem}
\usepackage{tabularx}
\usepackage{xstring}
\usepackage{multirow}
\usepackage{xspace}
\usepackage{url}
\usepackage{subcaption}
\usepackage{xcolor}
\usepackage[hang,flushmargin]{footmisc}
\usepackage{kotex}
\usepackage{arydshln}
\ifcamera \usepackage[accsupp]{axessibility} \fi

% ...

\ifarxiv  \fi

\newcommand{\R}[1]{{%
    \textbf{%
        \ifstrequal{#1}{1}{\textcolor{red}{R#1}}{%
        \ifstrequal{#1}{2}{\textcolor{blue}{R#1}}{%
        \ifstrequal{#1}{3}{\textcolor{magenta}{R#1}}{%
        \ifstrequal{#1}{4}{\textcolor{teal}{R#1}}{%
                           \textcolor{cyan}{R#1}%
        }}}}%
    }%
}}

\input{_math_commands}

\definecolor{Green}{rgb}{0.2, 0.7, 0.1}

\newlength\abovesecmargin
\newlength\belowsecmargin
\newlength\abovesubsecmargin
\newlength\belowsubsecmargin
\newlength\abovesubsubsecmargin
\newlength\belowsubsubsecmargin
\newlength\paramargin
\newlength\abovetabcapmargin
\newlength\belowtabcapmargin
\newlength\abovefigcapmargin
\newlength\belowfigcapmargin

%% file: _math_commands.tex
\usepackage{amsmath,amsfonts,bm}
\usepackage{mathtools}

\def\eqref#1{equation~\ref{#1}}

\def\1{\bm{1}}

\def\rvc{{\mathbf{c}}}

\def\rvr{{\mathbf{r}}}

\def\rvv{{\mathbf{v}}}

\def\rvx{{\mathbf{x}}}
\def\rvy{{\mathbf{y}}}

\def\vl{{\bm{l}}}

\DeclareMathAlphabet{\mathsfit}{\encodingdefault}{\sfdefault}{m}{sl}
\SetMathAlphabet{\mathsfit}{bold}{\encodingdefault}{\sfdefault}{bx}{n}

\newcommand{\E}{\mathbb{E}}



%% file: contents/00_abstract.tex
\input{figs/teaser}

\begin{abstract}
Recent progress in single-image 3D generation highlights the importance of multi-view coherency, leveraging 3D priors from large-scale diffusion models pretrained on Internet-scale images.
However, the aspect of novel-view diversity remains underexplored within the research landscape due to the ambiguity in converting a 2D image into 3D content, where numerous potential shapes can emerge.
Here, we aim to address this research gap by simultaneously addressing both consistency and diversity.
Yet, striking a balance between these two aspects poses a considerable challenge due to their inherent trade-offs.
This work introduces \textbf{HarmonyView}, a simple yet effective diffusion sampling technique adept at decomposing two intricate aspects in single-image 3D generation: consistency and diversity.
This approach paves the way for a more nuanced exploration of the two critical dimensions within the sampling process.
Moreover, we propose a new evaluation metric based on CLIP image and text encoders to comprehensively assess the diversity of the generated views, which closely aligns with human evaluators' judgments.
In experiments, HarmonyView achieves a harmonious balance, demonstrating a win-win scenario in both consistency and diversity.
\end{abstract}

%% file: figs/teaser.tex
\begin{strip}
    \centering
    \vspace{-5.3em}
    \setlength\tabcolsep{1pt}
    \resizebox{\linewidth}{!}{
    \renewcommand{\arraystretch}{0.5}
    \begin{tabular}{c:cccccccc:cc}
    \includegraphics[width=0.086\textwidth]{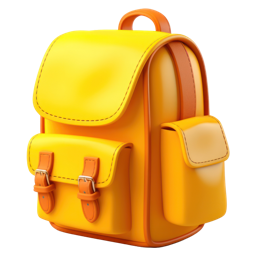} &
    \includegraphics[width=0.086\textwidth]{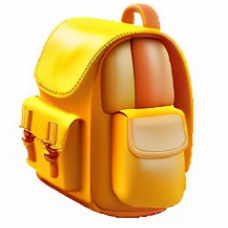} &
    \includegraphics[width=0.086\textwidth]{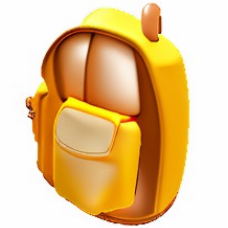} &
    \includegraphics[width=0.086\textwidth]{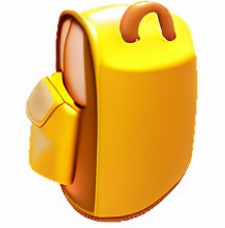} &
    \includegraphics[width=0.086\textwidth]{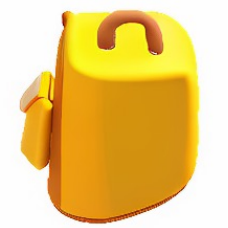} &
    \includegraphics[width=0.086\textwidth]{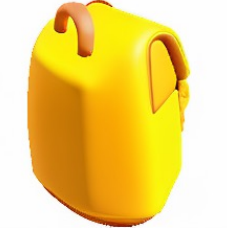} &
    \includegraphics[width=0.086\textwidth]{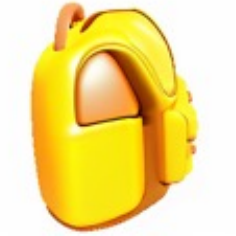} &
    \includegraphics[width=0.086\textwidth]{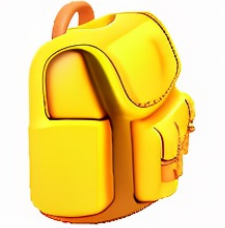} &
    \includegraphics[width=0.086\textwidth]{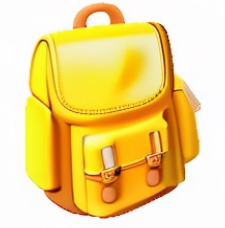} &
    \includegraphics[width=0.086\textwidth]{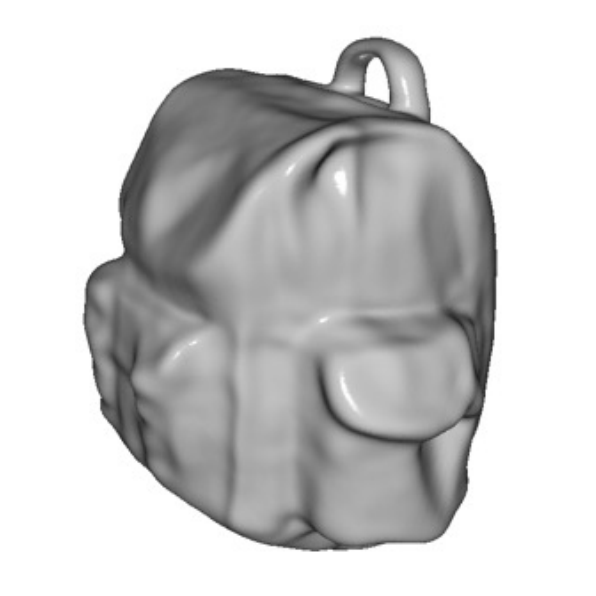} &
    \includegraphics[width=0.086\textwidth]{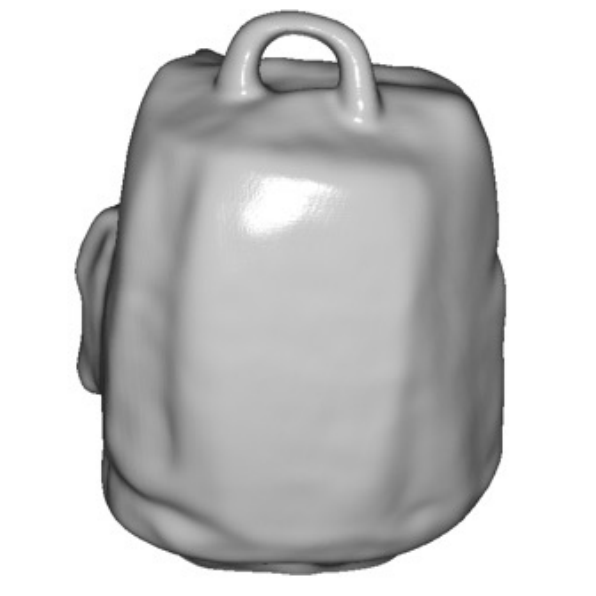} \\
     &
    \includegraphics[width=0.086\textwidth]{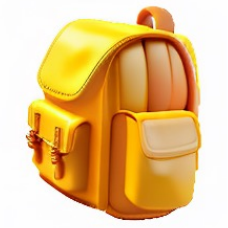} &
    \includegraphics[width=0.086\textwidth]{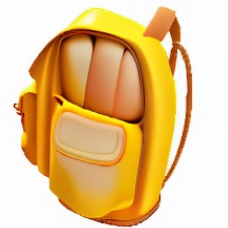} &
    \includegraphics[width=0.086\textwidth]{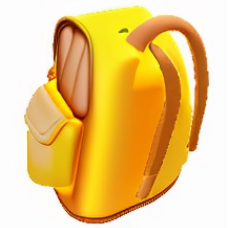} &
    \includegraphics[width=0.086\textwidth]{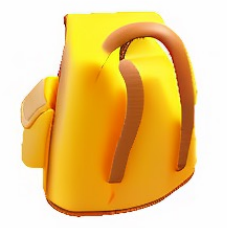} &
    \includegraphics[width=0.086\textwidth]{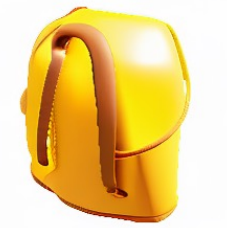} &
    \includegraphics[width=0.086\textwidth]{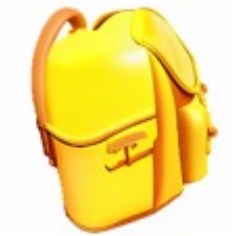} &
    \includegraphics[width=0.086\textwidth]{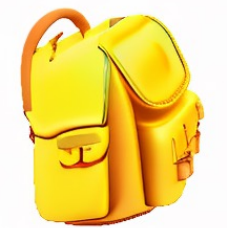} &
    \includegraphics[width=0.086\textwidth]{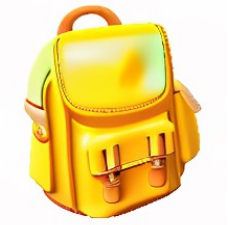} &
    \includegraphics[width=0.086\textwidth]{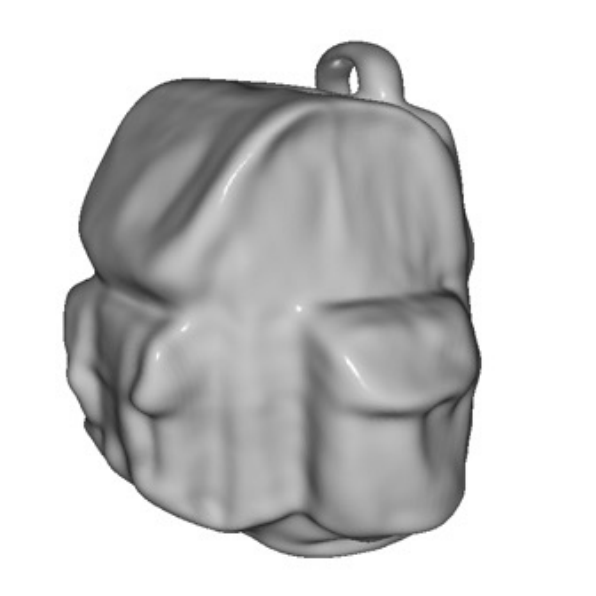} &
    \includegraphics[width=0.086\textwidth]{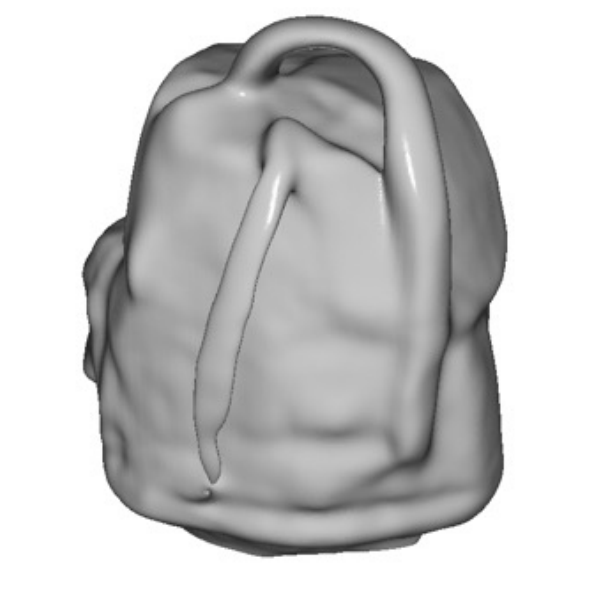} \\
    \includegraphics[width=0.086\textwidth]{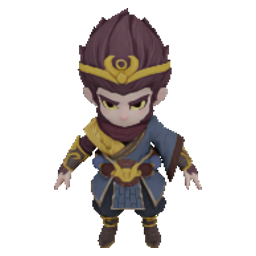} &
    \includegraphics[width=0.086\textwidth]{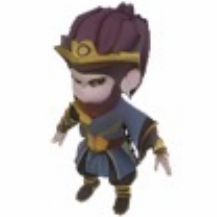} &
    \includegraphics[width=0.086\textwidth]{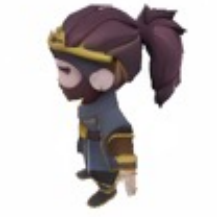} &
    \includegraphics[width=0.086\textwidth]{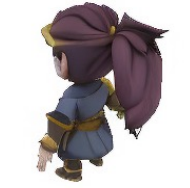} &
    \includegraphics[width=0.086\textwidth]{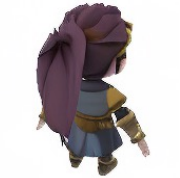} &
    \includegraphics[width=0.086\textwidth]{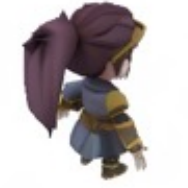} &
    \includegraphics[width=0.086\textwidth]{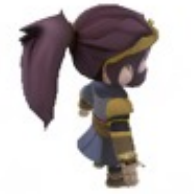} &
    \includegraphics[width=0.086\textwidth]{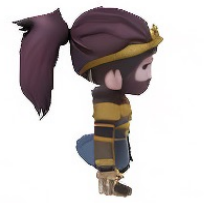} &
    \includegraphics[width=0.086\textwidth]{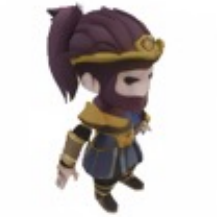} &
    \includegraphics[width=0.086\textwidth]{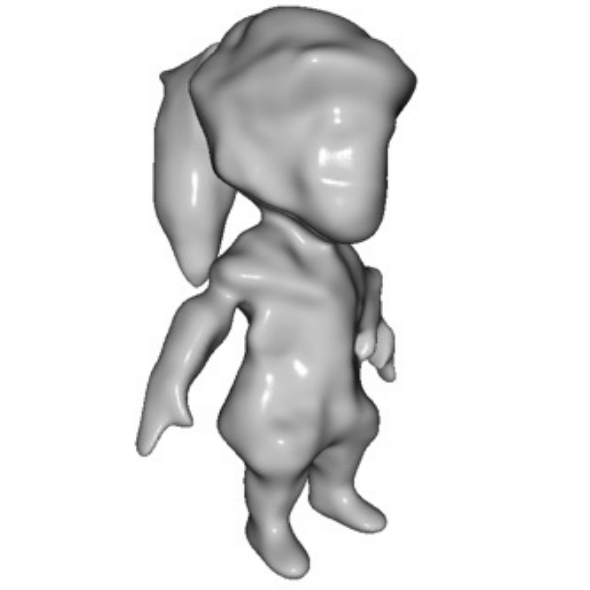} &
    \includegraphics[width=0.086\textwidth]{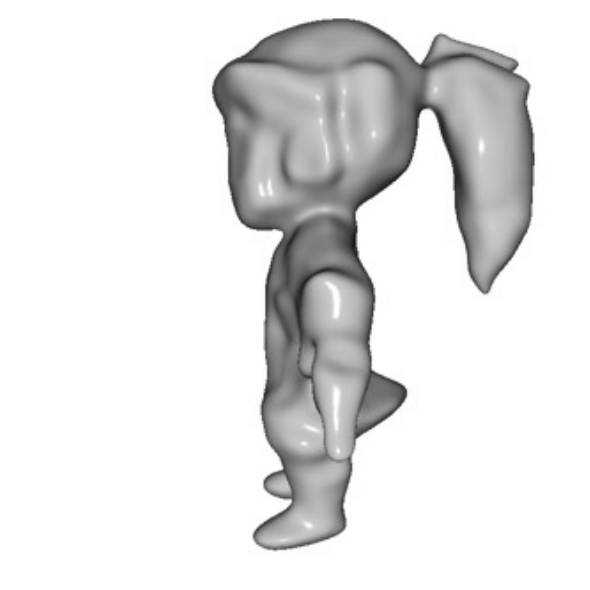} \\
     &
    \includegraphics[width=0.086\textwidth]{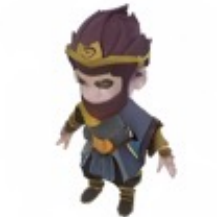} &
    \includegraphics[width=0.086\textwidth]{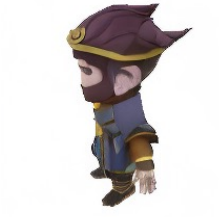} &
    \includegraphics[width=0.086\textwidth]{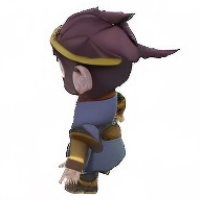} &
    \includegraphics[width=0.086\textwidth]{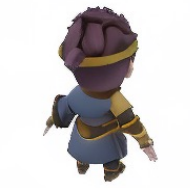} &
    \includegraphics[width=0.086\textwidth]{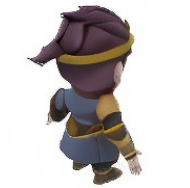} &
    \includegraphics[width=0.086\textwidth]{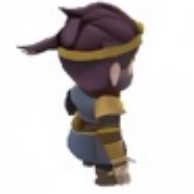} &
    \includegraphics[width=0.086\textwidth]{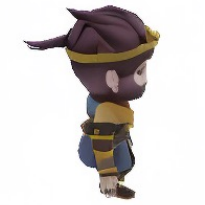} &
    \includegraphics[width=0.086\textwidth]{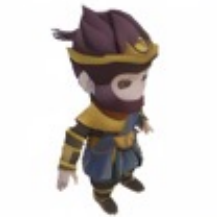} &
    \includegraphics[width=0.086\textwidth]{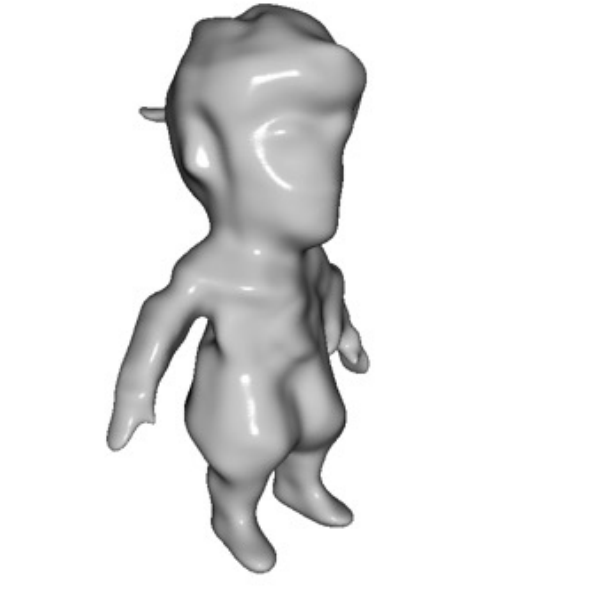} &
    \includegraphics[width=0.086\textwidth]{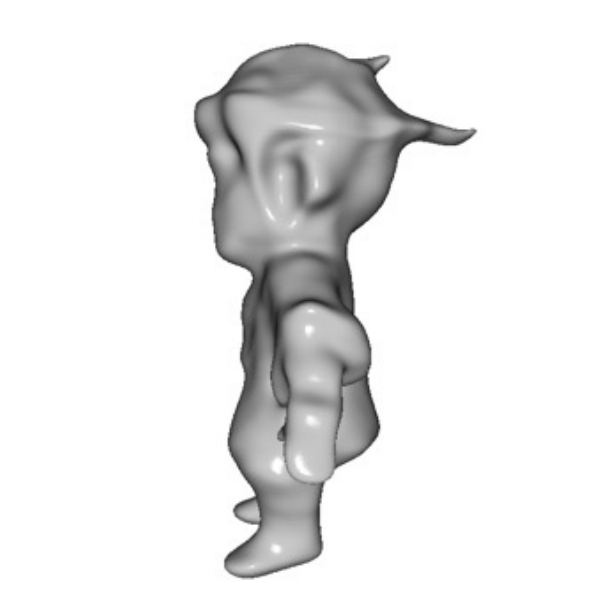} \\
    \includegraphics[width=0.086\textwidth]{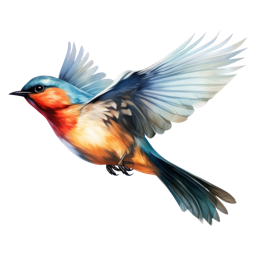} &
    \includegraphics[width=0.086\textwidth]{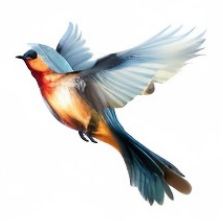} &
    \includegraphics[width=0.086\textwidth]{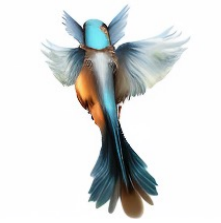} &
    \includegraphics[width=0.086\textwidth]{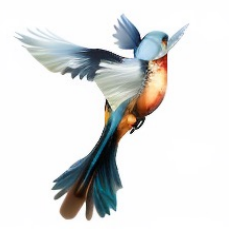} &
    \includegraphics[width=0.086\textwidth]{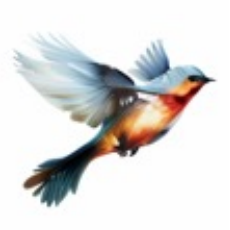} &
    \includegraphics[width=0.086\textwidth]{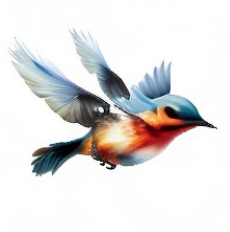} &
    \includegraphics[width=0.086\textwidth]{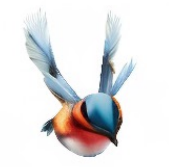} &
    \includegraphics[width=0.086\textwidth]{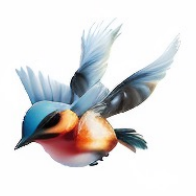} &
    \includegraphics[width=0.086\textwidth]{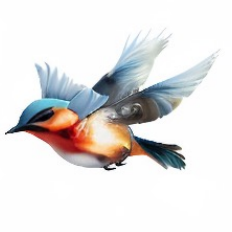} &
    \includegraphics[width=0.086\textwidth]{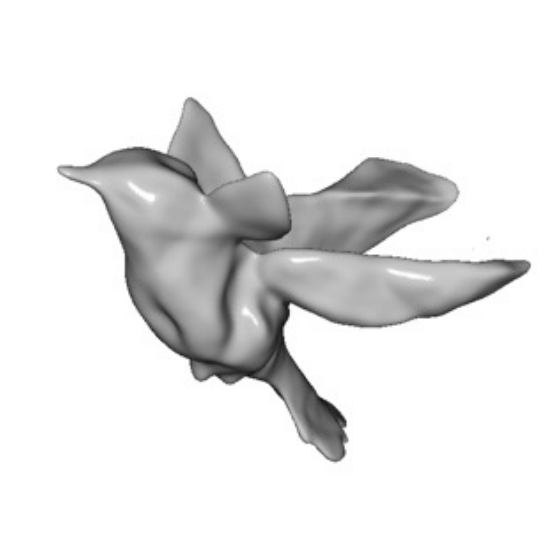} &
    \includegraphics[width=0.086\textwidth]{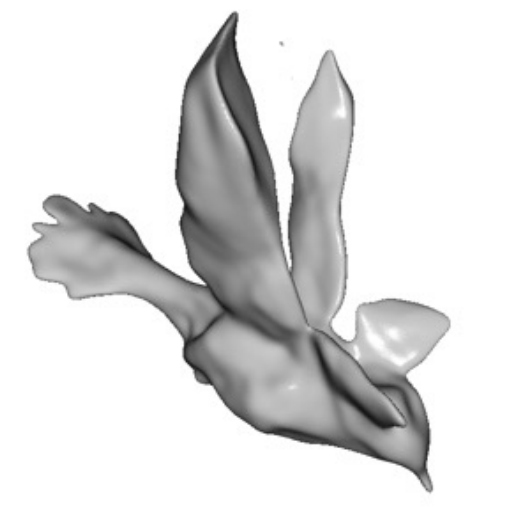} \\
     &
    \includegraphics[width=0.086\textwidth]{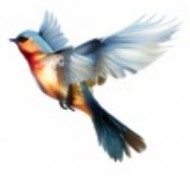} &
    \includegraphics[width=0.086\textwidth]{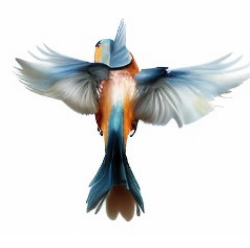} &
    \includegraphics[width=0.086\textwidth]{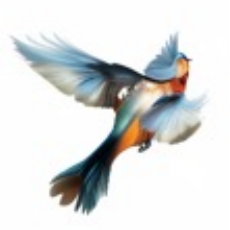} &
    \includegraphics[width=0.086\textwidth]{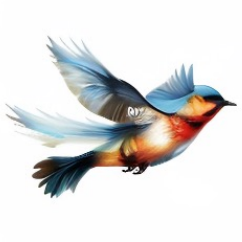} &
    \includegraphics[width=0.086\textwidth]{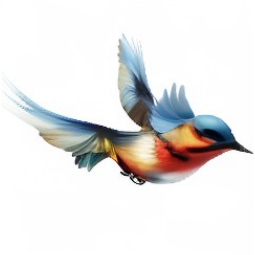} &
    \includegraphics[width=0.086\textwidth]{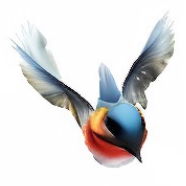} &
    \includegraphics[width=0.086\textwidth]{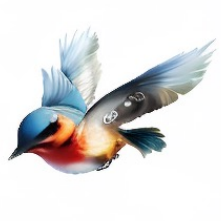} &
    \includegraphics[width=0.086\textwidth]{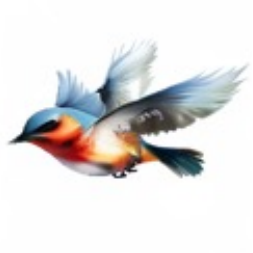} &
    \includegraphics[width=0.086\textwidth]{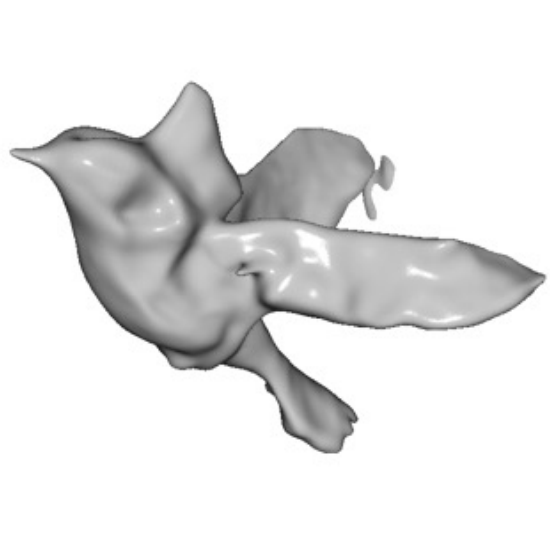} &
    \includegraphics[width=0.086\textwidth]{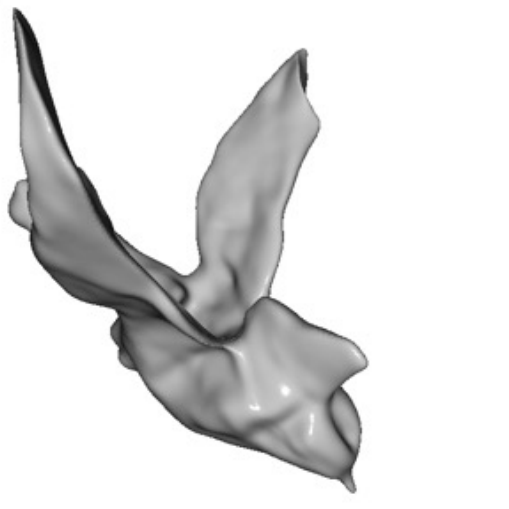} \\
    \multicolumn{1}{c}{Input} & \multicolumn{8}{c}{Generated diverse and multi-view coherent images} & \multicolumn{2}{c}{Mesh} \\
    \end{tabular}
    }

    \vspace{2mm}
    
    \resizebox{\linewidth}{!}{
    \begin{tabular}{c:ccccc:cccccc}
    \includegraphics[width=0.086\textwidth]{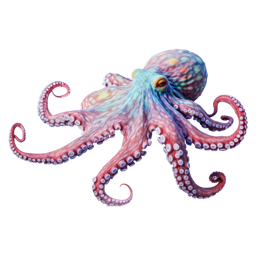} &
    \includegraphics[width=0.086\textwidth]{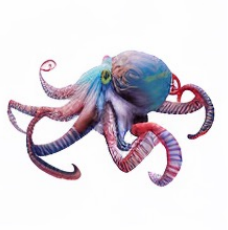} &
    \includegraphics[width=0.086\textwidth]{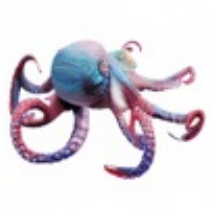} &
    \includegraphics[width=0.086\textwidth]{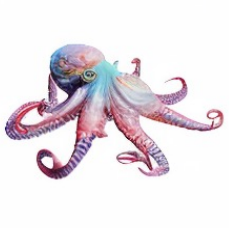} &
    \includegraphics[width=0.086\textwidth]{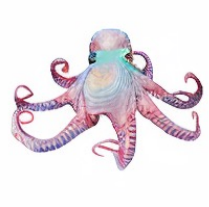} &
    \includegraphics[width=0.086\textwidth]{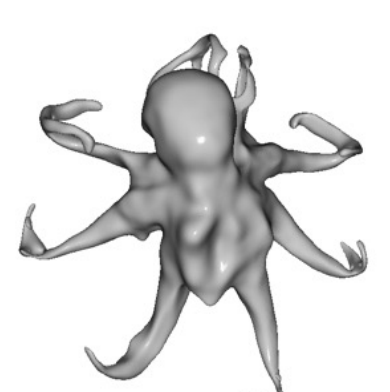} &
    \includegraphics[width=0.086\textwidth]{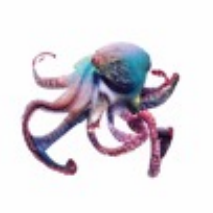} &
    \includegraphics[width=0.086\textwidth]{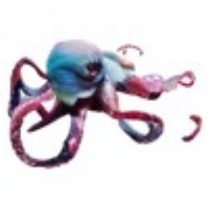} &
    \includegraphics[width=0.086\textwidth]{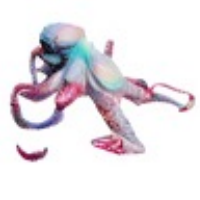} &
    \includegraphics[width=0.086\textwidth]{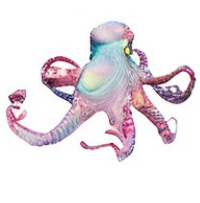} &
    \includegraphics[width=0.086\textwidth]{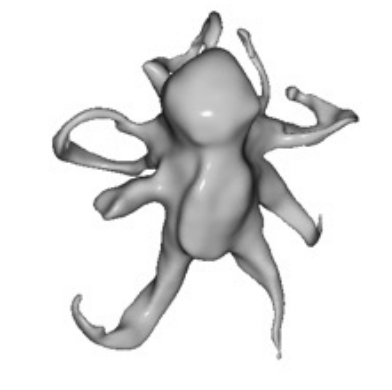} \\
    \includegraphics[width=0.086\textwidth]{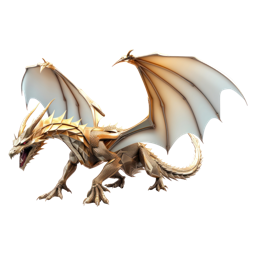} &
    \includegraphics[width=0.086\textwidth]{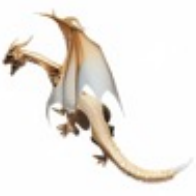} &
    \includegraphics[width=0.086\textwidth]{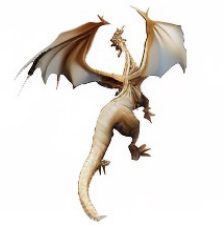} &
    \includegraphics[width=0.086\textwidth]{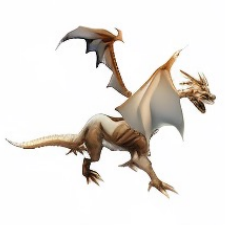} &
    \includegraphics[width=0.086\textwidth]{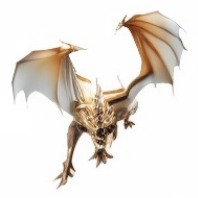} &
    \includegraphics[width=0.086\textwidth]{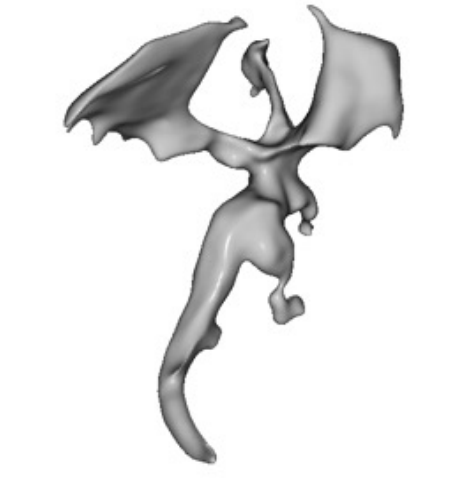} &
    \includegraphics[width=0.086\textwidth]{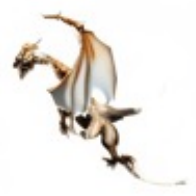} &
    \includegraphics[width=0.086\textwidth]{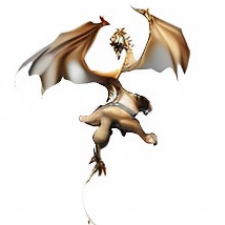} &
    \includegraphics[width=0.086\textwidth]{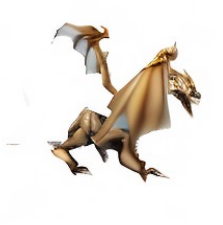} &
    \includegraphics[width=0.086\textwidth]{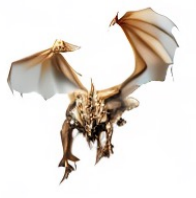} &
    \includegraphics[width=0.086\textwidth]{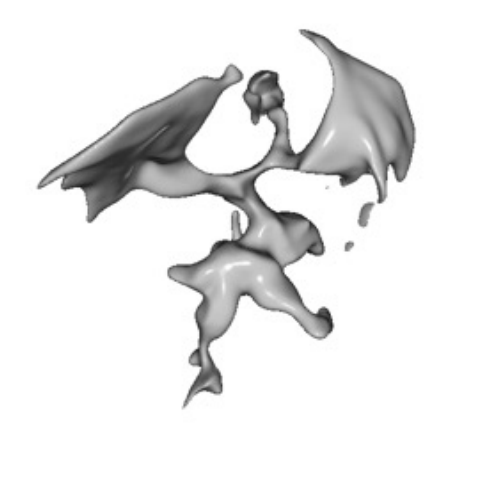} \\
    \includegraphics[width=0.086\textwidth]{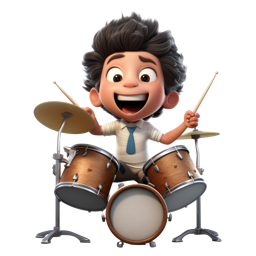} &
    \includegraphics[width=0.086\textwidth]{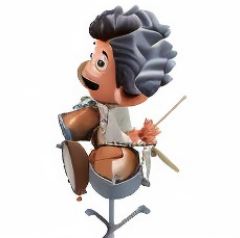} &
    \includegraphics[width=0.086\textwidth]{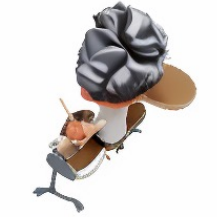} &
    \includegraphics[width=0.086\textwidth]{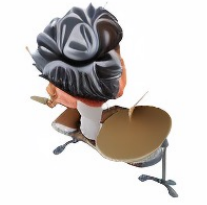} &
    \includegraphics[width=0.086\textwidth]{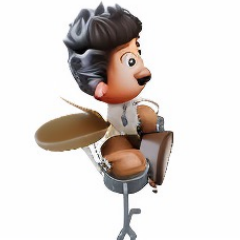} &
    \includegraphics[width=0.086\textwidth]{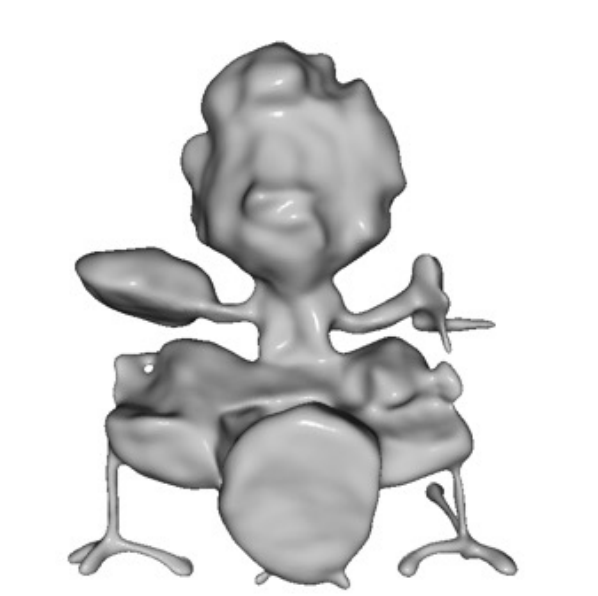} &
    \includegraphics[width=0.086\textwidth]{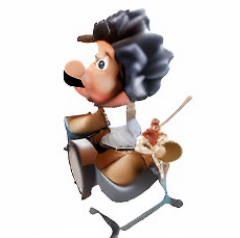} &
    \includegraphics[width=0.086\textwidth]{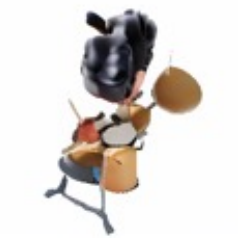} &
    \includegraphics[width=0.086\textwidth]{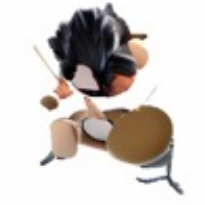} &
    \includegraphics[width=0.086\textwidth]{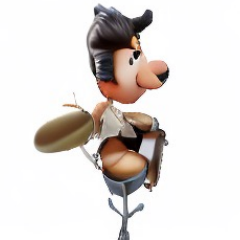} &
    \includegraphics[width=0.086\textwidth]{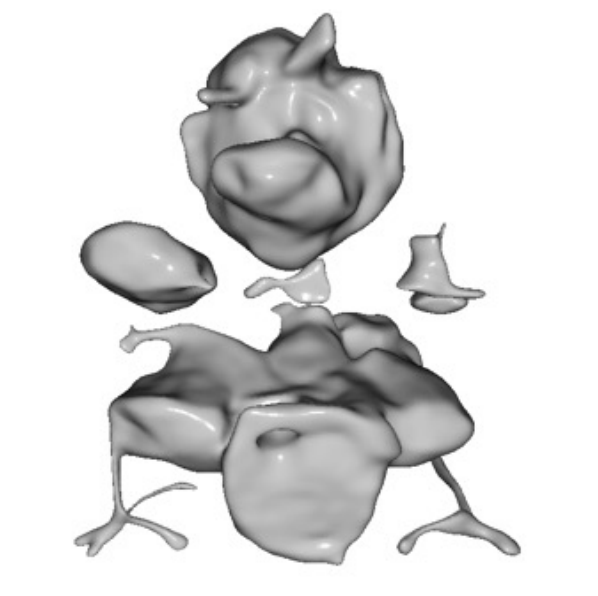} \\
    \includegraphics[width=0.086\textwidth]{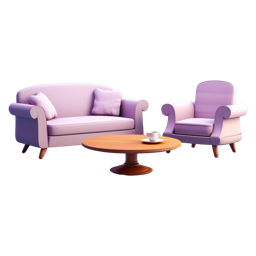} &
    \includegraphics[width=0.086\textwidth]{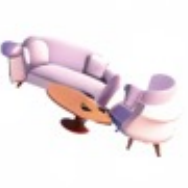} &
    \includegraphics[width=0.086\textwidth]{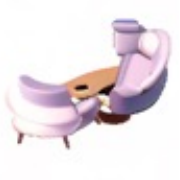} &
    \includegraphics[width=0.086\textwidth]{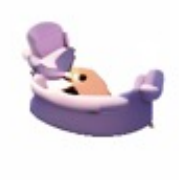} &
    \includegraphics[width=0.086\textwidth]{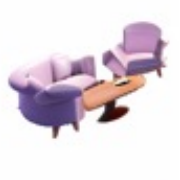} &
    \includegraphics[width=0.086\textwidth]{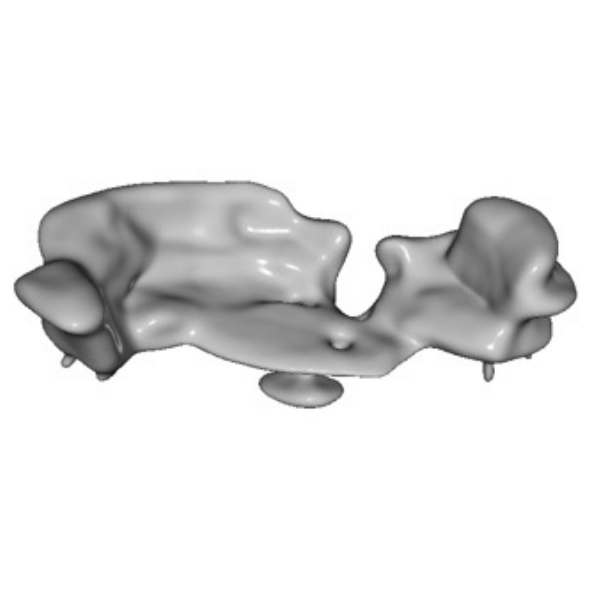} &
    \includegraphics[width=0.086\textwidth]{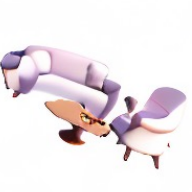} &
    \includegraphics[width=0.086\textwidth]{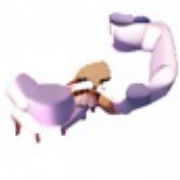} &
    \includegraphics[width=0.086\textwidth]{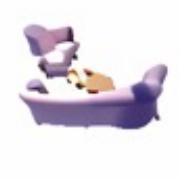} &
    \includegraphics[width=0.086\textwidth]{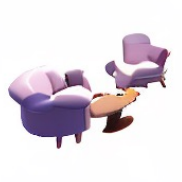} &
    \includegraphics[width=0.086\textwidth]{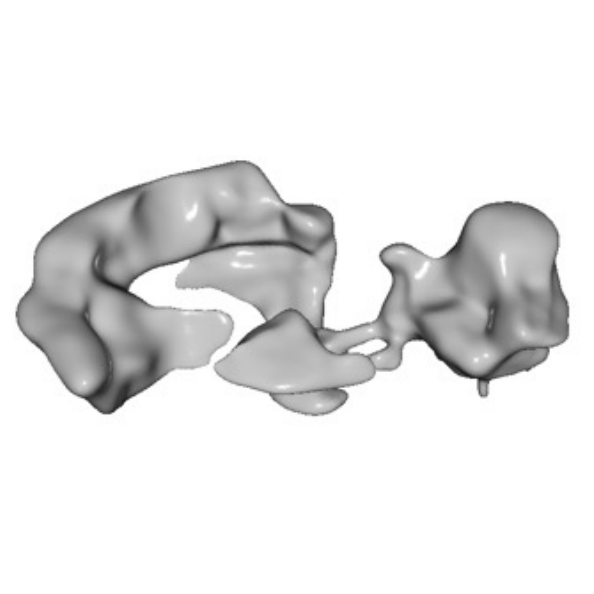} \\
    \multicolumn{1}{c}{Input} & \multicolumn{5}{c}{HarmonyView (Ours)} & \multicolumn{5}{c}{SyncDreamer~\cite{liu2023syncdreamer}} & \\
    \end{tabular}
    }
    \vspace{\abovefigcapmargin}
    \captionof{figure}{
        \textbf{HarmonyView for one-image-to-3D.}
        HarmonyView generates realistic 3D content using just a single image.
        It excels at maintaining visual and geometric consistency across generated views while enhancing the diversity of novel views, even in complex scenes.
    }
    \vspace{\belowfigcapmargin}
    \label{fig:teaser}
\end{strip}

%% file: contents/01_intro.tex
\section{Introduction}
\label{sec:intro}

Humans can effortlessly imagine the 3D form of an object from just a single camera view, drawing upon their prior knowledge of the 3D world.
Yet, emulating this human capability in machines remains a longstanding challenge in the field of computer vision~\cite{chang2015shapenet,tulsiani2016learning,sinha2017surfnet,wang2018pixel2mesh,zhang2018learning, park2022bridging}.
The fundamental hurdle lies in the inherent ambiguity of deducing 3D structure from a single 2D image since a single image essentially collapses the three dimensions of the real world into a 2D representation.
Consequently, countless 3D configurations of an object can be projected onto the same 2D image.
This ambiguity has ignited the quest for innovative solutions for single-image 3D generation~\cite{tang2023make,weng2023zeroavatar,szymanowicz2023viewset,liu2023one,qian2023magic123,tang2023mvdiffusion,zou2023sparse3d,shi2023mvdream,liu2023syncdreamer,lin2023consistent123,jiang2023efficient,ye2023consistent,weng2023consistent123,yang2023consistnet,shi2023toss,long2023wonder3d,shi2023zero123++,kant2023invs,sargent2023zeronvs,zhou2023sparsefusion,chan2023generative}.

One prevalent strategy is to generate multi-view images from a single 2D image~\cite{watson2022novel,liu2023zero,szymanowicz2023viewset,liu2023one}, and process them using techniques such as Neural Radiance Fields (NeRFs)~\cite{mildenhall2021nerf} to create 3D representations.
Regarding this, recent studies~\cite{watson2022novel,liu2023zero,liu2023syncdreamer,ye2023consistent,yang2023consistnet,szymanowicz2023viewset} highlight the importance of maintaining \textit{multi-view coherency}.
This ensures that the generated 3D objects to be coherent across diverse viewpoints, empowering NeRF to produce accurate and realistic 3D reconstructions. 
To achieve this, researchers harness the capabilities of large-scale diffusion models~\cite{rombach2022high}, particularly those trained on a vast collection of 2D images.
The abundance of 2D images provides a rich variety of views for the same object, allowing the model to learn view-to-view relationships and acquire geometric priors about the 3D world.
On top of this, some works~\cite{liu2023syncdreamer,szymanowicz2023viewset} introduce a refinement stage that fine-tunes the view alignment to accommodate variations in camera angles.
This adjustment is a key factor in achieving the desired multi-view coherency, which directly impacts the realism of the resulting 3D representation.
This progress has notably enhanced the utility of the generated 3D contents, making them more suitable for various applications~\cite{prince20023d,wibirama2020physical}.

An equally significant but often overlooked aspect in single-image 3D generation is the \textit{novel-view diversity}.
The ill-posed nature of this task necessitates dealing with numerous potential 3D interpretations of a given 2D image.
Recent works~\cite{wang2023prolificdreamer,liu2023zero,liu2023syncdreamer,szymanowicz2023viewset} showcase the potential of creating diverse 3D contents by leveraging the capability of diffusion models in generating diverse 2D samples.
However, balancing the pursuit of consistency and diversity remains a challenge due to their inherent trade-off: maintaining \textit{visual consistency} between generated multi-view images and the input view image directly contributes to sample quality but comes at the cost of limiting \textit{diversity}.
Although current multi-view diffusion models~\cite{szymanowicz2023viewset,liu2023syncdreamer} attempt to optimize both aspects simultaneously, they fall short of fully unraveling their intricacies.
This poses a crucial question: Can we navigate towards a harmonious balance between these two fundamental aspects in single-image 3D generation, thereby unlocking their full potential?

This work aims to address this question by introducing a simple yet effective diffusion sampling technique, termed HarmonyView.
This technique effectively decomposes the intricacies in balancing consistency and diversity, enabling a more nuanced exploration of these two fundamental facets in single-image 3D generation.
Notably, HarmonyView provides a means to exert explicit control over the sampling process, facilitating a more refined and controlled generation of 3D contents.
This versatility of HarmonyView is illustrated in~\cref{fig:teaser}.
Our method achieves a harmonious balance, demonstrating mutual benefits in both consistency and diversity.
HarmonyView generates geometrically coherent 3D contents that faithfully represent the input image for visible parts while also capturing diverse yet plausible modes for occluded parts.
Another challenge we face is the absence of standardized metrics for assessing the diversity of generated multi-views.
To address this gap and provide a more comprehensive assessment of the consistency and diversity of 3D contents, we introduce a novel evaluation metric based on both the CLIP image and text encoders~\cite{radford2021learning,hessel2021clipscore}.

In experiments, we quantitatively compare HarmonyView against state-of-the-art techniques, spanning two tasks: novel-view synthesis and 3D reconstruction.
In both tasks, HarmonyView consistently outperforms baseline methods across all metrics.
Our qualitative results further highlight the efficacy of HarmonyView, showcasing faithful reconstructions with remarkable visual quality, even in complex scenes.
Moreover, we show that our proposed metric closely aligns with the assessments made by human evaluators.
Lastly, HarmonyView can be seamlessly integrated with off-the-shelf text-to-image diffusion models (\eg, Stable Diffusion~\cite{rombach2022high}), enabling it to perform text-to-image-to-3D generation.

%% file: contents/02_related.tex
\section{Related Work}
\label{sec:related}
\paragraph{Lifting 2D pretrained models for 3D generation.}
Recent research endeavors~\cite{lin2023magic3d,chen2023fantasia3d,wang2023prolificdreamer,wang2023score,lorraine2023att3d,weng2023zeroavatar,tang2023mvdiffusion,zou2023sparse3d,shi2023mvdream} are centered on the idea of lifting 2D pre-trained models~\cite{rombach2022high,radford2021learning} to create 3D models from textual prompts, without the need for explicit 3D data.
The key insight lies in leveraging 3D priors acquired by diffusion models during pre-training on Internet-scale data.
This enables them to \textit{dream up} novel 3D shapes guided by text descriptions.
DreamFusion~\cite{poole2022dreamfusion} distills pre-trained Stable Diffusion~\cite{rombach2022high} using \textit{Score Distillation Sampling (SDS)} to extract a Neural Radiance Field (NeRF)~\cite{mildenhall2021nerf} from a given text prompt.
DreamFields~\cite{jain2022zero} generates 3D models based on text prompts by optimizing the CLIP~\cite{radford2021learning} distance between the CLIP text embedding and NeRF~\cite{mildenhall2021nerf} renderings.
However, accurately representing 3D details with word embeddings remains a challenge.

Similarly, some works~\cite{xu2023neurallift,melas2023realfusion,tang2023make,qian2023magic123} extend the distillation process to train NeRF for the 2D-to-3D task.
NeuralLift-360~\cite{xu2023neurallift} utilizes a depth-aware NeRF to generate scenes guided by diffusion models and incorporates a distillation loss for CLIP-guided diffusion prior~\cite{radford2021learning}.
Magic123~\cite{qian2023magic123} uses SDS loss to train a NeRF and then fine-tunes a mesh representation.
Due to the reliance on SDS loss, these methods necessitate textual inversion~\cite{gal2022image} to find a suitable text description for the input image.
Such a process needs per-scene optimization, making it time-consuming and requiring tedious parameter tuning for satisfactory quality.

Another line of work~\cite{watson2022novel,liu2023zero,szymanowicz2023viewset,liu2023one} uses 2D diffusion models to generate multi-view images then use them for 3D reconstruction with NeRF~\cite{mildenhall2021nerf,wang2021neus}.
3DiM~\cite{watson2022novel} views novel-view synthesis as an image-to-image translation problem and uses a pose-conditional diffusion model to predict novel views from an input view.
Zero-1-to-3~\cite{liu2023zero} enables zero-shot 3D creation from arbitrary images by fine-tuning Stable Diffusion~\cite{rombach2022high} with relative camera pose.
Our work, falling into this category, is able to convert arbitrary 2D images to 3D without SDS loss~\cite{poole2022dreamfusion}.
It seamlessly integrates with other frameworks, such as text-to-2D~\cite{ramesh2021zero,nichol2021glide,rombach2022high} and neural reconstruction methods~\cite{mildenhall2021nerf,wang2021neus}, streamlining the text-to-image-to-3D process.
Unlike prior distillation-based methods~\cite{xu2023neurallift,melas2023realfusion} confined to a singular mode, our approach offers greater flexibility for generating diverse 3D contents.

\vspace{\paramargin}
\paragraph{Consistency and diversity in 3D generation.}
The primary challenge in single-image 3D content creation lies in maintaining multi-view coherency. 
Various approaches~\cite{watson2022novel,liu2023zero,liu2023syncdreamer,ye2023consistent,yang2023consistnet} attempt to tackle this challenge:
Viewset Diffusion~\cite{szymanowicz2023viewset} utilizes a diffusion model trained on multi-view 2D data to output 2D viewsets and corresponding 3D models.
SyncDreamer~\cite{liu2023syncdreamer} introduces a 3D-aware feature attention that synchronizes intermediate states of noisy multi-views.
Despite these efforts, achieving complete geometric coherence in generated views remains a challenge.

On the other hand, diversity across generated 3D samples is another critical aspect in single-image 3D generation.
However, only a few works in the related literature specifically address this issue, often limited to domains such as face generation~\cite{dey2022generating} or starting from text for 3D generation~\cite{wang2023prolificdreamer}.
Recent studies~\cite{liu2023zero,szymanowicz2023viewset,liu2023syncdreamer,ye2023consistent} showcase the potential of pre-trained diffusion models~\cite{rombach2022high} in generating diverse multi-view images.
However, there is still significant room for exploration in balancing consistency and diversity.
In our work, we aim to unlock the potential of diffusion models, allowing for reasoning about diverse modes for novel views while being faithful to the input view for observable parts.
We achieve this by breaking down the formulation of multi-view diffusion model into two fundamental aspects: \textit{visual consistency} with input view and \textit{diversity} of novel views.
Additionally, we propose the CD score to address the absence of a standardized diversity measure in existing literature.

%% file: contents/03_method.tex
\section{Method}
\label{sec:method}
Our goal is to create a high-quality 3D object from a single input image, denoted as $\rvy$.
To achieve this, we use the diffusion model~\cite{song2020denoising} to generate a cohesive set of $N$ views at pre-defined viewpoints, denoted as ${\rvx}^{(1:N)}_0 = \{{\rvx}^{(1)}_0,...,{\rvx}^{(N)}_0\}$. 
These mutli-view images are then utilized in NeRF-like techniques~\cite{mildenhall2021nerf,wang2021neus} for 3D reconstruction.
The key to a realistic 3D object lies in the consistency across the generated views.
If they exhibit coherent appearance and geometry, the resulting 3D object will appear more natural. 
Therefore, ensuring consistency is crucial for achieving our goal.
Recent works~\cite{szymanowicz2023viewset,liu2023syncdreamer,shi2023zero123++} address multi-view generation by jointly optimizing the distribution of multiple views.
Building upon them, we aim to enhance both consistency and diversity by decomposing their formulation during diffusion sampling.

\subsection{Diffusion Models}
We address the challenge of generating a 3D representation from a single, partially observed image using diffusion models~\citep{sohl2015deep,song2020denoising}.
These models inherently possess the capability to capture diverse modes~\cite{xiao2021tackling}, making them well-suited for the task.
We adopt the setup of DDPM~\cite{ho2020denoising}, which defines a \textit{forward diffusion process} transforming an initial data sample ${\rvx}_0$ into a sequence of noisy samples ${\rvx}_1, \dots, {\rvx}_T$ over $T$ steps, approximating a Gaussian noise distribution.
In practice, we perform the forward process by directly transitioning to a noised version of a sample using the equation:
\begin{equation}
{\rvx}_t = \sqrt{\bar{\alpha}_t}{\rvx}_0 + \sqrt{1-\bar{\alpha}_t} {\bm{\epsilon}},
\label{eq:forward_process}
\end{equation}
where ${\bm{\epsilon}} \sim \mathcal{N}(0, \mathbf{I})$ is a Gaussian noise, $\bar{\alpha}_t$ is a noise schedule monotonically decreasing with timestep $t$ (with $\bar{\alpha}_0 = 1$), and ${\rvx}_t$ is a noisy version of the input ${\rvx}_0$ at timestep $t$.

The \textit{reverse denoising process} ``undo" the forward steps to recover the original data from noisy observations.
Typically, this process is learned by optimizing a noise prediction model ${\bm{\epsilon}}_{\theta}({\rvx}_t, t)$ on a data distribution $q(x_0)$.
DDPM~\cite{ho2020denoising} defines the following simple loss:
\begin{equation}
\mathcal{L}_{simple} = \E_{{\rvx}_0 \sim q({\rvx}_0), {\bm{\epsilon} \sim \mathcal{N}(0, 1)}, t \sim U[1, T]} \| {\bm{\epsilon}} - {\bm{\epsilon}}_{\theta}({\rvx}_t; t) \|_2^2.
\label{eq:simple_loss}
\end{equation}
\subsection{Multi-view Diffusion Models}
SyncDreamer~\cite{liu2023syncdreamer} introduces a multi-view diffusion model that captures the \textit{joint distribution} of $N$ novel views ${\rvx}^{(1:N)}_0$ given an input view ${\rvy}$.
This model extends the DDPM forward process (\cref{eq:forward_process}) by adding random noises independently to each view at every time step:
\begin{equation}
{\rvx}^{(n)}_t = \sqrt{\bar{\alpha}_t}{\rvx}^{(n)}_0 + \sqrt{1-\bar{\alpha}_t} {\bm{\epsilon}}^{(n)}.
\label{eq:mv_forward}
\end{equation}
Here, $n$ denotes the view index.
A noise prediction model $\bm{\epsilon}_\theta$ predicts the noise of the $n$-th view $\bm{\epsilon}^{(n)}$, given the condition of an input view $\rvy$, the view difference between the input view and the $n$-th target view $\Delta{\rvv}^{(n)}$, and noisy multi views ${\rvx}^{(1:N)}_t$.
Hereafter, we define the pair $({\rvy}, \Delta {\rvv}^{(n)})$ as the reference view condition ${\rvr}^{(n)}$ to simplify notation.
Similar to \cref{eq:simple_loss}, the loss for the noise prediction model is defined as:
\begin{equation}
    \mathcal{L} = \E_{{\rvx}^{(1:N)}_0, \bm{\epsilon}^{(1:N)}, t} \|\bm{\epsilon}^{(n)} - \bm{\epsilon}_\theta ({\rvx}^{(n)}; t, {\rvc}^{(n)})\|_2^2,
\label{eq:mv_loss}
\end{equation}
where ${\rvc}^{(n)} = ({\rvr}^{(n)}, {\rvx}^{(1:N)}_t)$ and $\bm{\epsilon}^{(1:N)}$ represents Gaussian noise of size $N \times H \times W$ added to all $N$ views.

\subsection{HarmonyView}
\paragraph{Diffusion sampling guidance.}
Classifier-guided diffusion~\cite{dhariwal2021diffusion} uses a noise-robust classifier $p({\vl}|{\rvx}_t)$, which estimates the class label $\vl$ given a noisy sample ${\rvx}_t$, to guide the diffusion process with gradients $\nabla_{{\rvx}_t} \log p({\vl}|{\rvx}_t)$.
This classifier requires bespoke training to cope with high noise levels (where timestep $t$ is large) and to provide meaningful signals all the way through the sampling process.
Classifier-free guidance~\cite{ho2022classifier} uses a single conditional diffusion model $p_{\theta}({\rvx}|{\vl})$ with conditioning dropout, which intermittently replaces $\vl$ (typically 10\%) with a null token $\phi$ (representing the absence of conditioning information) for unconditional predictions.
This models an \textit{implicit classifier} directly from a diffusion model without the need for an extra classifier trained on noisy input.
These conditional diffusion models~\cite{dhariwal2021diffusion,ho2022classifier} dramatically improve sample quality by enhancing the conditioning signal but with a trade-off in diversity.


\vspace{\paramargin}
\paragraph{What's wrong with multi-view diffusion sampling?}
From~\cref{eq:mv_loss}, we derive an \textit{unconditional} diffusion model $p({\rvx}^{(n)})$ parameterized by a score estimator $\bm{\epsilon}_{\theta}({\rvx}^{(n)}_t; t)$ and \textit{conditional} diffusion model $p({\rvx^{(n)}}|{\rvc}^{(n)})$ parameterized by $\bm{\epsilon}_{\theta}({\rvx}^{(n)}_t; t,{\rvc}^{(n)}_t)$.
These two models are learned via a single neural network following the classifier-free guidance~\cite{ho2022classifier}.
During sampling, the multi-view diffusion model adjusts its prediction as follows ($t$ is omitted for clarity):
\begin{equation}
    \resizebox{\linewidth}{!}{$
    \hat{\bm{\epsilon}}_{\theta}({\rvx}^{(n)}_t; {\rvc}^{(n)}) = \bm{\epsilon}_{\theta}({\rvx}^{(n)}_t; {\rvc}^{(n)}) + s \cdot (\bm{\epsilon}_{\theta}({\rvx}^{(n)}_t; {\rvc}^{(n)}) - {\bm{\epsilon}}_{\theta}({\rvx}^{(n)}_t)),
    $}
\label{eq:mv_cfg}
\end{equation}
where $s$ represents a guidance scale.

The model output is extrapolated further in the direction of $\bm{\epsilon}_{\theta}({\rvx}^{(n)}_t; {\rvc}^{(n)}_t)$ and away from $\bm{\epsilon}_{\theta}({\rvx}^{(n)}_t)$.
Remind that ${\rvc}^{(n)} = ({\rvr}^{(n)}, {\rvx}^{(1:N)}_t)$.
Thus, the scaling of $s$ affects both the input view condition ${\rvr}^{(n)}$ and the multi-view condition ${\rvx}^{(1:N)}_t$ simultaneously.
As evidenced by~\Cref{tab:scale_study}, increasing $s$ encourages \textit{multi-view coherency} and \textit{diversity} in the generated views.
Yet, this comes with a trade-off: it simultaneously diminishes the \textit{visual consistency} with the input view. 
While the inherent trade-off between these two dimensions is obvious in this context, managing competing objectives under a single guidance poses a considerable challenge.
In essence, the model tends to generate diverse and geometrically coherent multi-view images, but differ in visual aspects (\eg, color, texture) from the input view, resulting in sub-optimal quality.
Empirical observations, shown in~\cref{fig:qual_ablation} and~\Cref{tab:ablation_study}, substantiate that this formulation manifests a conflict between the objectives of consistency and diversity.

\vspace{\paramargin}
\paragraph{Harmonizing consistency and diversity.}
\input{figs/ablation}
To address the aforementioned challenge, we introduce a method termed ``HarmonyView".
Our approach leverages two implicit classifiers.
One classifier $p^{i}({\rvr}^{(n)}|{\rvx}^{(n)}_t, {\rvx}^{(1:N)}_{t})$ guides the target view ${\rvx}^{(n)}_t$ and multi-views ${\rvx}^{(1:N)}_t$ to be more \textit{visually consistent} with the input view ${\rvr}^{(n)}$.
Another classifier $p^i({\rvx}^{(1:N)}_{t} | {\rvx}^{(n)}_t, {\rvr}^{(n)})$ contains uncertainty in both the target (${\rvx}^{(1:N)}_{t}$) and conditional (${\rvx}^{(n)}_t$) elements.
This contributes to capturing \textit{diverse} modes.
Together, they synergistically guide the synchronization of noisy multi-views ${\rvx}^{(1:N)}_t$, facilitating \textit{geometric coherency} among clean multi-views.
Based on these, we redefine the score estimation as follows:
\begin{equation}
\begin{split}
    \tilde{\bm{\epsilon}}_\theta ({\rvx}^{(n)}_{t}; {\rvc}^{(n)})
    & = \bm{\epsilon}_\theta ({\rvx}^{(n)}_{t}; {\rvc}^{(n)}) \\
    & -s_{1}\sigma_{t}\nabla_{{\rvx}^{(n)}_t} \log p^{i}({\rvr}^{(n)}|{\rvx}^{(n)}_t, {\rvx}^{(1:N)}_{t}) \\
    & -s_{2}\sigma_{t}\nabla_{{\rvx}^{(n)}_t} \log p^{i}({\rvx}^{(1:N)}_{t} | {\rvx}^{(n)}_t, {\rvr}^{(n)}),
\end{split}
\label{eq:ours_initial}
\end{equation}
where $s_1$ and $s_2$ are guidance scales and ${\sigma}_{t}$ is a noise scheduling parameter.
By properly balancing these terms, we can obtain multi-view coherent images that align well with the semantic content of the input image while being diverse across different samples.

According to Bayes' rule, $p^{i}({\rvr}^{(n)}|{\rvx}^{(n)}_t, {\rvx}^{(1:N)}_{t}) \propto {p({\rvx}^{(n)}_t | {\rvc}^{(n)})} / {p({\rvx}^{(n)}_t | {\rvx}^{(1:N)}_{t})}$ and
$p^{i}({\rvx}^{(1:N)}_{t} | {\rvx}^{(n)}_t, {\rvr}^{(n)}) \propto {p({\rvx}^{(n)}_t | {\rvc}^{(n)})} / {p({\rvx}^{(n)}_t | {\rvr}^{(n)})}$.
Hence, the diffusion scores of these two implicit classifiers can be derived as follows:
\begin{equation}
\begin{split}
    \nabla_{{\rvx}^{(n)}_t} & \log p^{i}({\rvr}^{(n)}|{\rvx}^{(n)}_t, {\rvx}^{(1:N)}_{t}) \\
    & = -\frac{1}{\sigma_t}(\bm{\epsilon}_\theta ({\rvx}^{(n)}_t; {\rvc}^{(n)}) - \bm{\epsilon}_\theta ({\rvx}^{(n)}_t; {\rvx}^{(1:N)}_{t})).
\end{split}
\end{equation}
\begin{equation}
\begin{split}
    \nabla_{{\rvx}^{(n)}_t} & \log p^{i}({\rvx}^{(1:N)}_{t} | {\rvx}^{(n)}_t, {\rvr}^{(n)}) \\
    & = -\frac{1}{\sigma_t}(\bm{\epsilon}_\theta ({\rvx}^{(n)}_t; {\rvc}^{(n)}) - \bm{\epsilon}_{\theta}({\rvx}^{(n)}_{t};{\rvr}^{(n)}).
\end{split}
\end{equation}
Finally, these terms are plugged into~\cref{eq:ours_initial} and yields:
\begin{equation}
\begin{split}
    \tilde{\bm{\epsilon}}_\theta ({\rvx}^{(n)}_{t};& {\rvc}^{(n)}) = \bm{\epsilon}_\theta ({\rvx}^{(n)}_{t}; {\rvc}^{(n)}) \\
    & + s_{1} \cdot (\bm{\epsilon}_\theta ({\rvx}^{(n)}_{t}; {\rvc}^{(n)}) - \bm{\epsilon}_\theta ({\rvx}^{(n)}_{t}; {\rvx}^{(1:N)}_{t}) \\
    & + s_{2} \cdot (\bm{\epsilon}_\theta ({\rvx}^{(n)}_{t}; {\rvc}^{(n)}) - \bm{\epsilon}_{\theta}({\rvx}^{(n)}_{t}; {\rvr}^{(n)}).
\end{split}
\label{eq:ours_final}
\end{equation}
This formulation effectively decomposes \textit{consistency} and \textit{diversity}, offering a nuanced approach that grants control over both dimensions.
While simple, our decomposition achieves a win-win scenario, striking a harmonious balance in generating samples that are both consistent and diverse (see~\cref{fig:qual_ablation} and~\Cref{tab:ablation_study}).

\input{tables/ablation}

\subsection{Consistency-Diversity (CD) Score}
\label{subsec:CD}
We propose the CD score with two key principles:
(1) \textit{Diversity of novel views}: It is preferable that the generated images exhibit diverse and occasionally creative appearances that are not easily imaginable from the input image.
(2) \textit{Semantic consistency}: While pursuing diversity, it is crucial to maintain semantic consistency, \ie, the generated images should retain their semantic content consistently, regardless of variations in the camera viewpoint.
To operationalize this evaluation, CD score utilizes CLIP~\cite{radford2021learning} image (${\Psi}_{I}$) and text encoders (${\Psi}_{T}$), akin to CLIP score~\cite{hessel2021clipscore}.
%
\input{figs/qualitative_nvs}
\textit{Diversity} ($D$) measures the average dissimilarity of generated views $\{{\rvx}^{(1)}, \dots, {\rvx}^{(N)}\}$ from a reference view ${\rvy}$, reflecting how distinct the generated images are from the reference view, emphasizing creative variations.
The diversity is computed by averaging the cosine similarity of each generated view with the reference view using CLIP image encoders.
\begin{equation}
    D = \frac{1}{N} \sum_{n=1}^{N} \left[ 1 - cos({\Psi}_{I}({\rvy}), {\Psi}_{I}({\rvx}^{(n)})) \right].
\end{equation}
\textit{Semantic variance} ($\text{S}_{Var}$) quantifies the variance in semantic changes across views.
This measures how similar the generated images are to a given text prompt, ``\texttt{An image of \{OBJECT\}}."
The semantic variance is calculated by averaging the cosine similarity between the CLIP text embedding of the prompt and the CLIP image embedding of each generated view, followed by measuring the variance of these values across views.
\begin{equation}
\begin{split}
    & \bar{\text{S}} = \frac{1}{N} \sum_{n=1}^{N} \cos({\Psi}_{T}(\texttt{text}), {\Psi}_{I}({\rvx}^{(n)})), \\
    & \text{S}_{Var} = \frac{1}{N} \sum_{n=1}^{N} ( \cos({\Psi}_{T}(\texttt{text}), {\Psi}_{I}({\rvx}^{(n)})) - \bar{\text{S}} )^2.
\end{split}
\end{equation}
The CD score is then computed as the ratio of diversity to semantic variances across views:
\begin{equation}
    \text{CD Score} = {D} / \text{S}_{Var}.
\end{equation}
We note that the CD score is reference-free, \ie, it does not require any ground truth images to measure the score.

%% file: figs/ablation.tex
\begin{figure*}
    \centering
    \setlength\tabcolsep{1pt}
    \resizebox{\linewidth}{!}{
    \renewcommand{\arraystretch}{0.5}
    \begin{tabular}{c:cc:cc:cc:cc:cc}
    \includegraphics[width=0.085\textwidth]{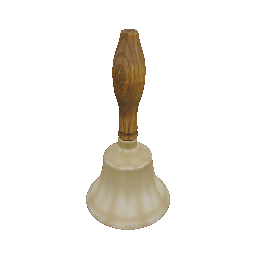} &
    \includegraphics[width=0.085\textwidth]{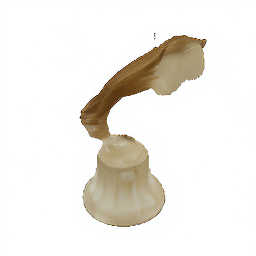} &
    \includegraphics[width=0.085\textwidth]{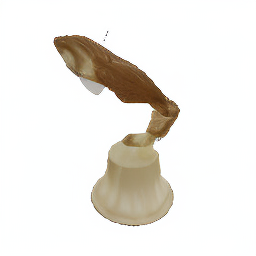} &
    \includegraphics[width=0.085\textwidth]{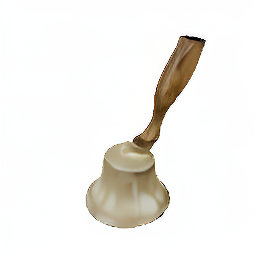} &
    \includegraphics[width=0.085\textwidth]{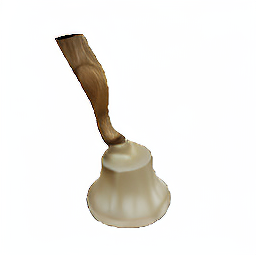} &
    \includegraphics[width=0.085\textwidth]{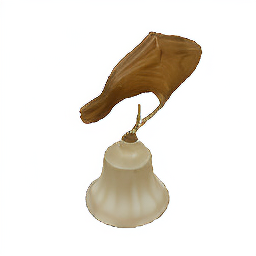} &
    \includegraphics[width=0.085\textwidth]{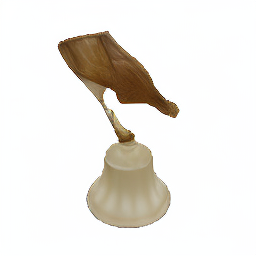} &
    \includegraphics[width=0.085\textwidth]{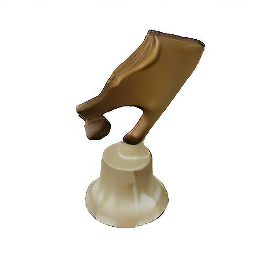} &
    \includegraphics[width=0.085\textwidth]{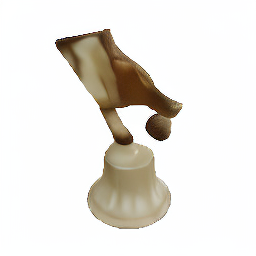} &
    \includegraphics[width=0.085\textwidth]{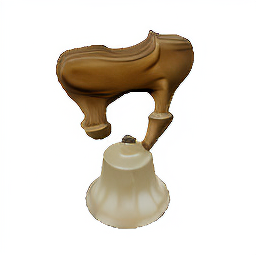} &
    \includegraphics[width=0.085\textwidth]{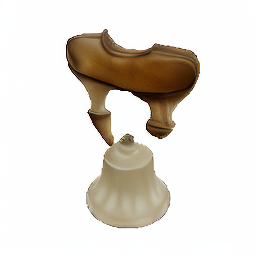} \\
    \includegraphics[width=0.085\textwidth]{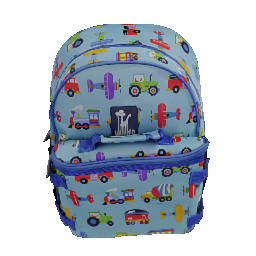} &
    \includegraphics[width=0.085\textwidth]{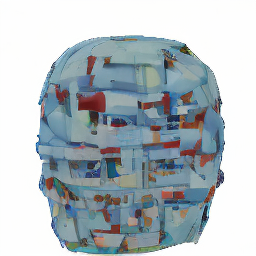} &
    \includegraphics[width=0.085\textwidth]{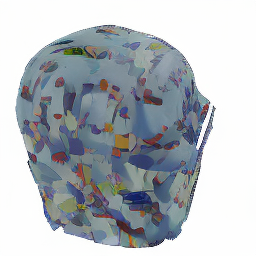} &
    \includegraphics[width=0.085\textwidth]{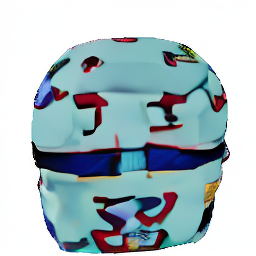} &
    \includegraphics[width=0.085\textwidth]{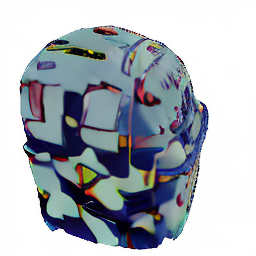} &
    \includegraphics[width=0.085\textwidth]{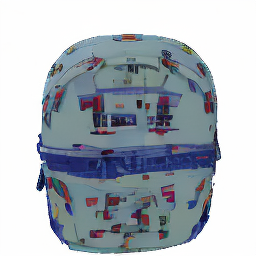} &
    \includegraphics[width=0.085\textwidth]{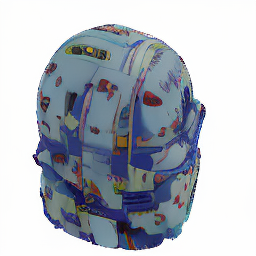} &
    \includegraphics[width=0.085\textwidth]{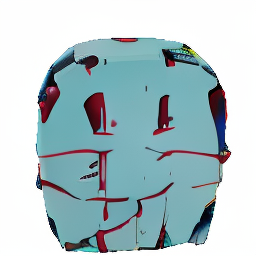} &
    \includegraphics[width=0.085\textwidth]{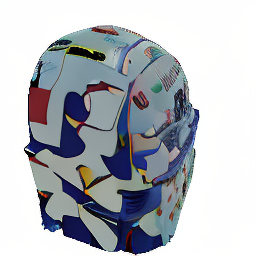} &
    \includegraphics[width=0.085\textwidth]{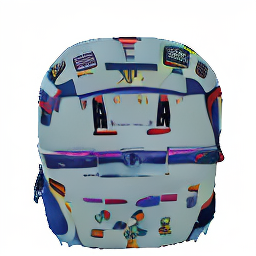} &
    \includegraphics[width=0.085\textwidth]{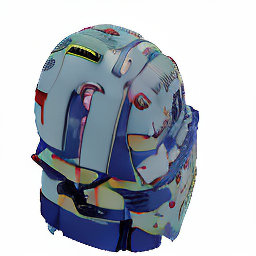} \\
     &
    \includegraphics[width=0.085\textwidth]{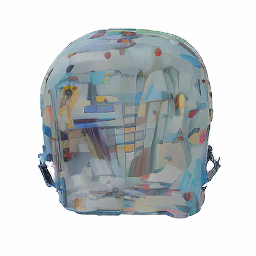} &
    \includegraphics[width=0.085\textwidth]{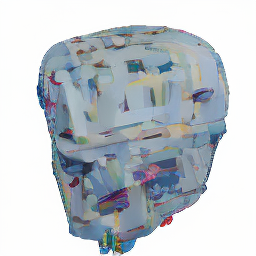} &
    \includegraphics[width=0.085\textwidth]{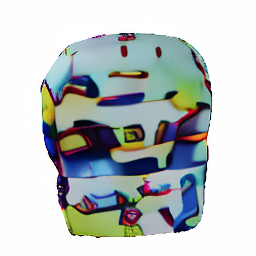} &
    \includegraphics[width=0.085\textwidth]{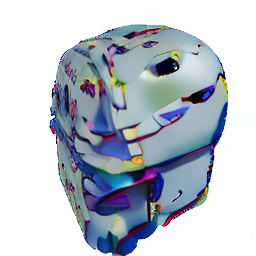} &
    \includegraphics[width=0.085\textwidth]{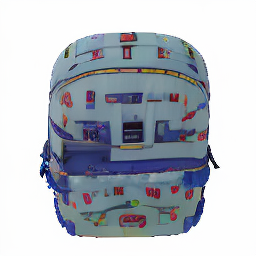} &
    \includegraphics[width=0.085\textwidth]{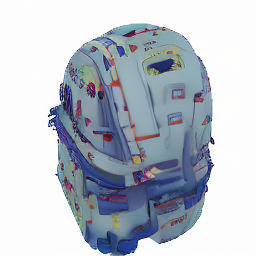} &
    \includegraphics[width=0.085\textwidth]{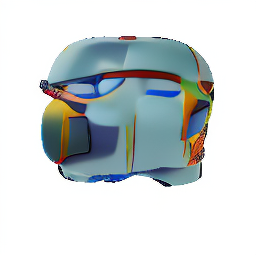} &
    \includegraphics[width=0.085\textwidth]{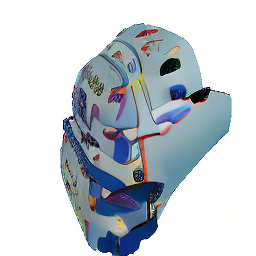} &
    \includegraphics[width=0.085\textwidth]{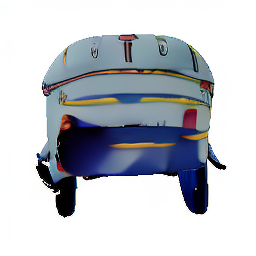} &
    \includegraphics[width=0.085\textwidth]{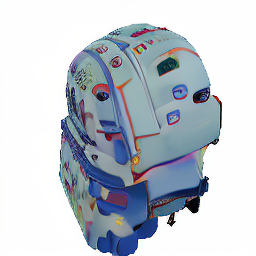} \\
    \multicolumn{1}{c}{Input} & \multicolumn{2}{c}{No Guidance} & \multicolumn{2}{c}{Baseline (\cref{eq:mv_cfg})} & \multicolumn{2}{c}{Only {\large $s_1$}} & \multicolumn{2}{c}{Only {\large $s_2$}} & \multicolumn{2}{c}{Ours (\cref{eq:ours_final})} \\
    \end{tabular}
    }
    \vspace{\abovefigcapmargin}
    \caption{
        \textbf{Qualitative comparison of several instantiations for multi-view diffusion guidance on novel-view synthesis.}
        Our decomposition of~\cref{eq:mv_cfg} yields two guidance parameters:
        $s_1$ for input-target visual consistency and $s_2$ for diversity in the novel views.
        With these parameters, our final formulation~\cref{eq:ours_final} enables the generation of a diverse set of multi-view coherent images that well reflect the input view.
    }
    \vspace{\belowfigcapmargin}
    \label{fig:qual_ablation}
\end{figure*}

%% file: tables/ablation.tex
\begin{table}[]
    \centering
    \setlength\tabcolsep{3pt}
    \resizebox{\linewidth}{!}{
   \begin{tabular}{lcccccccc}
       \toprule
       Method & {\large $s$} & {\large $s_1$} & {\large $s_2$} & PSNR$\uparrow$ & SSIM$\uparrow$ & LPIPS$\downarrow$ & $E_{flow}$$\downarrow$ & CD$\uparrow$ \\
       \midrule
       No Guidance &  &  &  & 20.51 & 0.818 & 0.144 & 2.270 & 0.640 \\
       Baseline (\cref{eq:mv_cfg}) & \checkmark &  &  & 20.19 & 0.819 & 0.140 & 2.071 & 0.717   \\
       \arrayrulecolor{gray}\cmidrule(lr){1-9}
       \multirow{3}{*}{Ours (\cref{eq:ours_final})} &  &  & \checkmark & 20.32 & 0.822 & 0.141 & 2.136 & 0.764 \\
       &  & \checkmark &  & \textbf{21.03} & \textbf{0.828} & \textbf{0.128} & 2.146 & 0.668 \\
       & \cellcolor{gray!25} & \cellcolor{gray!25}\checkmark & \cellcolor{gray!25}\checkmark & \cellcolor{gray!25}20.69 & \cellcolor{gray!25}0.825 & \cellcolor{gray!25}0.133 & \cellcolor{gray!25}\textbf{1.945} & \cellcolor{gray!25}\textbf{0.792} \\
       \arrayrulecolor{black}\bottomrule
    \end{tabular}
    }
    \vspace{\abovetabcapmargin}
    \caption{
        \textbf{Ablative study of multi-view diffusion guidance on novel-view synthesis.}
        Metrics measure sample quality with PSNR, SSIM, LPIPS; multi-view coherency with $E_{flow}$; and diversity with CD score. 
        Our final design strikes the best balance across the metrics. 
        Here, we set $s=1$, $s_{1}=2$, $s_{2}=1$.
    }
    \vspace{\belowtabcapmargin}
    \label{tab:ablation_study}
\end{table}

%% file: figs/qualitative_nvs.tex
\begin{figure*}
    \centering
    \setlength\tabcolsep{1pt}
    \resizebox{\linewidth}{!}{
    \renewcommand{\arraystretch}{0.5}
    \begin{tabular}{c:ccc:ccc:ccc}
        \includegraphics[width=0.095\textwidth]{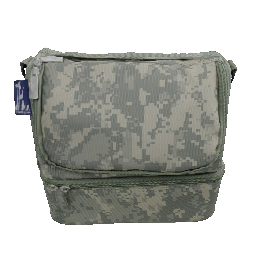} &
        \includegraphics[width=0.095\textwidth]{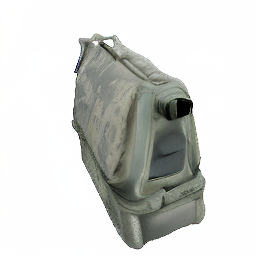} &
        \includegraphics[width=0.095\textwidth]{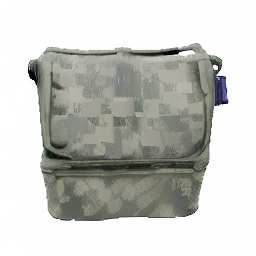} &
        \includegraphics[width=0.095\textwidth]{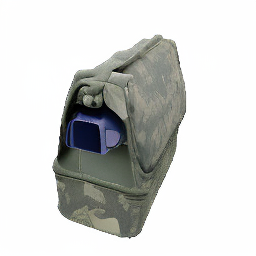} &
        \includegraphics[width=0.095\textwidth]{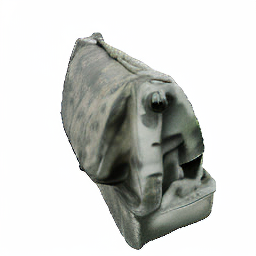} &
        \includegraphics[width=0.095\textwidth]{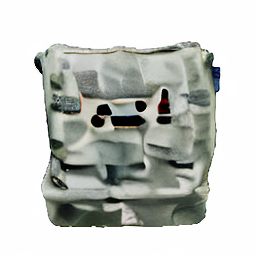} &
        \includegraphics[width=0.095\textwidth]{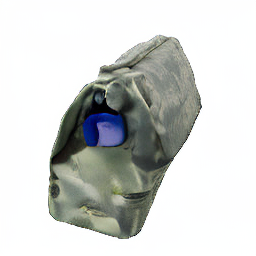} &
        \includegraphics[width=0.095\textwidth]{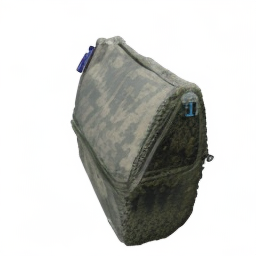} &
        \includegraphics[width=0.095\textwidth]{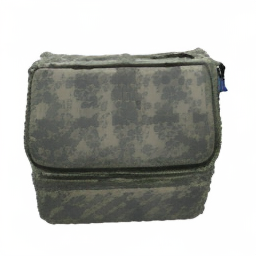} &
        \includegraphics[width=0.095\textwidth]{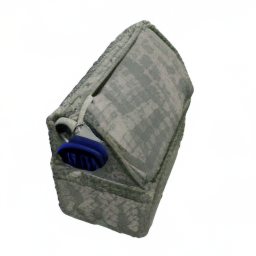} \\
         &
        \includegraphics[width=0.095\textwidth]{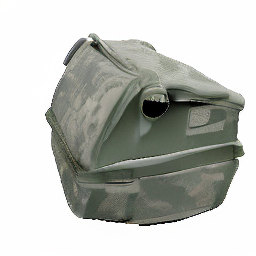} &
        \includegraphics[width=0.095\textwidth]{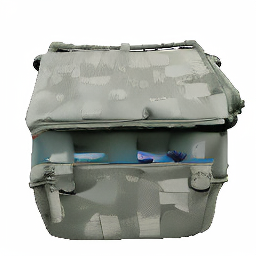} &
        \includegraphics[width=0.095\textwidth]{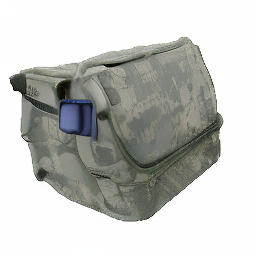} &
        \includegraphics[width=0.095\textwidth]{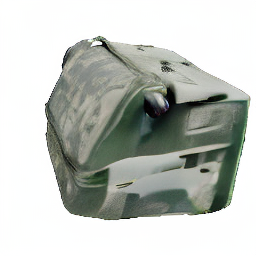} &
        \includegraphics[width=0.095\textwidth]{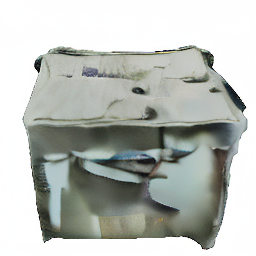} &
        \includegraphics[width=0.095\textwidth]{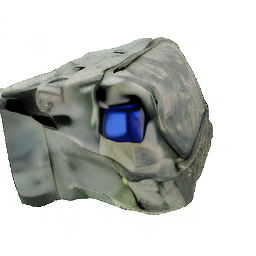} &
        \includegraphics[width=0.095\textwidth]{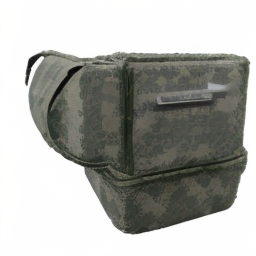} &
        \includegraphics[width=0.095\textwidth]{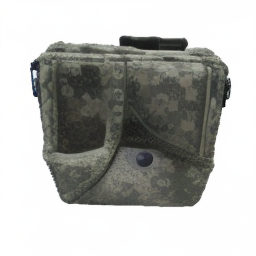} &
        \includegraphics[width=0.095\textwidth]{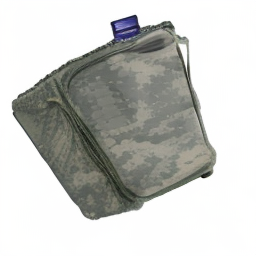} \\
        \includegraphics[width=0.095\textwidth]{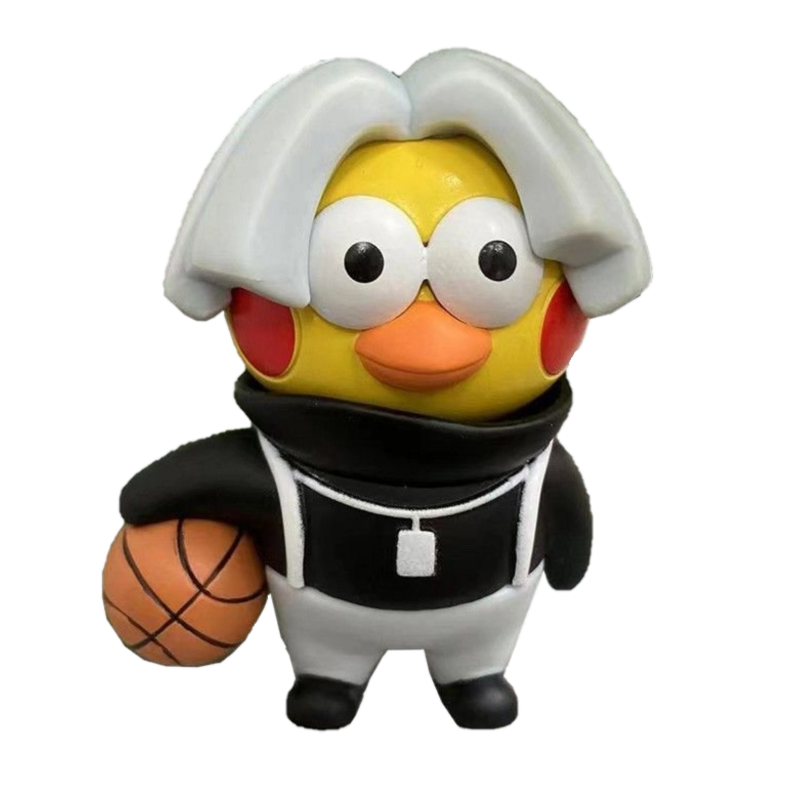} &
        \includegraphics[width=0.095\textwidth]{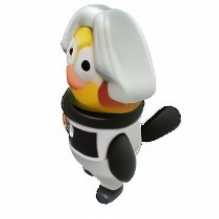} &
        \includegraphics[width=0.095\textwidth]{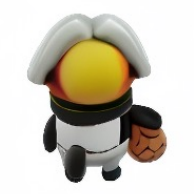} &
        \includegraphics[width=0.095\textwidth]{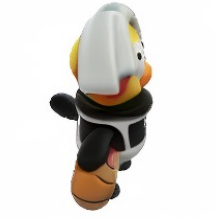} &
        \includegraphics[width=0.095\textwidth]{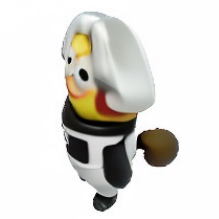} &
        \includegraphics[width=0.095\textwidth]{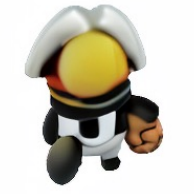} &
        \includegraphics[width=0.095\textwidth]{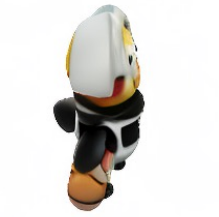} &
        \includegraphics[width=0.095\textwidth]{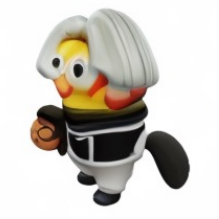} &
        \includegraphics[width=0.095\textwidth]{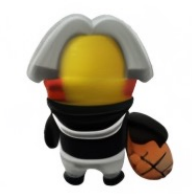} &
        \includegraphics[width=0.095\textwidth]{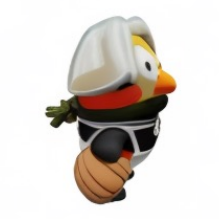} \\
         &
        \includegraphics[width=0.095\textwidth]{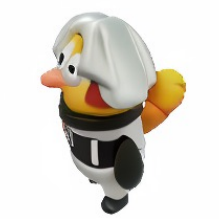} &
        \includegraphics[width=0.095\textwidth]{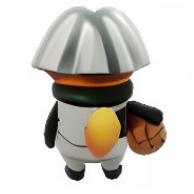} &
        \includegraphics[width=0.095\textwidth]{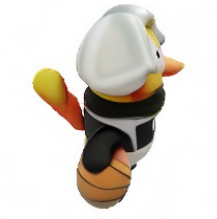} &
        \includegraphics[width=0.095\textwidth]{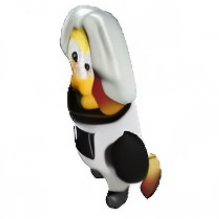} &
        \includegraphics[width=0.095\textwidth]{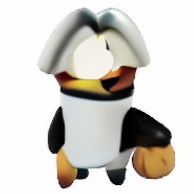} &
        \includegraphics[width=0.095\textwidth]{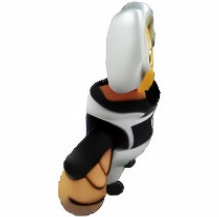} &
        \includegraphics[width=0.095\textwidth]{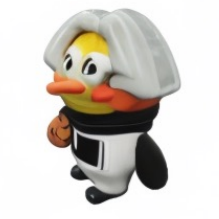} &
        \includegraphics[width=0.095\textwidth]{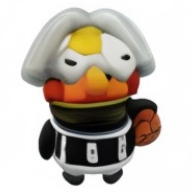} &
        \includegraphics[width=0.095\textwidth]{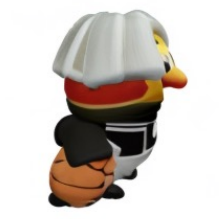} \\
        \multicolumn{1}{c}{Input} & \multicolumn{3}{c}{HarmonyView} & \multicolumn{3}{c}{SyncDreamer~\cite{liu2023syncdreamer}} & \multicolumn{3}{c}{Zero123~\cite{liu2023zero}} \\
    \end{tabular}
    }
    \vspace{\abovefigcapmargin}
    \caption{
        \textbf{Novel-view synthesis comparison.}
        HarmonyView generates plausible novel views while preserving coherence across views.
    }
    \vspace{\belowfigcapmargin}
    \label{fig:nvs}
\end{figure*}

%% file: contents/04_experiments.tex
\input{tables/novel_view_synthesis}

\section{Experiments}
\label{sec:experiments}
Due to space constraints, we provide detailed information regarding implementation details and baselines in Appendix.

\noindent\textbf{Dataset.}
Following~\cite{liu2023zero,liu2023one,liu2023syncdreamer}, we used the Google Scanned Object (GSO)~\cite{downs2022google} dataset, adopting the same data split as in~\cite{liu2023syncdreamer}, for our evaluation.
In addition, we utilized Internet-collected images, including those curated by~\cite{liu2023syncdreamer}, to assess the generation ability for complex objects or scenes.

\noindent\textbf{Tasks and metrics.}
For the novel-view synthesis task, we used three standard metrics -- PSNR, SSIM~\cite{wang2004image}, LPIPS~\cite{zhang2018unreasonable} -- to measure sample quality compared to GT images.
We measured diversity using the CD score.
As a multi-view coherency metric, we propose $E_{flow}$, which measures the $\ell_{1}$ distance between optical flow estimates from RAFT~\cite{teed2020raft} for both GT and generated images.
For the single-view 3D reconstruction task, we used Chamfer distance to evaluate point-by-point shape similarity and volumetric IoU to quantify the overlap between reconstructed and GT shapes.

\input{tables/testset_nvs}
\input{figs/qualitative_3d_recon}

\subsection{Comparative Results}
\paragraph{Novel-view synthesis.}
\Cref{tab:nvs} shows the quantitative results for novel-view synthesis on the GSO~\cite{downs2022google} dataset.
Here, HarmonyView outperforms state-of-the-art methods across all metrics.
We confirm that HarmonyView generates images of superior quality, as indicated by PSNR, SSIM and LPIPS.
It particularly excels in achieving multi-view coherency (indicated by $E_{flow}$) and generating diverse views that are faithful to the semantics of the input view (indicated by CD score).
%
In~\cref{fig:nvs}, we present the qualitative results.
Zero123~\cite{liu2023zero} produces multi-view incoherent images or implausible images, \eg, eyes on the back.
SyncDreamer~\cite{liu2023syncdreamer} generates images that lack visual similarity to the input view or contain deficiencies, \eg, flatness or hole on the back.
In contrast, HarmonyView generates diverse yet plausible multi-view images while maintaining geometric coherence across views.
In \Cref{tab:testset_nvs}, we examine novel-view synthesis methods on in-the-wild images curated by~\cite{liu2023syncdreamer}.
For evaluation, we use CD score and user Likert ratings (1 to 5) along three criteria: quality, consistency, and diversity.
While SyncDreamer~\cite{liu2023syncdreamer} excels in quality and consistency scores when compared to Zero123~\cite{liu2023zero}, Zero123 performs better in diversity and CD score.
Notably, HarmonyView stands out with the highest CD score and superior user ratings.
This suggests that HarmonyView effectively produces visually pleasing, realistic, and diverse images while being coherent across multiple views.
The correlation between the CD score and the diversity score underscores the efficacy of the CD score in capturing the diversity of generated images.

\input{tables/3d_reconstruction}

\vspace{\paramargin}
\paragraph{3D reconstruction.}
In~\Cref{tab:recon}, we quantitatively compare our approach against various other 3D generation methods~\cite{liu2023syncdreamer,liu2023zero,liu2023one,nichol2022point,jun2023shap,qian2023magic123,melas2023realfusion}.
Both our method and SDS-free methods~\cite{liu2023zero,liu2023syncdreamer} utilize NeuS~\cite{wang2021neus}, a neural reconstruction method for converting multi-view images into 3D shapes.
To achieve faithful reconstruction of 3D mesh that aligns well with ground truth, the generated multi-view images should be geometrically coherent.
Notably, HarmonyView achieves the best results by a significant margin in both Chamfer distance and volumetric IoU metrics, demonstrating the proficiency of HarmonyView in producing multi-view coherent images.
We also present a qualitative comparison in~\cref{fig:recon}.
The results showcase the remarkable quality of HarmonyView.
While competing methods often struggle with incomplete reconstructions (\eg, Point-E, Shap-E), fall short in capturing small details (\eg, Zero123), and show discontinuities (\eg, SyncDreamer) or artifacts (\eg, One-2-3-45), our method produces high-quality 3D meshes characterized by accurate geometry and a realistic appearance.

\input{tables/scale_study}

\subsection{Analysis}
\paragraph{Scale study.}
In~\Cref{tab:scale_study}, we investigate two instantiations of multi-view diffusion guidance with different scale configurations: baseline (\cref{eq:mv_cfg}) and our approach (\cref{eq:ours_final}).
As $s$ increases from 0.5 to 1.5 in the baseline method, $E_{flow}$ (indicating \textit{multi-view coherency}) and CD score (indicating \textit{diversity}) show an increasing trend.
Simultaneously, PSNR, SSIM, and LPIPS (indicating \textit{visual consistency}) show a declining trend.
This implies a trade-off between visual consistency and diversity.
In contrast, our method involves parameters $s_1$ and $s_2$.
We observe that increasing $s_1$ provides stronger guidance in aligning multi-view images with the input view, leading to direct improvements in PSNR, SSIM, and LPIPS.
Keeping $s_1$ fixed at 2.0, elevating $s_2$ tends to yield improved CD score, indicating an enhanced diversity in the generated images.
However, given the inherent conflict between consistency and diversity, an increase in $s_2$ introduces a trade-off.
We note that our approach consistently outperforms the baseline across various configurations, striking a nuanced balance between consistency and diversity.
Essentially, our decomposition provides more explicit control over those two dimensions, enabling a better balance.
Additionally, the synergy between $s_1$ and $s_2$ notably enhances $E_{flow}$, leading to improved 3D alignment across multiple views.


\input{figs/qualitative_complex_scene}
\input{figs/qualitative_text_to_3d}

\vspace{\paramargin}
\paragraph{Generalization to complex objects or scenes.}
Even in challenging scenarios, either with a highly detailed single object or multiple objects within a single scene, HarmonyView excels at capturing intricate details that SyncDreamer~\cite{liu2023syncdreamer} might miss.
The results are shown in~\cref{fig:complex}.
Our model well generates multi-view coherent images even in such scenarios, enabling the smooth reconstruction of natural-looking meshes without any discontinuities.


\vspace{\paramargin}
\paragraph{Compatibility with text-to-image models.}
HarmonyView seamlessly integrates with off-the-shelf text-to-image models~\cite{ramesh2021zero,rombach2022high}.
These models convert textual descriptions into 2D images, which our model further transforms into high-quality multi-view images and 3D meshes.
Visual examples are shown in~\cref{fig:t23d}.
Notably, our model excels in capturing the essence or mood of the given 2D image, even managing to create plausible details for occluded parts.
This demonstrates strong generalization capability, allowing it to perform well even with unstructured real-world images.


\vspace{\paramargin}
\paragraph{Runtime.}
HarmonyView generates 64 images (\ie, 4 instances $\times$ 16 views) in only one minute, with 50 DDIM~\cite{song2020denoising} sampling steps on an 80GB A100 GPU.
Despite the additional forward pass through the diffusion model, HarmonyView takes less runtime than SyncDreamer~\cite{liu2023syncdreamer}, which requires about 2.7 minutes with 200 DDIM sampling steps.

\vspace{\paramargin}
\paragraph{Additional results \& analysis.}
Please see Appendix for more qualitative examples and analysis on the CD score, \etc.

%% file: tables/novel_view_synthesis.tex
\begin{table}[]
    \centering
    \resizebox{\linewidth}{!}{
    \begin{tabular}{lccccc}
       \toprule
       Method  & PSNR$\uparrow$ & SSIM$\uparrow$ & LPIPS$\downarrow$ & $E_{flow}$$\downarrow$ & CD$\uparrow$ \\
       \midrule
       Realfusion~\cite{melas2023realfusion}    
       & 15.26 & 0.722 & 0.283 &  - & -  \\
       Zero123~\cite{liu2023zero}    
       & 18.98 & 0.795 & 0.166 &  3.820 & 0.628  \\
       SyncDreamer~\cite{liu2023syncdreamer}   
       & 20.19 & 0.819 & 0.140 & 2.071 & 0.717   \\
       \rowcolor{gray!25}
       HarmonyView & \textbf{20.69} & \textbf{0.825} & \textbf{0.133} & \textbf{1.945} & \textbf{0.792} \\
       \bottomrule
    \end{tabular}
    }
    \vspace{\abovetabcapmargin}
    \caption{
        \textbf{Novel-view synthesis on GSO~\cite{downs2022google} dataset.}
        We report PSNR, SSIM, LPIPS, $E_{flow}$, and CD score.
    }
    \vspace{\belowtabcapmargin}
    \label{tab:nvs}
\end{table}

%% file: tables/testset_nvs.tex
\begin{table}[]
    \centering
    \resizebox{\linewidth}{!}{
    \begin{tabular}{lcccc}
       \toprule
       \multirow{2}{*}{Methods} & \multirow{2}{*}{CD$\uparrow$} & \multicolumn{3}{c}{User Likert Score (1-5)$\uparrow$} \\
       \arrayrulecolor{gray}\cmidrule(lr){3-5}
       & & Quality & Consistency & Diversity \\
       \arrayrulecolor{black}\midrule
       Zero123~\cite{liu2023zero} & 0.752 & 3.208 & 3.167 & 2.854 \\
       SyncDreamer~\cite{liu2023syncdreamer} & 0.722 & 3.417 & 3.208 & 2.708 \\
       \rowcolor{gray!25}
       HarmonyView & \textbf{0.804} & \textbf{3.958} & \textbf{3.479} & \textbf{3.813} \\
       \bottomrule
    \end{tabular}
    }
    \vspace{\abovetabcapmargin}
    \caption{
        \textbf{Novel-view synthesis on in-the-wild images.}
        We report the CD score and 5-scale user Likert score, assessing quality, consistency, and diversity.
        Notably, the CD score shows strong alignment with human judgments.
        The test images are collected by~\cite{liu2023syncdreamer}.
    }
    \vspace{\belowtabcapmargin}
    \vspace{-1mm}
    \label{tab:testset_nvs}
\end{table}

%% file: figs/qualitative_3d_recon.tex
\begin{figure*}
    \centering
    \setlength\tabcolsep{5pt}
    \resizebox{\linewidth}{!}{
    \renewcommand{\arraystretch}{0.5}
    \begin{tabular}{ccccccc}
        \includegraphics[width=0.13\textwidth]{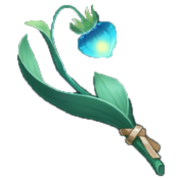} &
        \includegraphics[width=0.13\textwidth]{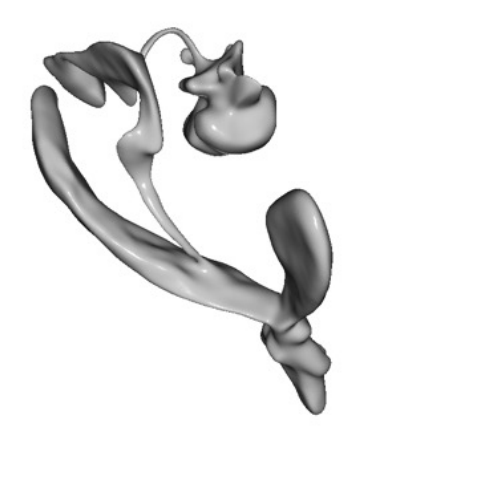} &
        \includegraphics[width=0.13\textwidth]{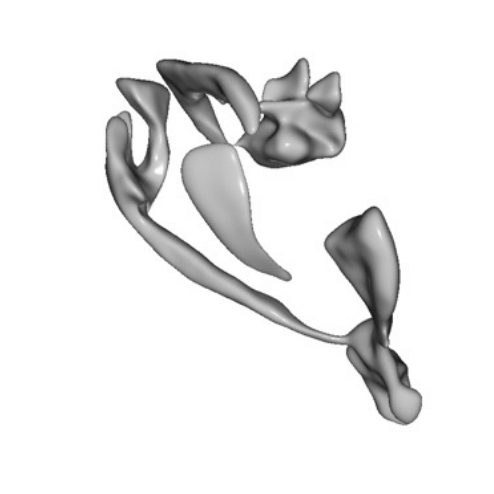} &
        \includegraphics[width=0.13\textwidth]{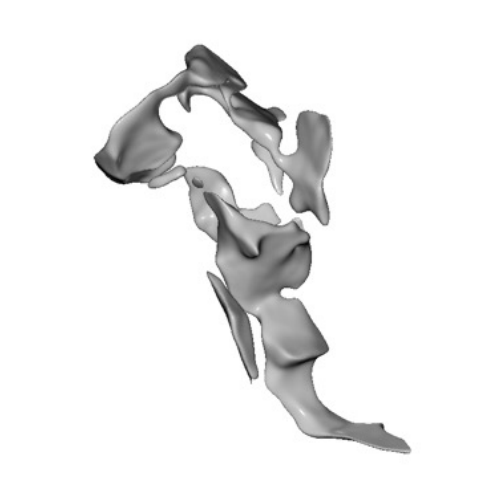} &
        \includegraphics[width=0.13\textwidth]{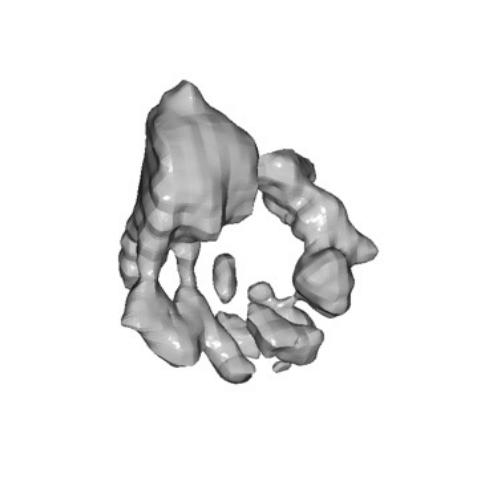} &
        \includegraphics[width=0.13\textwidth]{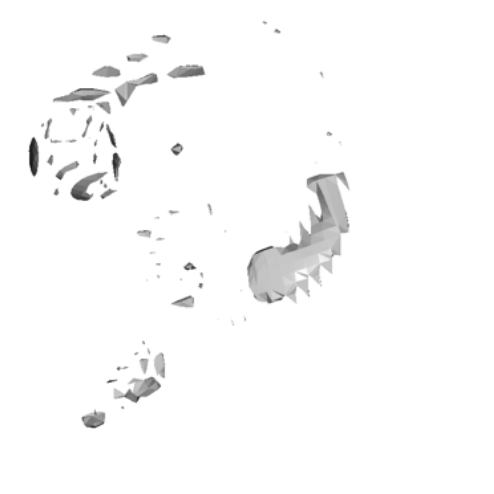} &
        \includegraphics[width=0.13\textwidth]{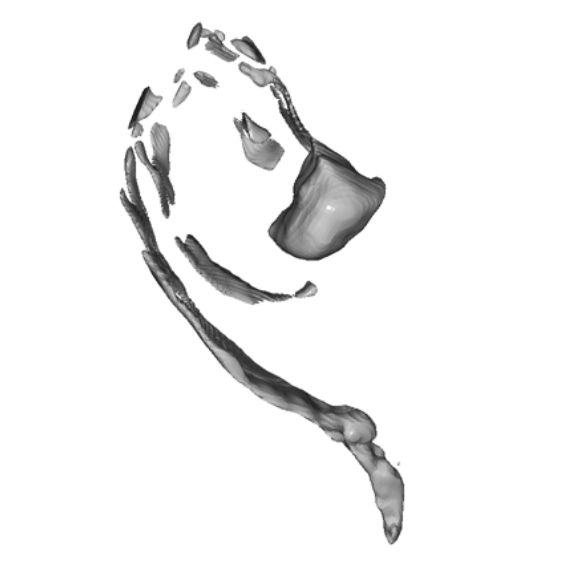} \\
        \includegraphics[width=0.13\textwidth]{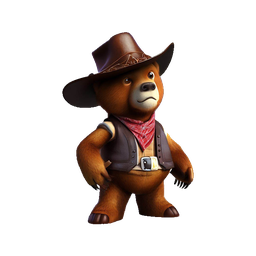} &
        \includegraphics[width=0.13\textwidth]{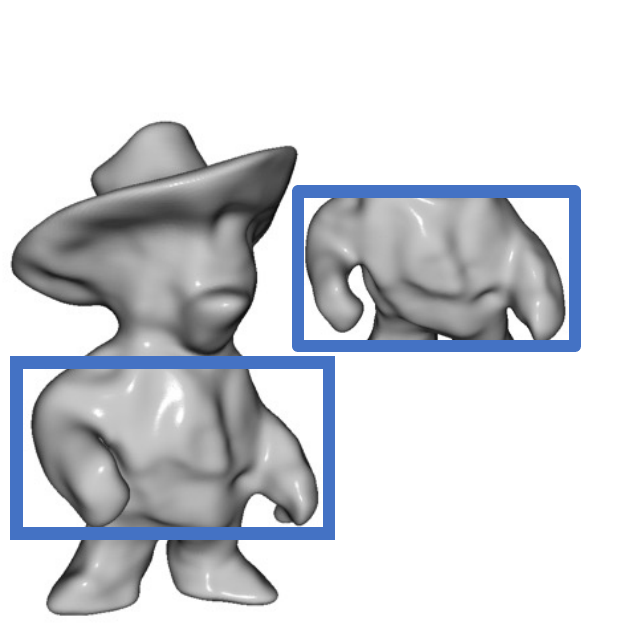} &
        \includegraphics[width=0.13\textwidth]{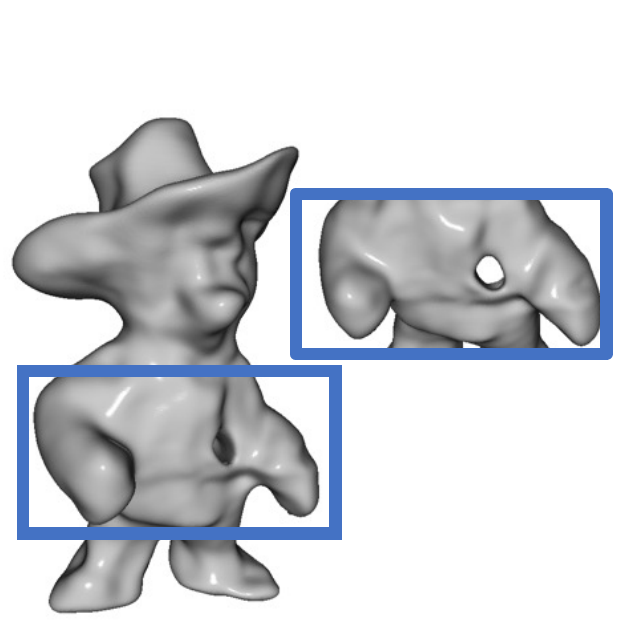} &
        \includegraphics[width=0.13\textwidth]{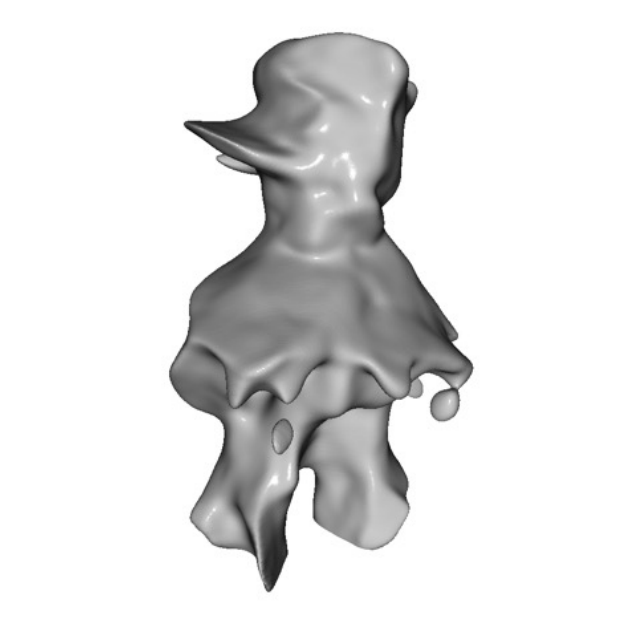} &
        \includegraphics[width=0.13\textwidth]{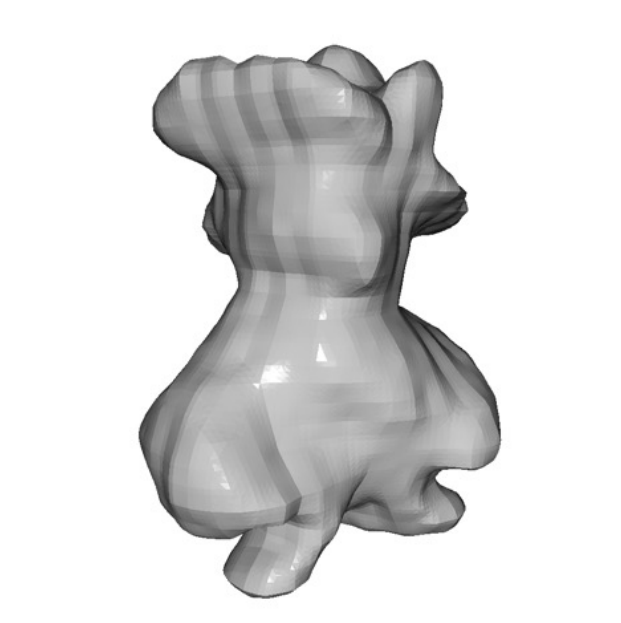} &
        \includegraphics[width=0.13\textwidth]{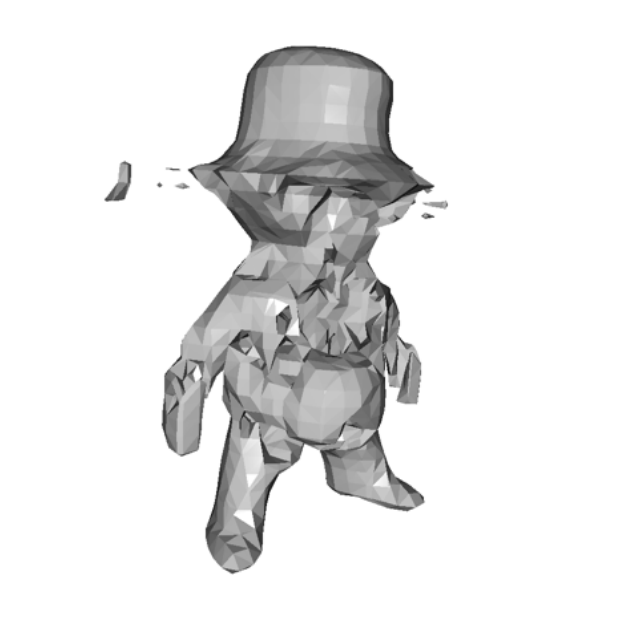} &
        \includegraphics[width=0.13\textwidth]{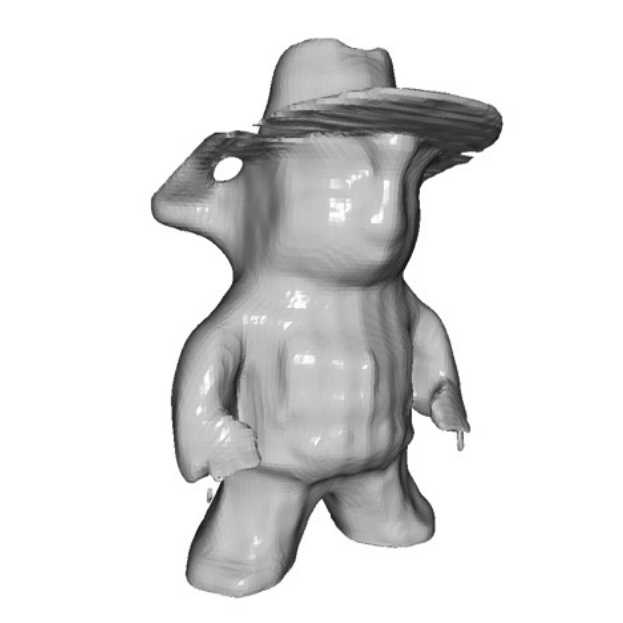} \\
        \includegraphics[width=0.13\textwidth]{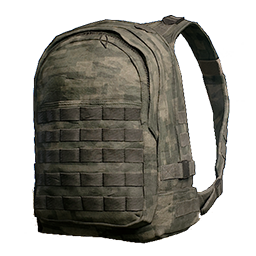} &
        \includegraphics[width=0.13\textwidth]{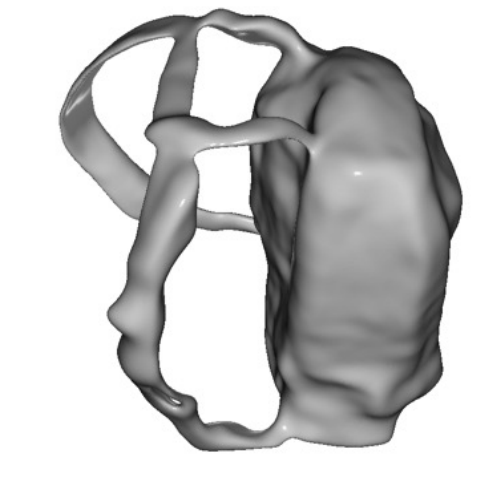} &
        \includegraphics[width=0.13\textwidth]{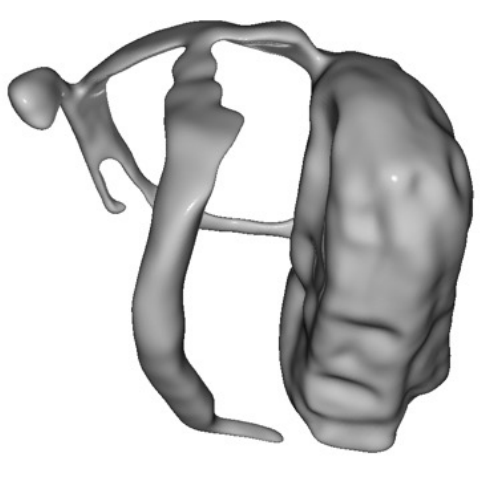} &
        \includegraphics[width=0.13\textwidth]{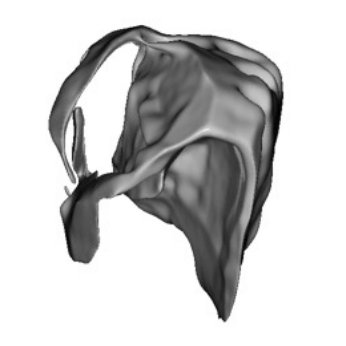} &
        \includegraphics[width=0.13\textwidth]{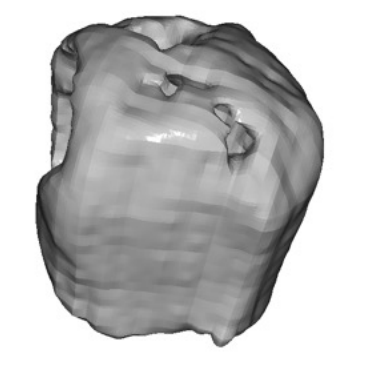} &
        \includegraphics[width=0.13\textwidth]{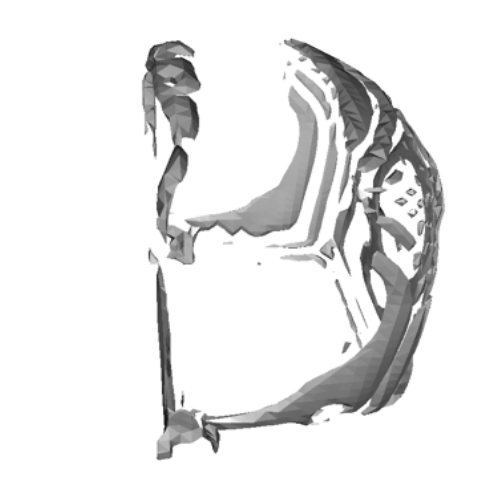} &
        \includegraphics[width=0.13\textwidth]{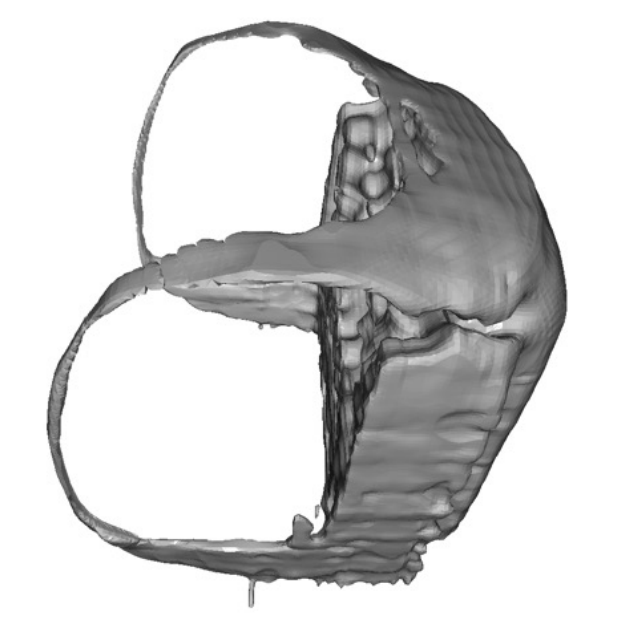} \\
        Input & HarmonyView & {\small SyncDreamer~\cite{liu2023syncdreamer}} & Zero123~\cite{liu2023zero} & One-2-3-45~\cite{liu2023one} & Point-E~\cite{nichol2022point}& Shap-E~\cite{jun2023shap} \\
    \end{tabular}
    }
    \vspace{\abovefigcapmargin}
    \caption{
        \textbf{3D reconstruction comparison.}
        HarmonyView stands out in creating high-quality 3D meshes where other often fails.
        HarmonyView, SyncDreamer~\cite{liu2023syncdreamer}, and Zero123~\cite{liu2023zero} use the vanilla NeuS~\cite{wang2021neus} for 3D reconstruction.
    }
    \vspace{\belowfigcapmargin}
    \label{fig:recon}
\end{figure*}

%% file: tables/3d_reconstruction.tex
\begin{table}[]
    \centering
    \resizebox{0.8\linewidth}{!}{
    \begin{tabular}{lcc}
       \toprule
       Method  & Chamfer Dist.$\downarrow$ & Volume IoU$\uparrow$ \\
       \midrule
       Realfusion~\cite{melas2023realfusion}    
       & 0.0819  & 0.2741   \\
       Magic123~\cite{qian2023magic123}
       & 0.0516 &  0.4528 \\
       One-2-3-45~\cite{liu2023one}    
       & 0.0629 &  0.4086 \\
       Point-E~\cite{nichol2022point}    
       & 0.0426 & 0.2875 \\
       Shap-E~\cite{jun2023shap}    
       & 0.0436 &  0.3584  \\
       Zero123~\cite{liu2023zero}    
       & 0.0339 &  0.5035 \\
       SyncDreamer~\cite{liu2023syncdreamer}    
       &  0.0261  &  0.5421   \\
       \rowcolor{gray!25}
       HarmonyView & \textbf{0.0187} & \textbf{0.6401} \\
       \bottomrule
    \end{tabular}
    }
    \vspace{\abovetabcapmargin}
    \caption{
        \textbf{3D reconstruction on GSO~\cite{downs2022google} dataset.} HarmonyView demonstrates substantial improvements over competitive baselines. 
    }
    \vspace{\belowtabcapmargin}
    \label{tab:recon}
\end{table}

%% file: tables/scale_study.tex
\begin{table}[]
    \centering
    \setlength\tabcolsep{3pt}
    \resizebox{\linewidth}{!}{
    \begin{tabular}{lcccccccc}
       \toprule
       Method & {\large $s$} & {\large $s_1$} & {\large $s_2$} & PSNR$\uparrow$ & SSIM$\uparrow$ & LPIPS$\downarrow$ & $E_{flow}$$\downarrow$ & CD$\uparrow$ \\
       \midrule
       \multirow{3}{*}{Baseline (\cref{eq:mv_cfg})} & 0.5 & - & - & 20.55 & 0.822 & 0.137 & 2.074 & 0.685   \\
       & 1.0 & - & - & 20.19 & 0.819 & 0.140 & 2.071 & 0.717   \\
       & 1.5 & - & - & 19.76 & 0.814 & 0.146 & 2.011 & 0.711   \\
       \arrayrulecolor{gray}\cmidrule(lr){1-9}
       \multirow{8}{*}{Ours (\cref{eq:ours_final})} & - & 0.0 & 1.0 & 20.32 & 0.822 & 0.141 & 2.136 & 0.764 \\
       & - & 1.0 & 1.0 & 20.55 & 0.824 & 0.135 & 2.009 & 0.772 \\
       & - & 3.0 & 1.0 & 20.73 & 0.825 & 0.132 & 1.950 & 0.737 \\
       & - & 2.0 & 0.0 & \textbf{21.03} & \textbf{0.828} & \textbf{0.128} & 2.146 & 0.668 \\
       & - & 2.0 & 0.6 & 20.90 & 0.827 & 0.130 & 1.996 & 0.770 \\
       & - & 2.0 & 0.8 & 20.80 & 0.826 & 0.131 & 2.009 & 0.774 \\
       & - & 2.0 & 1.2 & 20.56 & 0.824 & 0.135 & 1.996 & 0.760 \\
       & - & \cellcolor{gray!25}\textbf{2.0} & \cellcolor{gray!25}\textbf{1.0} & \cellcolor{gray!25}20.69 & \cellcolor{gray!25}0.825 & \cellcolor{gray!25}0.133 & \cellcolor{gray!25}\textbf{1.945} & \cellcolor{gray!25}\textbf{0.792} \\
       \arrayrulecolor{black}\bottomrule
    \end{tabular}
    }
    \vspace{\abovetabcapmargin}
    \caption{
        \textbf{Guidance scale study on novel-view synthesis.}
        We compare two instantiations of multi-view diffusion guidance: \cref{eq:mv_cfg} and \cref{eq:ours_final}.
        Our approach consistently outperforms the baseline.
        Increasing $s_1$ tends to enhance PSNR, SSIM, and LPIPS, while higher $s_2$ tends to improve CD score.
        Notably, the combined effect of $s_1$ and $s_2$ synergistically improves $E_{flow}$.
    }
    \vspace{\belowtabcapmargin}
    \label{tab:scale_study}
\end{table}

%% file: figs/qualitative_complex_scene.tex
\begin{figure*}
    \centering
    \setlength\tabcolsep{1pt}
    \resizebox{\linewidth}{!}{
    \renewcommand{\arraystretch}{0.5}
    \begin{tabular}{c:cccc:cccc}
    \includegraphics[width=0.105\textwidth]{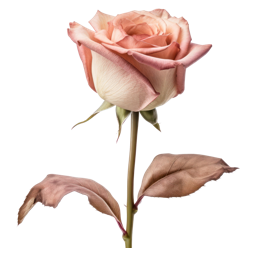} &
    \includegraphics[width=0.105\textwidth]{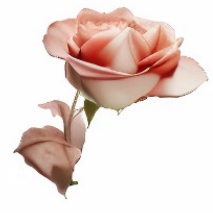} &
    \includegraphics[width=0.105\textwidth]{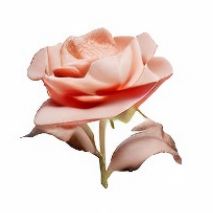} &
    \includegraphics[width=0.105\textwidth]{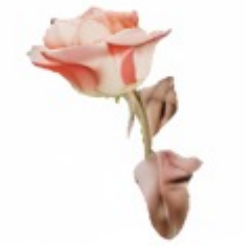} &
    \includegraphics[width=0.105\textwidth]{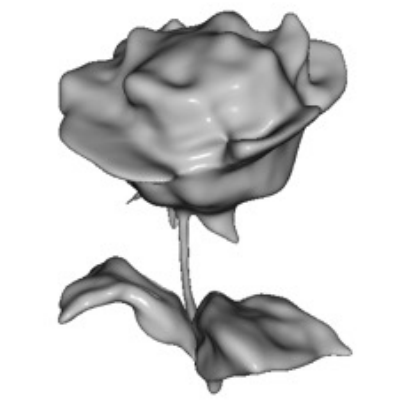} &
    \includegraphics[width=0.105\textwidth]{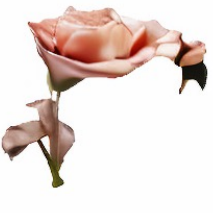} &
    \includegraphics[width=0.105\textwidth]{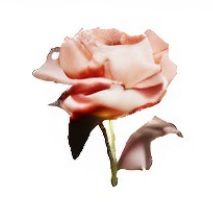} &
    \includegraphics[width=0.105\textwidth]{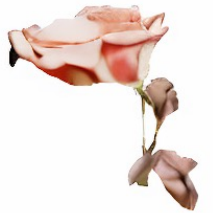} &
    \includegraphics[width=0.105\textwidth]{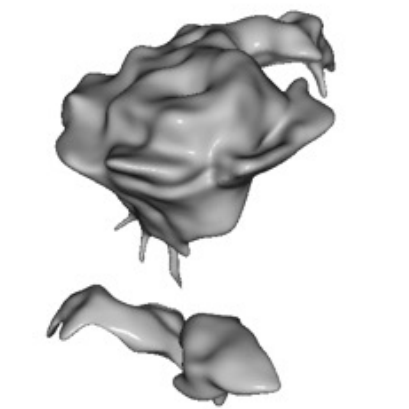} \\
    \includegraphics[width=0.105\textwidth]{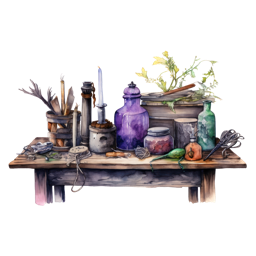} &
    \includegraphics[width=0.105\textwidth]{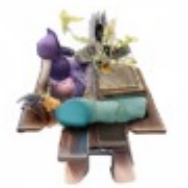} &
    \includegraphics[width=0.105\textwidth]{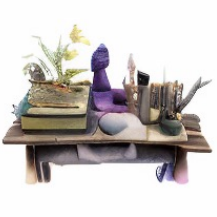} &
    \includegraphics[width=0.105\textwidth]{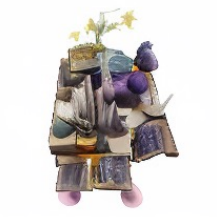} &
    \includegraphics[width=0.105\textwidth]{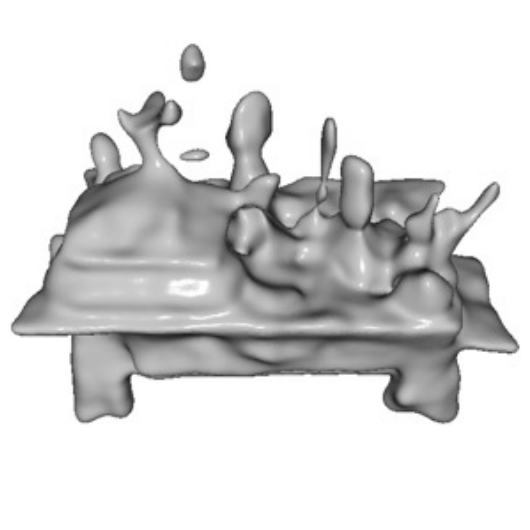} &
    \includegraphics[width=0.105\textwidth]{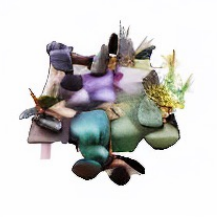} &
    \includegraphics[width=0.105\textwidth]{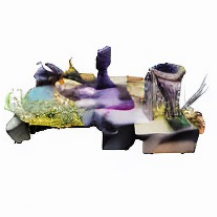} &
    \includegraphics[width=0.105\textwidth]{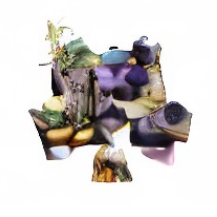} &
    \includegraphics[width=0.105\textwidth]{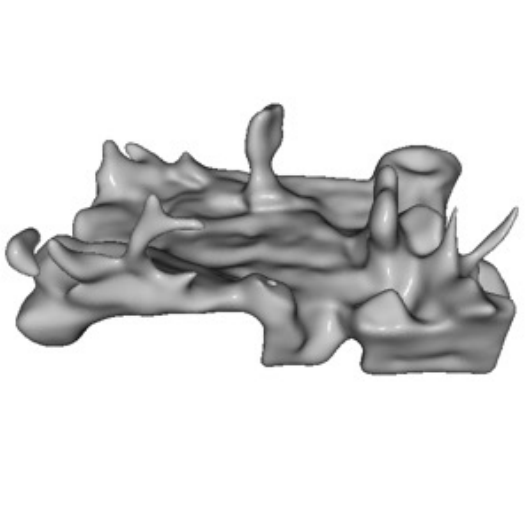} \\
    \multicolumn{1}{c}{Input} & \multicolumn{4}{c}{HarmonyView} & \multicolumn{4}{c}{SyncDreamer~\cite{liu2023syncdreamer}}
    \end{tabular}
    }
    \vspace{\abovefigcapmargin}
    \caption{
        \textbf{3D reconstruction for complex object or scene.}
        HarmonyView successfully reconstructs the details, while SyncDreamer fails.
    }
    \label{fig:complex}
\end{figure*}

%% file: figs/qualitative_text_to_3d.tex
\begin{figure*}
    \centering
    \setlength\tabcolsep{1pt}
    \resizebox{\linewidth}{!}{
    \renewcommand{\arraystretch}{0.5}
    \begin{tabular}{cccccccc}
    \includegraphics[width=0.12\textwidth]{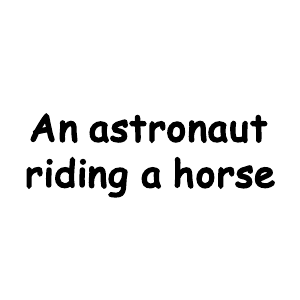} &
    \includegraphics[width=0.12\textwidth]{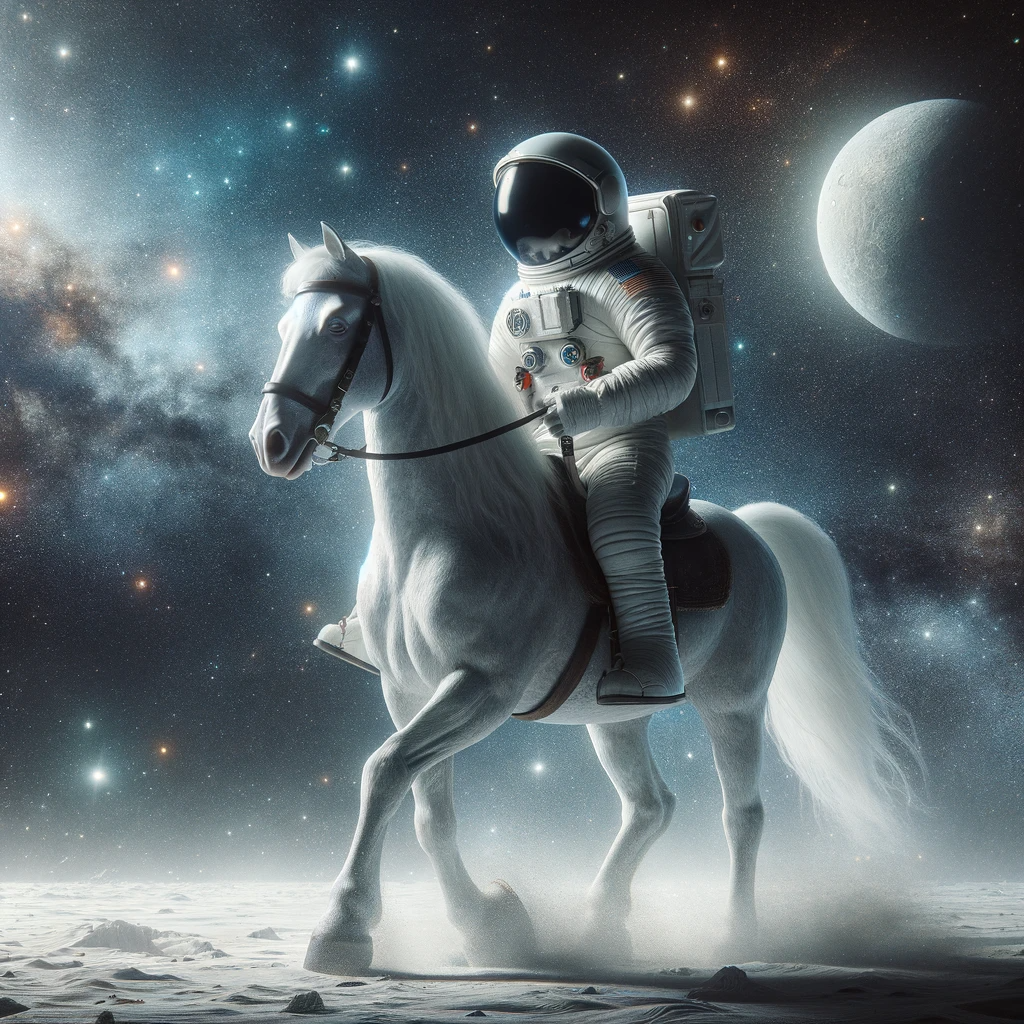} &
    \includegraphics[width=0.12\textwidth]{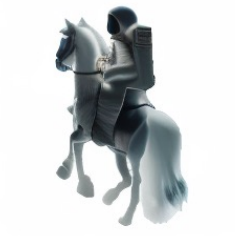} &
    \includegraphics[width=0.12\textwidth]{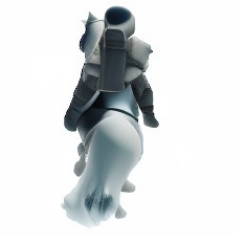} &
    \includegraphics[width=0.12\textwidth]{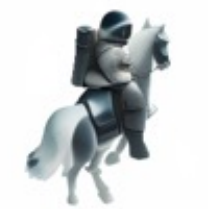} &
    \includegraphics[width=0.12\textwidth]{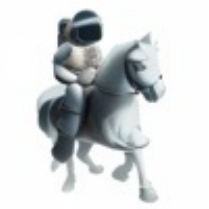} &
    \includegraphics[width=0.12\textwidth]{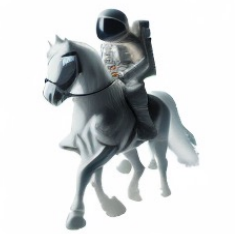} &
    \includegraphics[width=0.12\textwidth]{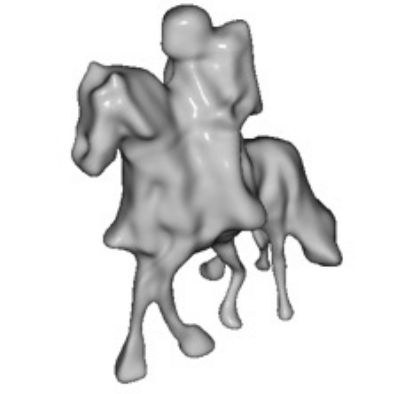} \\
    \includegraphics[width=0.12\textwidth]{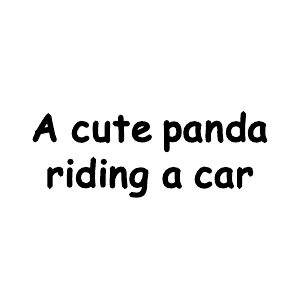} &
    \includegraphics[width=0.12\textwidth]{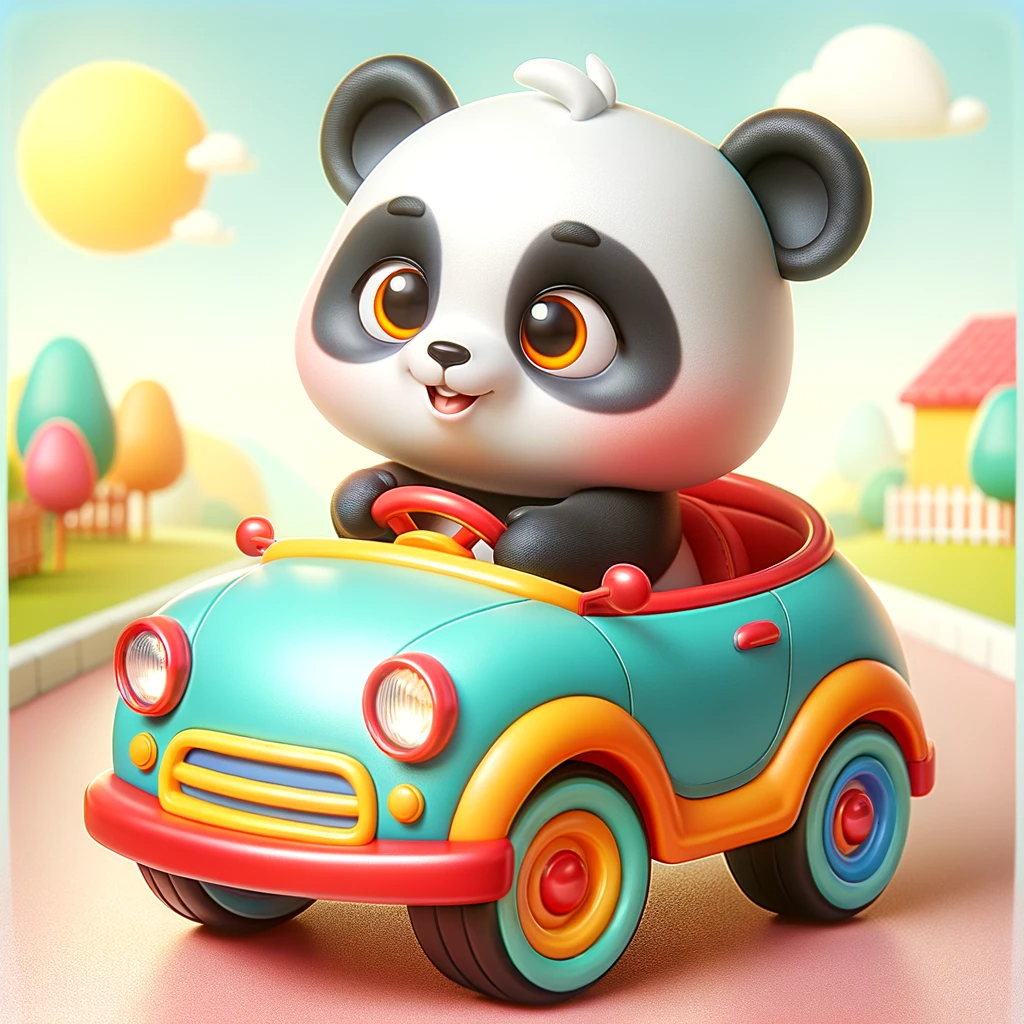} &
    \includegraphics[width=0.12\textwidth]{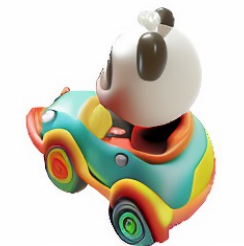} &
    \includegraphics[width=0.12\textwidth]{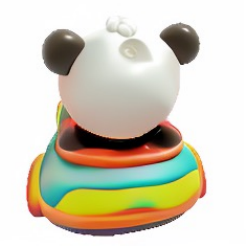} &
    \includegraphics[width=0.12\textwidth]{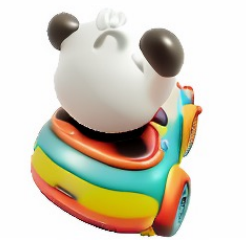} &
    \includegraphics[width=0.12\textwidth]{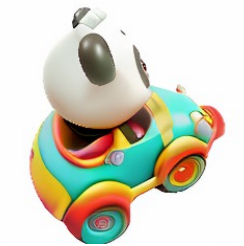} &
    \includegraphics[width=0.12\textwidth]{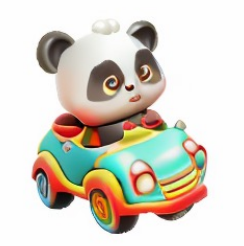} &
    \includegraphics[width=0.12\textwidth]{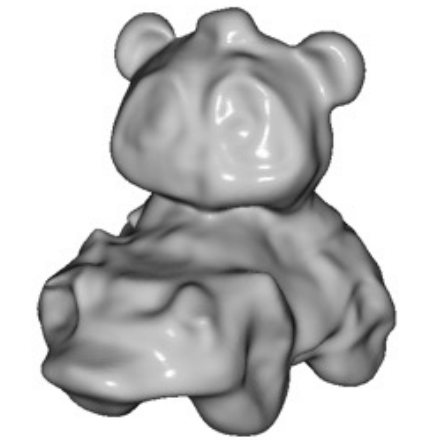} \\
    \includegraphics[width=0.12\textwidth]{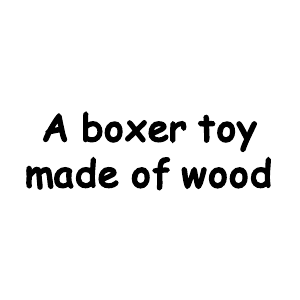} &
    \includegraphics[width=0.12\textwidth]{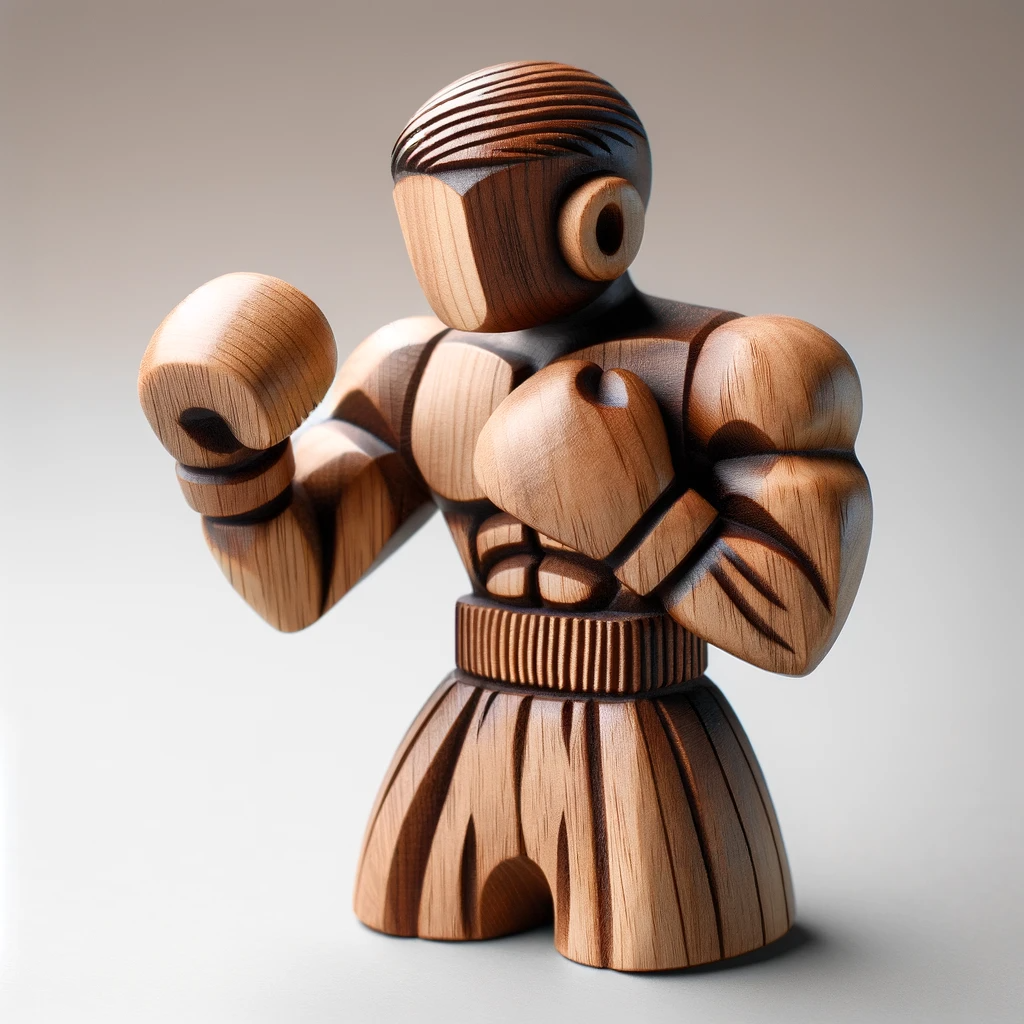} &
    \includegraphics[width=0.12\textwidth]{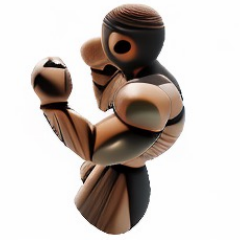} &
    \includegraphics[width=0.12\textwidth]{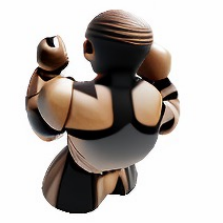} &
    \includegraphics[width=0.12\textwidth]{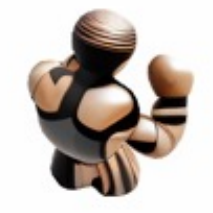} &
    \includegraphics[width=0.12\textwidth]{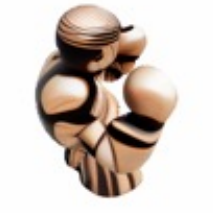} &
    \includegraphics[width=0.12\textwidth]{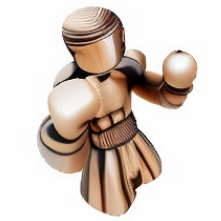} &
    \includegraphics[width=0.12\textwidth]{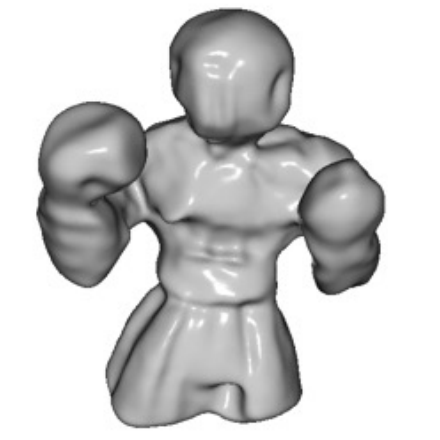} \\
    Input text & Text to image & \multicolumn{5}{c}{Generated images} & Mesh \\
    \end{tabular}
    }
    \vspace{\abovefigcapmargin}
    \caption{
        \textbf{Text-to-Image-to-3D.}
        HarmonyView, when combined with text-to-image frameworks~\cite{ramesh2021zero,nichol2021glide,rombach2022high}, enables text-to-3D.
    }
    \vspace{\belowfigcapmargin}
    \label{fig:t23d}
\end{figure*}


%% file: contents/10_conclusion.tex
\section{Conclusion}
\label{sec:conclusion}
In this study, we have introduced HarmonyView, a simple yet effective technique that adeptly balances two fundamental aspects in a single-image 3D generation: \textit{consistency} and \textit{diversity}.
By providing explicit control over the diffusion sampling process, HarmonyView achieves a harmonious equilibrium, facilitating the generation of diverse yet plausible novel views while enhancing consistency.
Our proposed evaluation metric CD score effectively measures the diversity of generated multi-views, closely aligning with human evaluators' judgments.
Experiments show the superiority of HarmonyView over state-of-the-art methods in both novel-view synthesis and 3D reconstruction tasks.
The visual fidelity and faithful reconstructions achieved by HarmonyView highlight its efficacy and potential for various applications.

%% file: contents/12_appendix.tex

\appendix
\label{sec:appendix}





\section{More Related Work}
The challenge of \textit{one-image 3D generation} has recently attracted significant attention, with various approaches and methods proposed to address this complex problem~\cite{chen2023review}.
In this section, we provide a brief review of the literature.

\vspace{\paramargin}
\paragraph{Classical 3D generative methods.}
Early works can be broadly categorized into two main groups: primitive-based approaches and depth estimation approaches.
Primitive-based approaches~\cite{tulsiani2016learning,wang2018pixel2mesh,groueix2018papier}, focus on the fitting of primitive 3D shapes to 2D images, seeking to align synthetic models with observed image features.
They often employ iterative optimization to refine the pose and shape of the model until a satisfactory fit is achieved.
On the other hand, depth estimation approaches~\cite{zhang2018learning,wu2020unsupervised} typically follow a two-step process:
They first use a monocular depth estimator (\eg, MiDaS~\cite{ranftl2020towards}) to predict the 3D geometry, which is then used to render artistic effects through multi-plane images~\citep{jampani2021slide,shih20203d} or point clouds~\cite{mu20223d}.
To address imperfections, a pre-trained inpainting model~\cite{yu2019free} is often applied to fill in missing holes.
However, these early approaches may struggle with generalization to real-world data or new object categories.


\vspace{\paramargin}
\paragraph{3D native models.}
A line of research~\cite{choy20163dr2d2,sinha2017surfnet,wu2018learning,groueix2018papier,chen2020bsp,deng2020cvxnet,genova2020local} follows an encoder-decoder framework for modeling the image-to-3D data distribution, which involves the use of global shape latent codes to directly encode the shape information from 3D assets (\eg, ShapeNet~\cite{chang2015shapenet}, Pix3D~\cite{sun2018pix3d}).
In contrast, other works utilize local features and representation-specific 3D generative models that leverage priors constructed from 3D primitives in various formats: point clouds~\cite{fan2017point,groueix2018papier,wu2020pq,zeng2022lion,melas2023pc2}, voxels~\cite{choy20163dr2d2,tulsiani2017drc,chen2019imnet,cheng2023sdfusion}, meshes~\cite{liao2018deep,wang2018pixel2mesh,gkioxari2019meshrcnn,chen2021neural,liu2023meshdiffusion}, or parametric surfaces~\cite{sharma2020parsenet,gupta20233dgen}.
While these \textit{3D native} models show impressive performance, they often require extensive 3D data and are constrained to specific object classes within that data.
They also suffer from quality degradation when handling real-world images due to domain disparities.
Recently, Point-E~\cite{nichol2022point} and Shap-E~\cite{jun2023shap} propose learning text-to-3D diffusion models on large-scale 3D assets to mitigate some of these limitations.








\section{Additional Experimental Setup}
\paragraph{Diversity evaluation.}
Due to the inherent stochastic nature of diffusion models, the outputs they generate can be different \wrt the random seed used for their generation.
Therefore, the computed metrics can differ depending on the seed we use.
To evaluate the diversity of generated samples from each model, we randomly sample 4 instances using different random seeds from the same input image.
We then use the CD score to quantify the diversity.
By calculating the CD score across these sampled instances (each derived from a different random seed but originating from the sample input images), we obtain an average CD score.
This average CD score represents the overall dissimilarity or diversity observed among the generated samples.
The reported values in the main paper are the average CD score calculated across these sampled instances.

\vspace{\paramargin}
\paragraph{Technical details.}
HarmonyView is built upon the pre-trained models of SyncDreamer~\cite{liu2023syncdreamer}, which generates a set of $N=16$ multi-view images, each with an elevation of $30^\circ$ and azimuths evenly distributed in the range of $[0^\circ, 360^\circ]$.
We assume that the azimuth of both the input view and the first target view is set to $0^\circ$.
The viewpoint differences $\Delta{\rvv}^{(n)}$ are calculated based on the differences in elevation and azimuth between the input view and target view.
At test time, similar to~\cite{melas2023realfusion,liu2023zero,qian2023magic123,liu2023syncdreamer}, we estimate an elevation angle and use it as an input.
To reconstruct the 3D mesh, we use foreground masks for generated images using CarveKit\footnote{\href{https://github.com/OPHoperHPO/image-background-remove-tool}{https://github.com/OPHoperHPO/image-background-remove-tool}}, and train the NeuS~\cite{wang2021neus} for 2$k$ steps.
For text-to-image-to-3D, 2D images from the input text are created with the assistance of DALL-E-3\footnote{\href{https://cdn.openai.com/papers/dall-e-3.pdf}{https://cdn.openai.com/papers/dall-e-3.pdf}}.

\vspace{\paramargin}
\paragraph{Baselines.}
In our work, we employ several state-of-the-art methods as baseline models: Zero123~\cite{liu2023zero}, RealFusion~\cite{melas2023realfusion}, Magic123~\cite{qian2023magic123}, One-2-3-45~\cite{liu2023one}, Point-E~\cite{nichol2022point}, Shap-E~\cite{jun2023shap}, and SyncDreamer~\cite{liu2023syncdreamer}.
Zero123~\cite{liu2023zero} is able to generate novel-view images of an object from various viewpoints given a single-view image.
Moreover, its integration with the SDS loss~\cite{poole2022dreamfusion} bolsters its capability for 3D reconstruction from single-view images.
RealFusion~\cite{melas2023realfusion} leverages Stable Diffusion~\cite{rombach2022high} and the SDS loss for achieving high-quality single-view reconstruction.
Magic123~\cite{qian2023magic123} builds upon the strengths of Zero123~\cite{liu2023zero} and RealFusion~\cite{melas2023realfusion}, resulting in a method that further improves the overall quality of 3D reconstruction.
One-2-3-45~\cite{liu2023one} takes a direct approach by regressing Signed Distance Functions (SDFs) from the output images of Zero123~\cite{liu2023zero}.
Point-E~\cite{nichol2022point} and Shap-E~\cite{jun2023shap} represent 3D generative models trained on an extensive 3D dataset.
Both models exhibit the capability to convert a single-view image into either a point cloud or a shape encoded in an MLP.
SyncDreamer~\cite{liu2023syncdreamer} produces multi-view coherent images from a single-view image by synchronizing intermediate states of generated images using a 3D-aware feature attention mechanism.


\section{Correlation between CD score and Human Evaluation}
\input{figs/user_study}
To assess the efficacy of HarmonyView against SyncDreamer~\cite{liu2023syncdreamer} and Zero123~\cite{liu2023zero}, we conducted a user study where participants rated the three approaches using a 5-point Likert-scale (1-5), evaluating (a) Quality, (b) Consistency, and (c) Diversity.
Our user study, showcased in~\cref{fig:user_study}, reveals a consistent alignment between the CD Score ($\text{CD Score} = D / \text{S}_{Var}$) and human evaluation metrics.
Throughout the study, we observed that the $\text{CD Score}$ reliably reflects the correlation between two key factors: $\text{S}_{Var}$, measuring the diversity in generated images' alignment with a given text prompt, and $D$, evaluating creative variation against a reference view using CLIP image encoders.

\noindent 1. \textbf{Semantic Variance ($\text{S}_{Var}$) and Consistency}: Lower Semantic Variance consistently corresponds to higher consistency in human evaluation. In simpler terms, when the generated images are more aligned in their interpretation of the text prompt, human evaluators tend to agree more on the perceived consistency. This correlation implies that there's a negative relationship between Semantic Variance and Consistency --- lower variance often leads to higher agreement among evaluators.

\noindent 2. \textbf{Diversity Score ($D$) and Quality Perception}: Higher Diversity Scores tend to lead to lower quality perceptions in human evaluation. This suggests a somewhat negative correlation between Diversity Score and Quality Perception. Put differently, when the diversity among the generated images is higher --- meaning they deviate more from the reference image --- human evaluators tend to perceive lower quality. Conversely, higher similarity between the generated images and the reference image correlates with higher perceived quality. In essence, when the visual similarity between the input and target views is higher, the quality tends to be perceived as better by human evaluators.

These findings collectively underscore the critical balance needed between semantic diversity and adherence to the reference image in the pursuit of generating high-quality images aligned with text prompts.
Achieving this delicate equilibrium is pivotal to ensure that generated images are diverse enough to capture different interpretations while also being faithful enough to the reference to maintain perceived quality.
HarmonyView demonstrated the highest CD score compared to SyncDreamer~\cite{liu2023syncdreamer} and Zero123~\cite{liu2023zero}, indicating that our generated images strike a winning balance between consistency and diversity, excelling in both aspects of fidelity to the reference image and semantic variation.




\section{Additional Results}

\input{figs/appendix_ablation}
\input{figs/appendix_qualitative_nvs}

\subsection{Novel-view Synthesis}

\paragraph{Qualitative ablation study.}
Our HarmonyView decomposes multi-view diffusion guidance into two distinct guidance components (see \cref{eq:ours_final}): $s_1$ primarily serves to ensure visual consistency between the input and target views, while $s_2$ focuses on amplifying diversity across novel viewpoints.
The significance of this approach is showcased in~\cref{fig:appendix_ablation}, where we visually demonstrate how each guidance factor influences the synthesized images.
When prioritizing $s_1$, the quality of synthesis improves significantly as it focuses on aligning the visual consistency between the input and target views.
However, in specific cases, like the deer sample, it generates multiple faces of the deer, leading to what's known as the ``Janus problem" --- creating facial features on the rear side akin to the front, causing visual anomalies.
On the other hand, emphasizing $s_2$ results in increased diversity across the generated samples.
However, a fundamental trade-off exists between these two aspects --- quality and diversity --- making it challenging to optimize for both simultaneously.
Yet, by employing both $s_1$ and $s_2$ in tandem, we can achieve a win-win scenario.
This division allows us to precisely discern the impact of each guidance factor on the generation process.
By skillfully balancing these guiding principles, our method becomes empowered to generate a rich and varied array of images, exhibiting both multi-view coherence and fidelity to the input view.

\input{tables/statistical_analysis}

\vspace{\paramargin}
\paragraph{Qualitative comparison.}
Figure~\ref{fig:appendix_nvs} provides a glimpse into the capabilities and limitations of different novel-view synthesis methods.
Zero123~\cite{liu2023zero} frequently generates images that lack coherence across multiple viewpoints.
These synthesized images often contain implausible variations, such as alterations in the number of cymbals or trees based on the view, or even changes in the shape of eyes.
These inconsistencies underscore the struggle of Zero123 to maintain coherence and realism across different perspectives, leading to discrepancies that compromise the overall quality of multi-view synthesis.
SyncDreamer~\cite{liu2023syncdreamer} faces challenges in preserving the expected visual similarity across different viewpoints.
The generated images often display deviations in overall size, empty or missing regions, or distorted forms, leading to an overall loss of visual completeness and integrity.
Instances where facial features are erased or distorted represent the difficulties SyncDreamer encounters in maintaining the visual fidelity expected across diverse views.
In stark contrast, HarmonyView stands out for its ability to generate diverse yet plausible multi-view images while preserving geometric coherence across these views.
Unlike its counterparts, HarmonyView maintains a harmonious relationship between different views, ensuring the consistent appearance, shapes, and elements of objects.
In addition, HarmonyView can extrapolate realistic frontal views from the rear-view input image (see third sample).
This further underscores the versatility and robustness of HarmonyView.
Overall, HarmonyView is able to generate a diverse set of images while maintaining a sense of realism and coherence across the multiple views.

\vspace{\paramargin}
\paragraph{Statistical analysis.}
In~\Cref{tab:statistic_analysis}, we conduct a comprehensive statistical analysis on the GSO~\cite{downs2022google} dataset, evaluating the performance of three methods: HarmonyView, Zeor123~\cite{liu2023zero}, and SyncDreamer~\cite{liu2023syncdreamer}.
We report PSNR, SSIM, LPIPS, and $E_{flow}$ for the best-matched instance with ground truth, as well as the average and variance across four instances.
Upon comparison, HarmonyView demonstrates superior performance across all metrics when compared to Zeor123 and SyncDreamer.
It attains the highest scores in PSNR and SSIM, indicating better image quality in terms of both fidelity and structural similarity when compared to the ground truth.
Moreover, HarmonyView also exhibits the lowest LPIPS and $E_{flow}$ scores, signifying reduced perceptual differences and flow errors when matched against the ground truth.
Interestingly, HarmonyView shows higher variability (indicated by larger variance values) across instances compared to other methods.
This variability might imply that while HarmonyView generally performs well, its performance might fluctuate more across different instances or scenarios compared to the Zero123 and SyncDreamer.
Nevertheless, it is essential to note that this variability in performance also reflects its diversity in samples.
This could imply that while HarmonyView showcases a broader range of outputs, it still maintains a high level of image quality.


\input{figs/appendix_qualitative_3d_recon}

\subsection{3D Reconstruction}
In~\cref{fig:appendix_recon}, the results exemplify HarmonyView's exceptional quality compared to other methods evaluated for 3D reconstruction.
While contrasting with competing methods, it is evident that these approaches encounter various challenges in handling the reconstruction process.
For instance, both Point-E~\cite{nichol2022point} and Shap-E~\cite{jun2023shap} struggle significantly with incomplete reconstructions, failing to capture the entirety of the intended 3D shapes.
This deficiency results in reconstructions that lack certain crucial elements, undermining the fidelity of the output.
In the case of One-2-3-45~\cite{liu2023one}, the method exhibits a tendency to produce ambiguous shapes, failing to accurately represent the intended shape contours.
Furthermore, Zero123~\cite{liu2023zero} faces difficulties in capturing fine elements within the reconstructed shapes, which diminishes the overall fidelity and detail level of the output.
SyncDremaer~\cite{liu2023syncdreamer} also shows discontinuities or holes within the generated 3D meshes.
These imperfections detract from the coherence and completeness of the reconstructed shape.
In contrast, HarmonyView produces high-quality 3D meshes that achieve accurate geometry while maintaining a realistic appearance.
Its ability to circumvent the pitfalls experienced by other methods speaks volumes about its capability to generate comprehensive, detailed, and visually compelling reconstructions.

\section{Discussion}
\subsection{Limitations \& Future Work}
While HarmonyView demonstrates promising results in enhancing both visual consistency and novel-view diversity in single-image 3D content generation, several limitations warrant further investigation.
Firstly, our multi-view diffusion formulation somewhat mitigates inherent trade-offs between consistency and diversity to achieve a certain level of Pareto optimality.
However, the complete separation of these aspects to eliminate the trade-off entirely remains a challenging pursuit.
Secondly, HarmonyView's current focus primarily revolves around object-centric scenes.
This poses limitations when dealing with complex scenarios involving multiple interacting objects, varying scales, and intricate geometries.
Expanding the technique to encompass such diverse and intricate scenes demands innovative approaches that account for object interactions, spatial relationships, and contextual understanding within the scene.
Moreover, our current setting typically involves single objects without backgrounds, simplifying the requirements for realism and diversity.
The ignorance of background significantly reduces the expectations of synthesizing diverse images.
To accommodate in-the-wild multi-object scenes with complex backgrounds, HarmonyView requires the use of an external background removal tool (\eg, CarveKit).
Addressing these limitations effectively presents ample opportunities for innovation and refinement within the field.
Exploring these avenues promises to advance the field towards more comprehensive and realistic 3D content generation from single images.

\subsection{Ethical Considerations}
The advancements in one-image-to-3D bring forth several ethical considerations that demand careful attention.
One key concern is the potential misuse of generated 3D content.
These advancements could be exploited to create deceptive or misleading visual information, leading to misinformation or even malicious activities like deepfakes, where fabricated content is passed off as genuine, potentially causing harm, misinformation, or manipulation.
It is essential to establish responsible usage guidelines and ethical standards to prevent the abuse of this technology.
Another critical concern is the inherent bias within the training data, which might lead to biased representations or unfair outcomes.
Ensuring diverse and representative training datasets and continuously monitoring and addressing biases are essential to mitigate such risks.
Moreover, the technology poses privacy implications, as it could be used to reconstruct 3D models of objects and scenes from any images.
Images taken without consent or from public spaces could be used to reconstruct detailed 3D models, potentially violating personal privacy boundaries.
As such, it is crucial to implement appropriate safeguards and obtain informed consent when working with images containing personal information.

%% file: figs/user_study.tex
\begin{figure*}
    \centering
    \setlength\tabcolsep{1pt}
    \resizebox{\linewidth}{!}{
    \renewcommand{\arraystretch}{0.5}
    \begin{tabular}{cc}
        \includegraphics[width=0.49\textwidth]{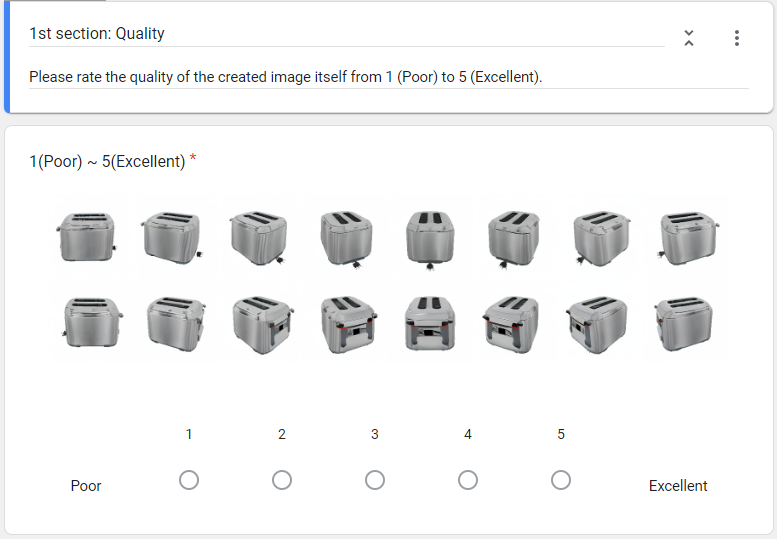} & \includegraphics[width=0.49\textwidth]{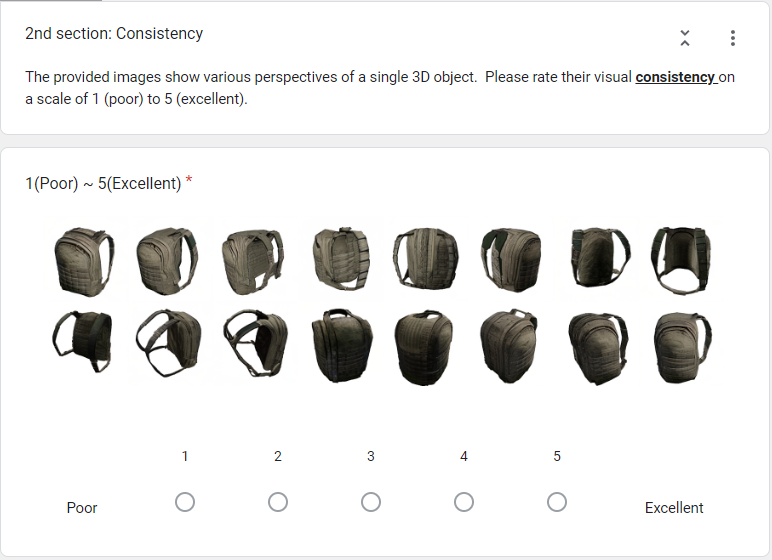} \\
        (a) Quality & (b) Consistency \\
        {} & {} \\
        \multicolumn{2}{c}{\includegraphics[width=0.49\textwidth]{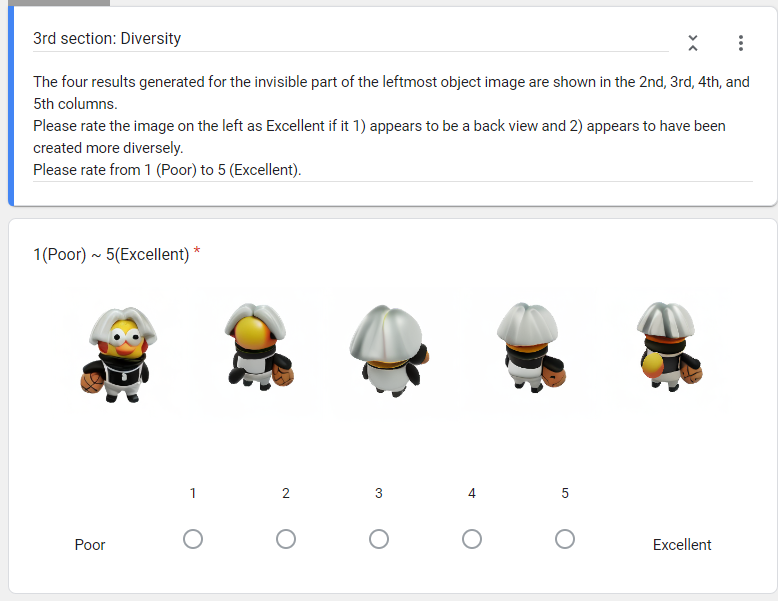}} \\
        \multicolumn{2}{c}{(c) Diversity}
    \end{tabular}
    }
    \vspace{\abovefigcapmargin}
    \caption{
        \textbf{User evaluation examples.}
        We perform a user study to evaluate the effectiveness of our approach, HarmonyView, in comparison to SyncDreamer~\cite{liu2023syncdreamer} and Zero123~\cite{liu2023zero}.
        Participants were asked to rate the three approaches using a 5-point Likert-scale (1-5), assessing (a) Quality, (b) Consistency, and (c) Diversity.
    }
    \vspace{\belowfigcapmargin}
    \label{fig:user_study}
\end{figure*}

%% file: figs/appendix_ablation.tex
\begin{figure*}
    \centering
    \setlength\tabcolsep{1pt}
    \resizebox{\linewidth}{!}{
    \renewcommand{\arraystretch}{0.5}
    \begin{tabular}{c:cccc:cccc:cccc}
    \includegraphics[width=0.075\textwidth]{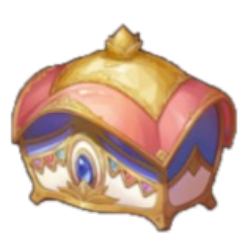} &
    \includegraphics[width=0.075\textwidth]{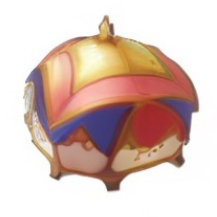} &
    \includegraphics[width=0.075\textwidth]{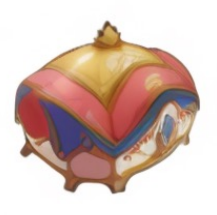} &
    \includegraphics[width=0.075\textwidth]{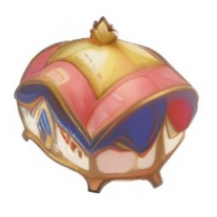} &
    \includegraphics[width=0.075\textwidth]{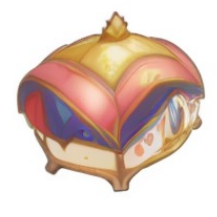} &
    \includegraphics[width=0.075\textwidth]{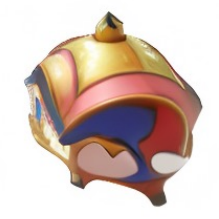} &
    \includegraphics[width=0.075\textwidth]{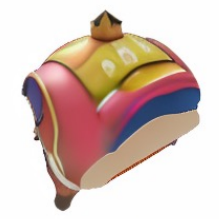} &
    \includegraphics[width=0.075\textwidth]{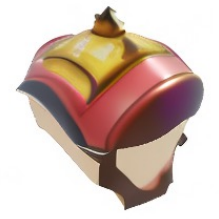} &
    \includegraphics[width=0.075\textwidth]{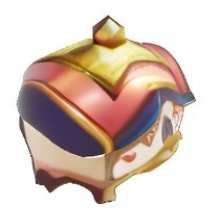} &
    \includegraphics[width=0.075\textwidth]{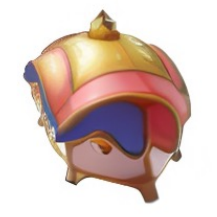} &
    \includegraphics[width=0.075\textwidth]{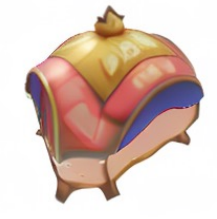} &
    \includegraphics[width=0.075\textwidth]{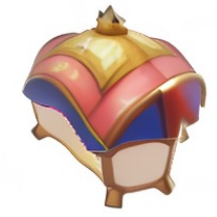} &
    \includegraphics[width=0.075\textwidth]{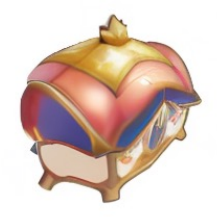} \\
         &
    \includegraphics[width=0.075\textwidth]{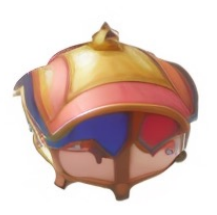} &
    \includegraphics[width=0.075\textwidth]{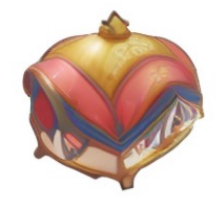} &
    \includegraphics[width=0.075\textwidth]{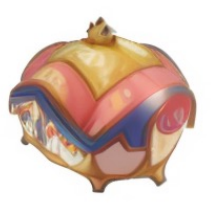} &
    \includegraphics[width=0.075\textwidth]{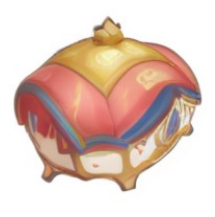} &
    \includegraphics[width=0.075\textwidth]{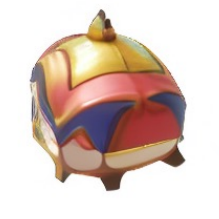} &
    \includegraphics[width=0.075\textwidth]{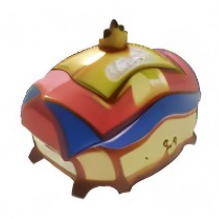} &
    \includegraphics[width=0.075\textwidth]{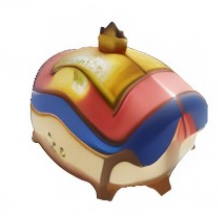} &
    \includegraphics[width=0.075\textwidth]{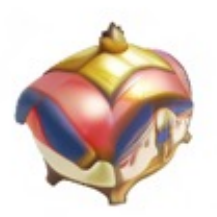} &
    \includegraphics[width=0.075\textwidth]{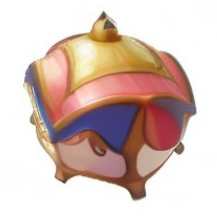} &
    \includegraphics[width=0.075\textwidth]{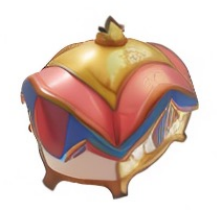} &
    \includegraphics[width=0.075\textwidth]{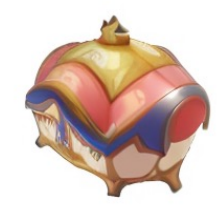} &
    \includegraphics[width=0.075\textwidth]{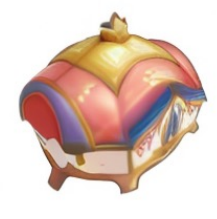} \\
    \includegraphics[width=0.075\textwidth]{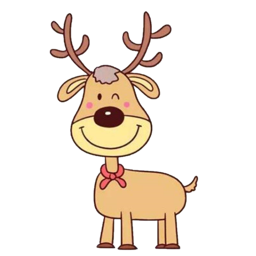} &
    \includegraphics[width=0.075\textwidth]{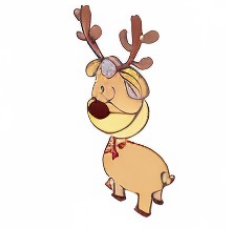} &
    \includegraphics[width=0.075\textwidth]{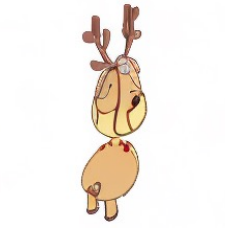} &
    \includegraphics[width=0.075\textwidth]{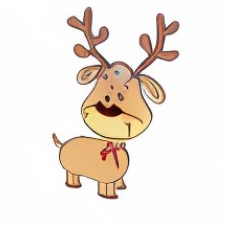} &
    \includegraphics[width=0.075\textwidth]{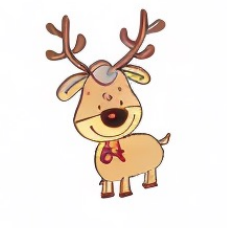} &
    \includegraphics[width=0.075\textwidth]{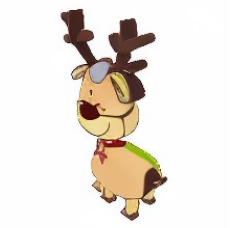} &
    \includegraphics[width=0.075\textwidth]{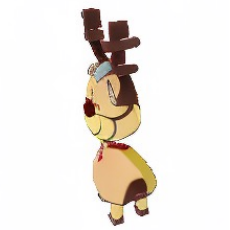} &
    \includegraphics[width=0.075\textwidth]{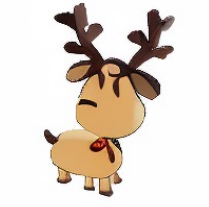} &
    \includegraphics[width=0.075\textwidth]{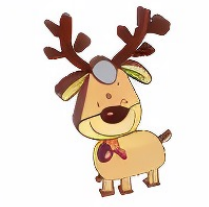} &
    \includegraphics[width=0.075\textwidth]{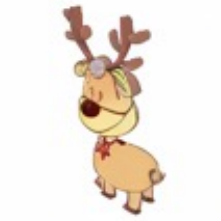} &
    \includegraphics[width=0.075\textwidth]{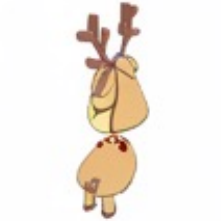} &
    \includegraphics[width=0.075\textwidth]{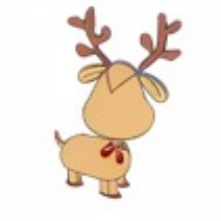} &
    \includegraphics[width=0.075\textwidth]{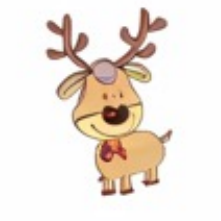} \\
         &
    \includegraphics[width=0.075\textwidth]{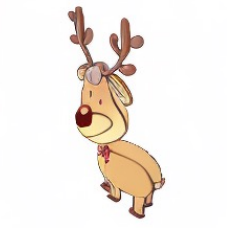} &
    \includegraphics[width=0.075\textwidth]{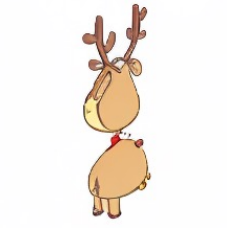} &
    \includegraphics[width=0.075\textwidth]{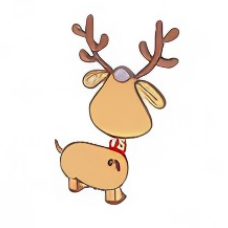} &
    \includegraphics[width=0.075\textwidth]{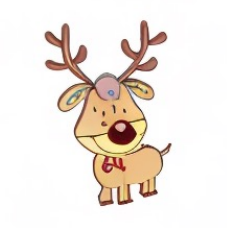} &
    \includegraphics[width=0.075\textwidth]{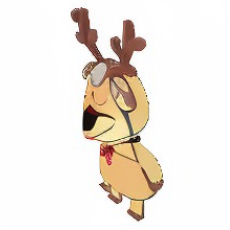} &
    \includegraphics[width=0.075\textwidth]{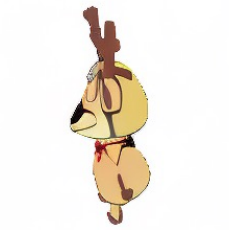} &
    \includegraphics[width=0.075\textwidth]{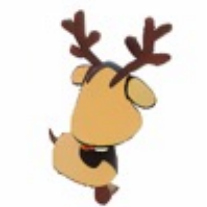} &
    \includegraphics[width=0.075\textwidth]{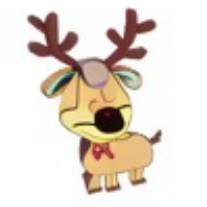} &
    \includegraphics[width=0.075\textwidth]{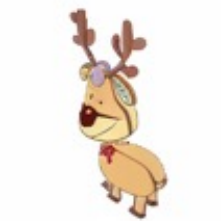} &
    \includegraphics[width=0.075\textwidth]{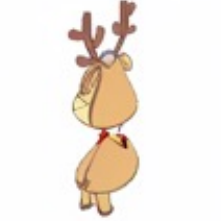} &
    \includegraphics[width=0.075\textwidth]{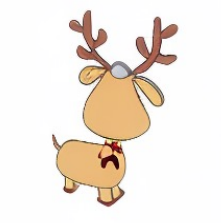} &
    \includegraphics[width=0.075\textwidth]{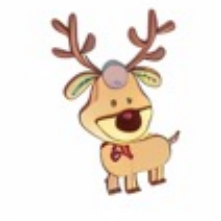} \\
    \includegraphics[width=0.075\textwidth]{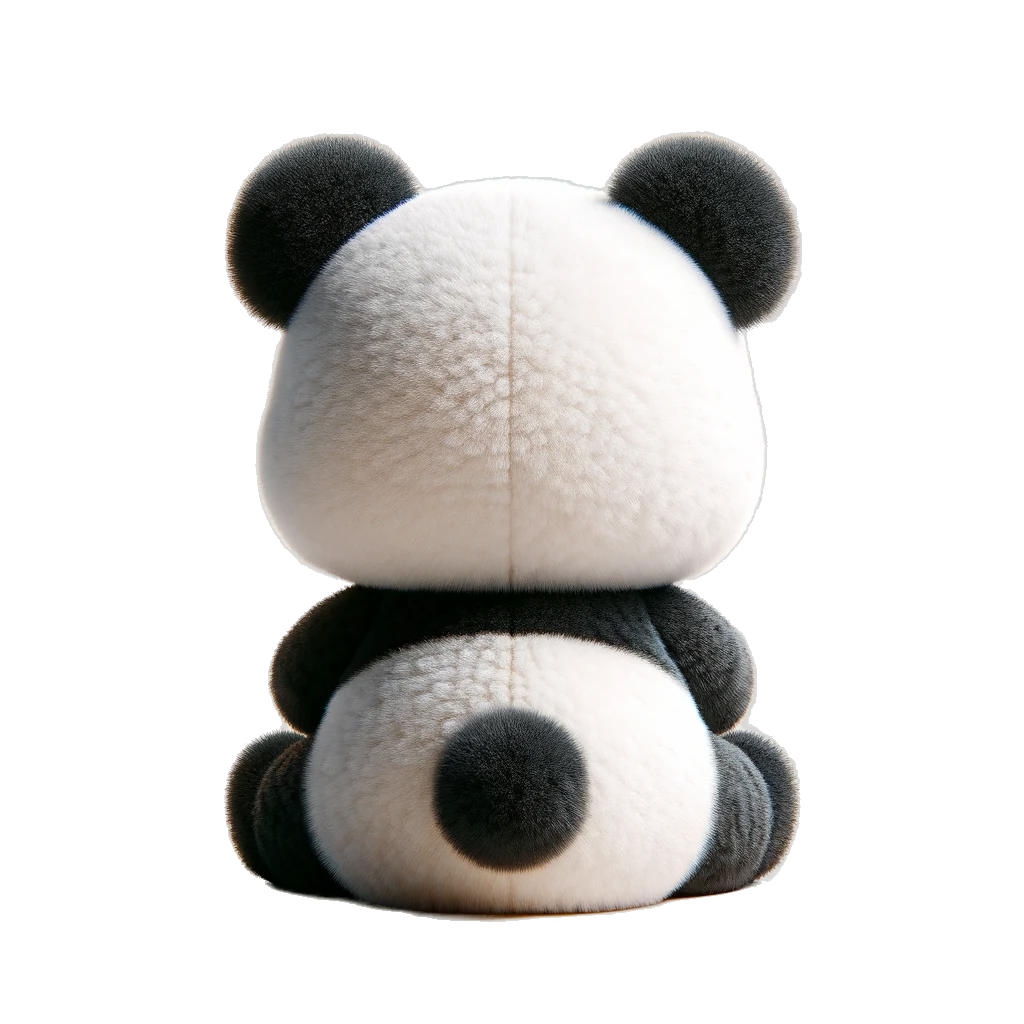} &
    \includegraphics[width=0.075\textwidth]{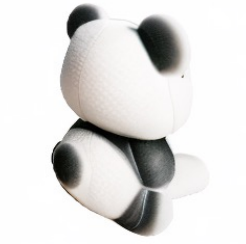} &
    \includegraphics[width=0.075\textwidth]{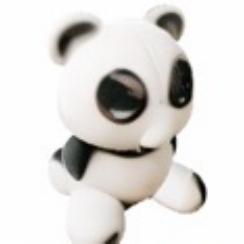} &
    \includegraphics[width=0.075\textwidth]{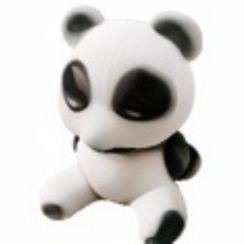} &
    \includegraphics[width=0.075\textwidth]{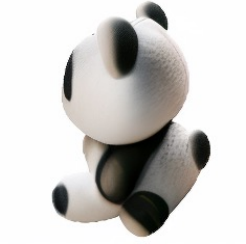} &
    \includegraphics[width=0.075\textwidth]{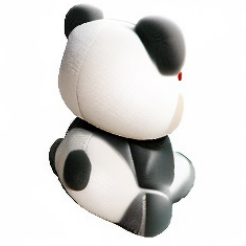} &
    \includegraphics[width=0.075\textwidth]{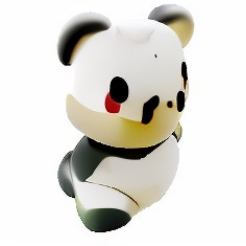} &
    \includegraphics[width=0.075\textwidth]{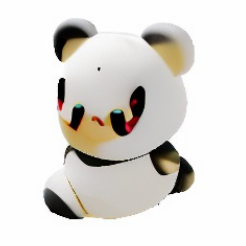} &
    \includegraphics[width=0.075\textwidth]{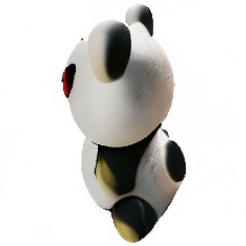} &
    \includegraphics[width=0.075\textwidth]{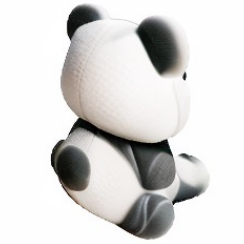} &
    \includegraphics[width=0.075\textwidth]{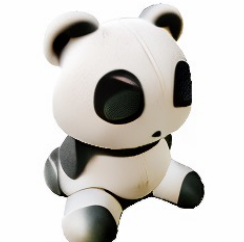} &
    \includegraphics[width=0.075\textwidth]{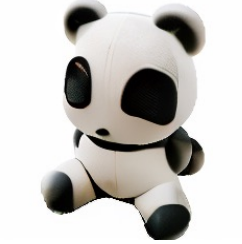} &
    \includegraphics[width=0.075\textwidth]{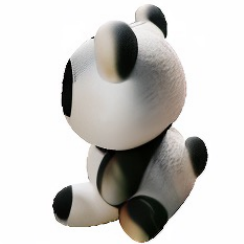}  \\
     &
    \includegraphics[width=0.075\textwidth]{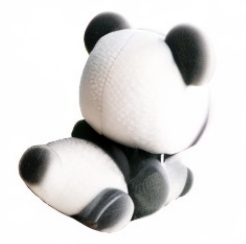} &
    \includegraphics[width=0.075\textwidth]{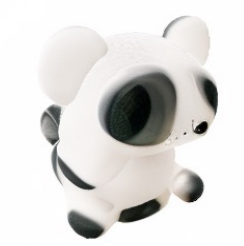} &
    \includegraphics[width=0.075\textwidth]{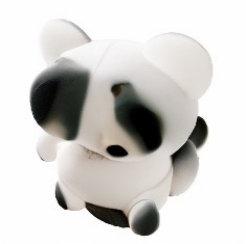} &
    \includegraphics[width=0.075\textwidth]{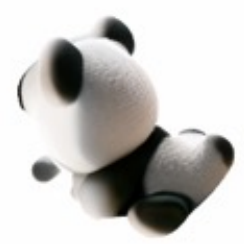} &
    \includegraphics[width=0.075\textwidth]{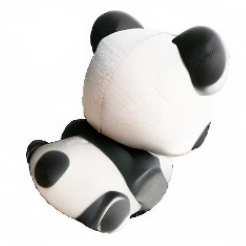} &
    \includegraphics[width=0.075\textwidth]{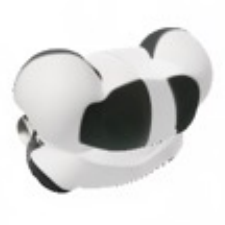} &
    \includegraphics[width=0.075\textwidth]{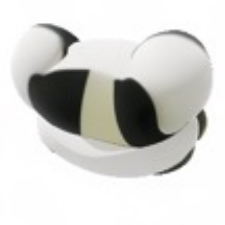} &
    \includegraphics[width=0.075\textwidth]{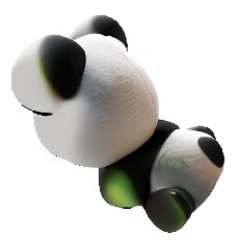} &
    \includegraphics[width=0.075\textwidth]{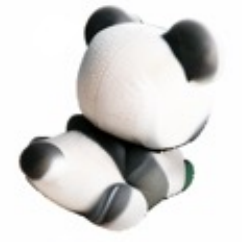} &
    \includegraphics[width=0.075\textwidth]{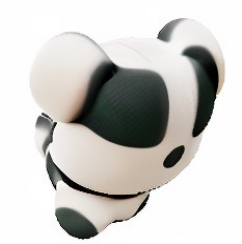} &
    \includegraphics[width=0.075\textwidth]{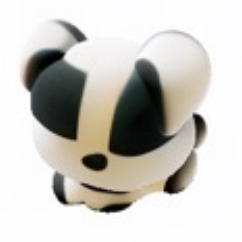} &
    \includegraphics[width=0.075\textwidth]{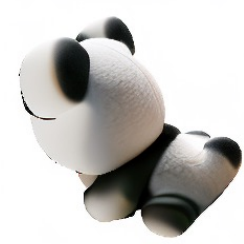} \\
         &
    \includegraphics[width=0.075\textwidth]{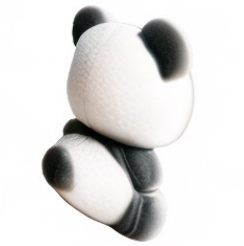} &
    \includegraphics[width=0.075\textwidth]{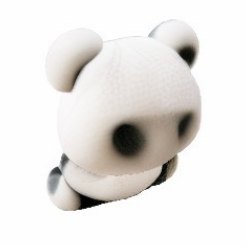} &
    \includegraphics[width=0.075\textwidth]{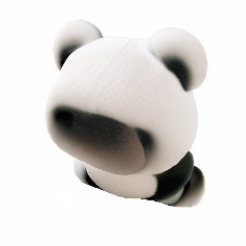} &
    \includegraphics[width=0.075\textwidth]{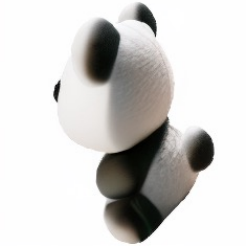} &
    \includegraphics[width=0.075\textwidth]{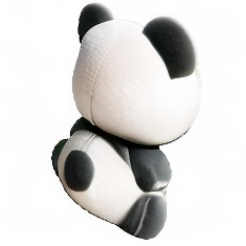} &
    \includegraphics[width=0.075\textwidth]{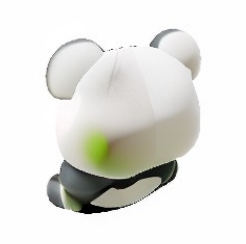} &
    \includegraphics[width=0.075\textwidth]{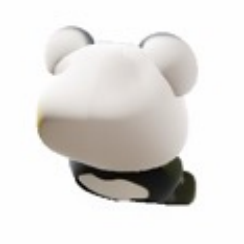} &
    \includegraphics[width=0.075\textwidth]{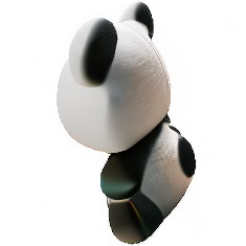} &
    \includegraphics[width=0.075\textwidth]{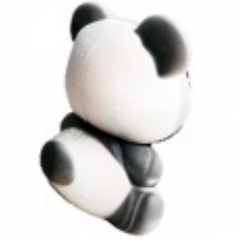} &
    \includegraphics[width=0.075\textwidth]{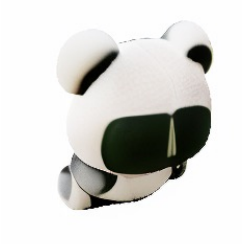} &
    \includegraphics[width=0.075\textwidth]{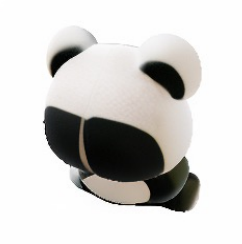} &
    \includegraphics[width=0.075\textwidth]{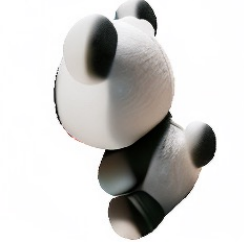}  \\
    \includegraphics[width=0.075\textwidth]{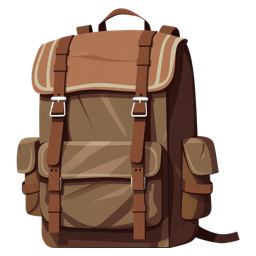} &
    \includegraphics[width=0.075\textwidth]{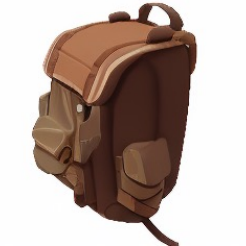} &
    \includegraphics[width=0.075\textwidth]{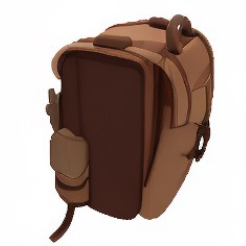} &
    \includegraphics[width=0.075\textwidth]{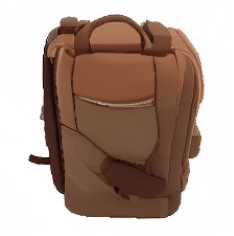} &
    \includegraphics[width=0.075\textwidth]{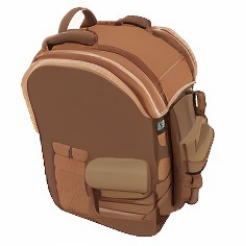} &
    \includegraphics[width=0.075\textwidth]{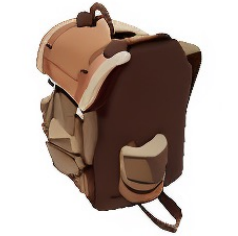} &
    \includegraphics[width=0.075\textwidth]{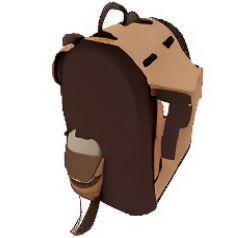} &
    \includegraphics[width=0.075\textwidth]{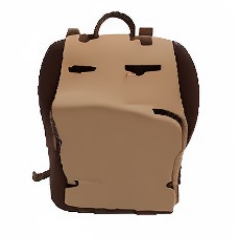} &
    \includegraphics[width=0.075\textwidth]{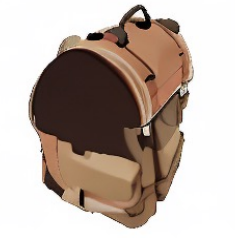} &
    \includegraphics[width=0.075\textwidth]{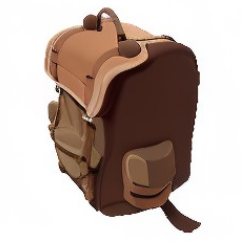} &
    \includegraphics[width=0.075\textwidth]{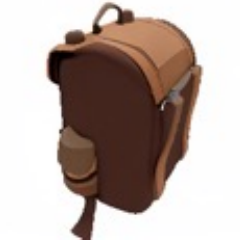} &
    \includegraphics[width=0.075\textwidth]{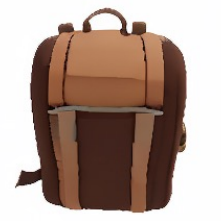} &
    \includegraphics[width=0.075\textwidth]{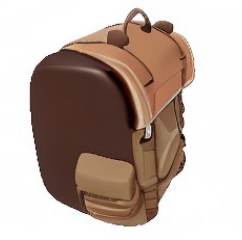}  \\
     &
    \includegraphics[width=0.075\textwidth]{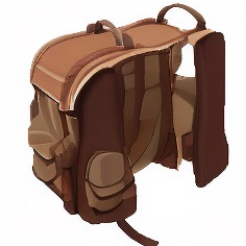} &
    \includegraphics[width=0.075\textwidth]{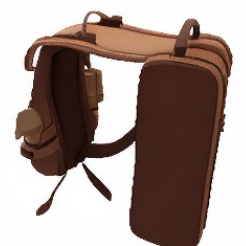} &
    \includegraphics[width=0.075\textwidth]{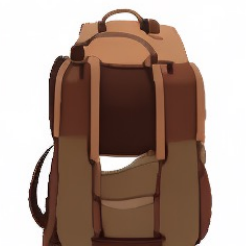} &
    \includegraphics[width=0.075\textwidth]{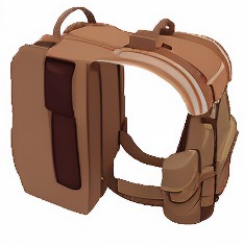} &
    \includegraphics[width=0.075\textwidth]{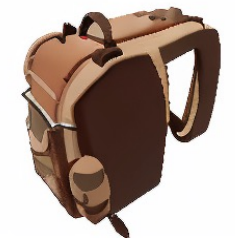} &
    \includegraphics[width=0.075\textwidth]{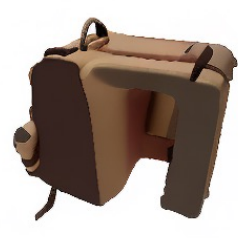} &
    \includegraphics[width=0.075\textwidth]{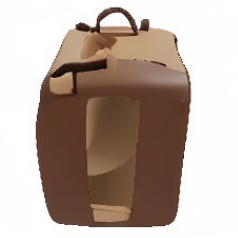} &
    \includegraphics[width=0.075\textwidth]{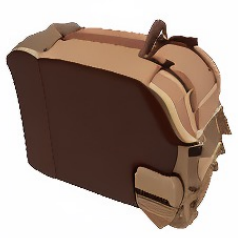} &
    \includegraphics[width=0.075\textwidth]{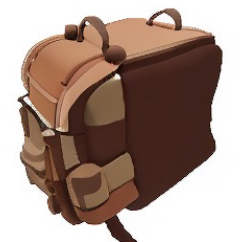} &
    \includegraphics[width=0.075\textwidth]{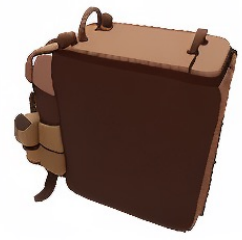} &
    \includegraphics[width=0.075\textwidth]{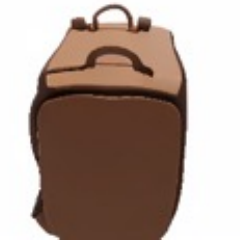} &
    \includegraphics[width=0.075\textwidth]{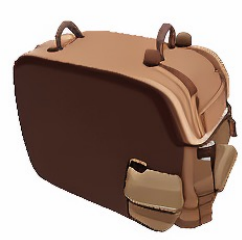} \\
         &
    \includegraphics[width=0.075\textwidth]{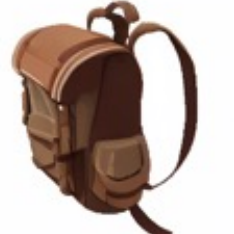} &
    \includegraphics[width=0.075\textwidth]{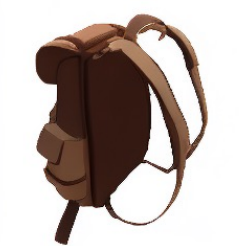} &
    \includegraphics[width=0.075\textwidth]{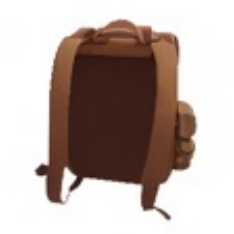} &
    \includegraphics[width=0.075\textwidth]{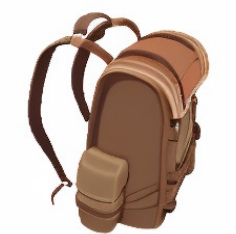} &
    \includegraphics[width=0.075\textwidth]{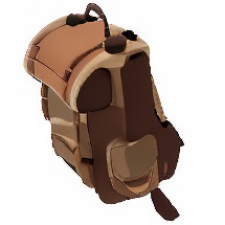} &
    \includegraphics[width=0.075\textwidth]{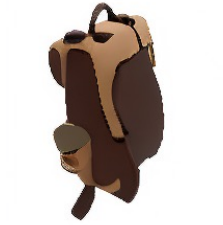} &
    \includegraphics[width=0.075\textwidth]{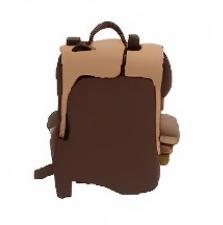} &
    \includegraphics[width=0.075\textwidth]{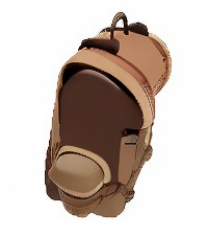} &
    \includegraphics[width=0.075\textwidth]{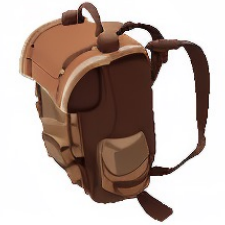} &
    \includegraphics[width=0.075\textwidth]{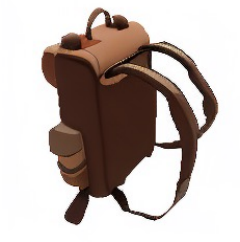} &
    \includegraphics[width=0.075\textwidth]{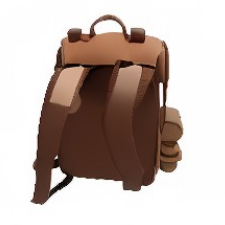} &
    \includegraphics[width=0.075\textwidth]{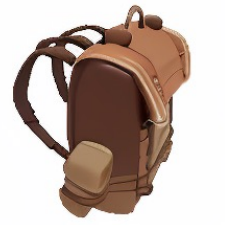}  \\    
    \multicolumn{1}{c}{Input} & \multicolumn{4}{c}{only $s_{1}$} & \multicolumn{4}{c}{only $s_{2}$} & \multicolumn{4}{c}{Both} \\
    \end{tabular}
    }
    \vspace{\abovefigcapmargin}
    \caption{
        \textbf{Qualitative ablation study on novel-view synthesis.}
        Our HarmonyView guides the multi-view diffusion process with two parameters, $s_1$ and $s_2$ (see \cref{eq:ours_final}).
        The nuanced interplay between $s_1$ and $s_2$ impacts consistency and diversity throughout the generation process.
        By skillfully balancing these guiding principles, we can achieve a win-win scenario: generate diverse images that maintain coherence across multiple views and stay faithful to the input view.
    }
    \vspace{\belowfigcapmargin}
    \label{fig:appendix_ablation}
\end{figure*}

%% file: figs/appendix_qualitative_nvs.tex
\begin{figure*}
    \centering
    \setlength\tabcolsep{1pt}
    \resizebox{\linewidth}{!}{
    \renewcommand{\arraystretch}{0.5}
    \begin{tabular}{c:cccc:cccc:cccc}
        \includegraphics[width=0.075\textwidth]{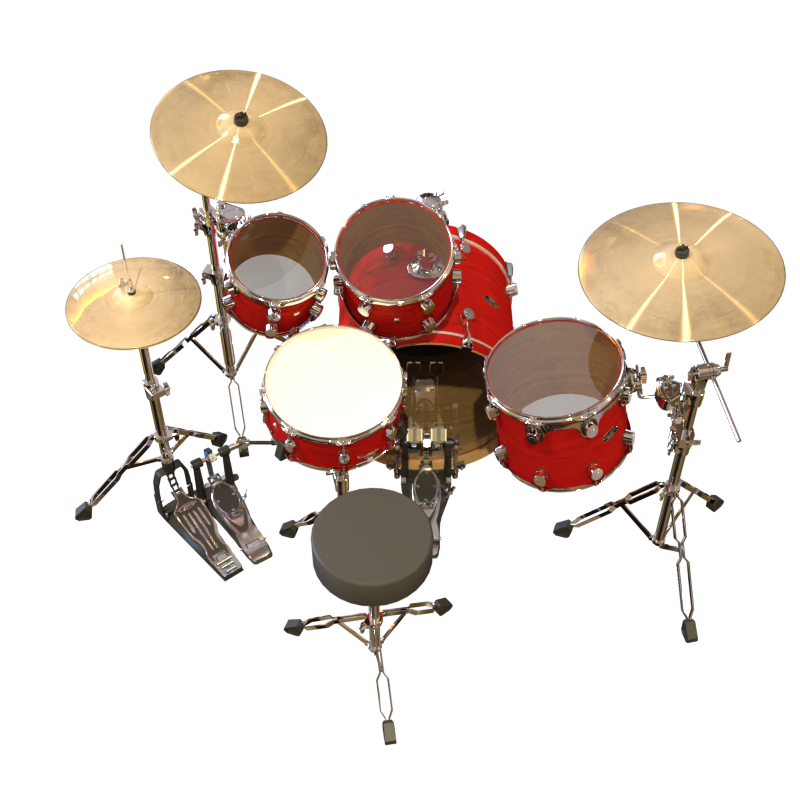} &
        \includegraphics[width=0.075\textwidth]{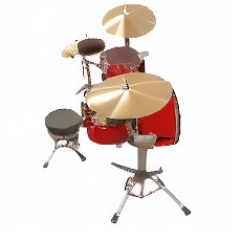} &
        \includegraphics[width=0.075\textwidth]{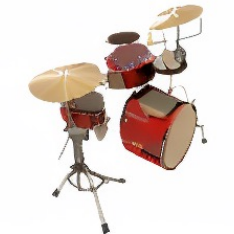} &
        \includegraphics[width=0.075\textwidth]{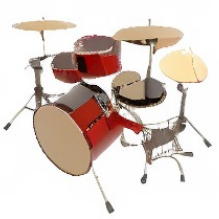} &
        \includegraphics[width=0.075\textwidth]{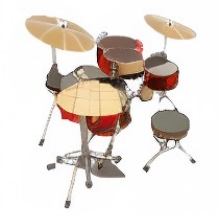} &
        \includegraphics[width=0.075\textwidth]{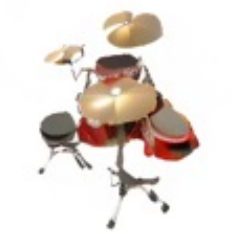} &
        \includegraphics[width=0.075\textwidth]{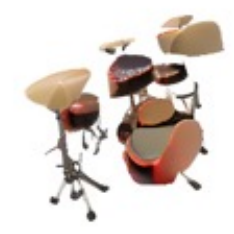} &
        \includegraphics[width=0.075\textwidth]{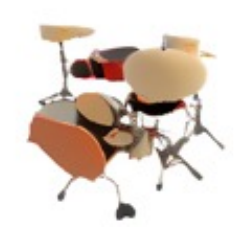} &
        \includegraphics[width=0.075\textwidth]{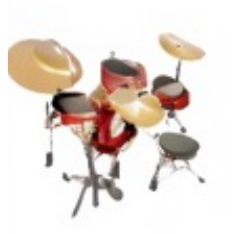} &
        \includegraphics[width=0.075\textwidth]{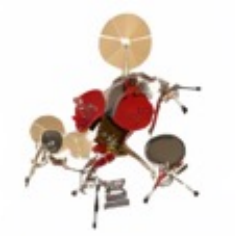} &
        \includegraphics[width=0.075\textwidth]{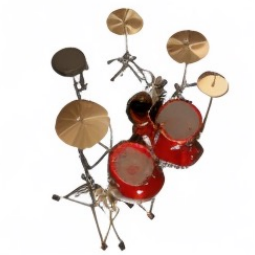} &
        \includegraphics[width=0.075\textwidth]{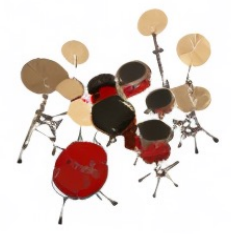} &
        \includegraphics[width=0.075\textwidth]{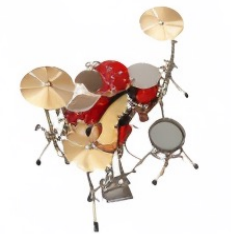} \\
         & 
        \includegraphics[width=0.075\textwidth]{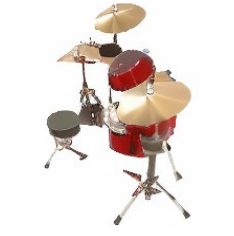} &
        \includegraphics[width=0.075\textwidth]{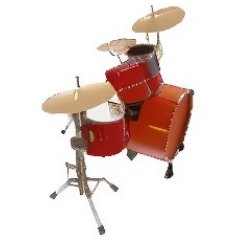} &
        \includegraphics[width=0.075\textwidth]{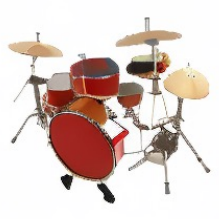} &
        \includegraphics[width=0.075\textwidth]{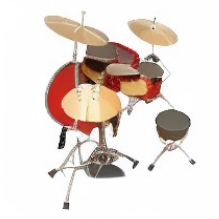} &
        \includegraphics[width=0.075\textwidth]{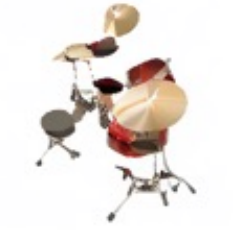} &
        \includegraphics[width=0.075\textwidth]{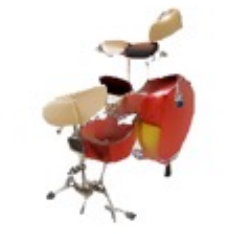} &
        \includegraphics[width=0.075\textwidth]{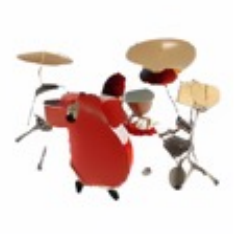} &
        \includegraphics[width=0.075\textwidth]{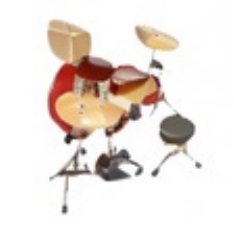} &
        \includegraphics[width=0.075\textwidth]{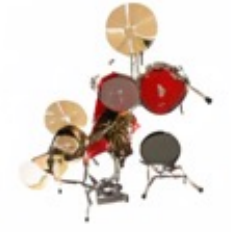} &
        \includegraphics[width=0.075\textwidth]{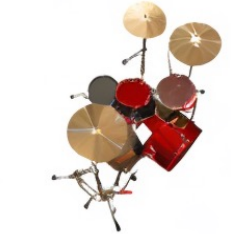} &
        \includegraphics[width=0.075\textwidth]{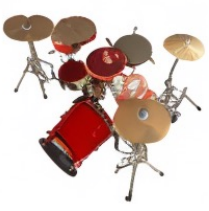} &
        \includegraphics[width=0.075\textwidth]{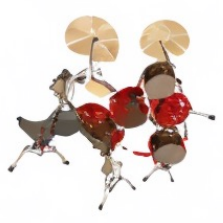} \\
        \includegraphics[width=0.075\textwidth]{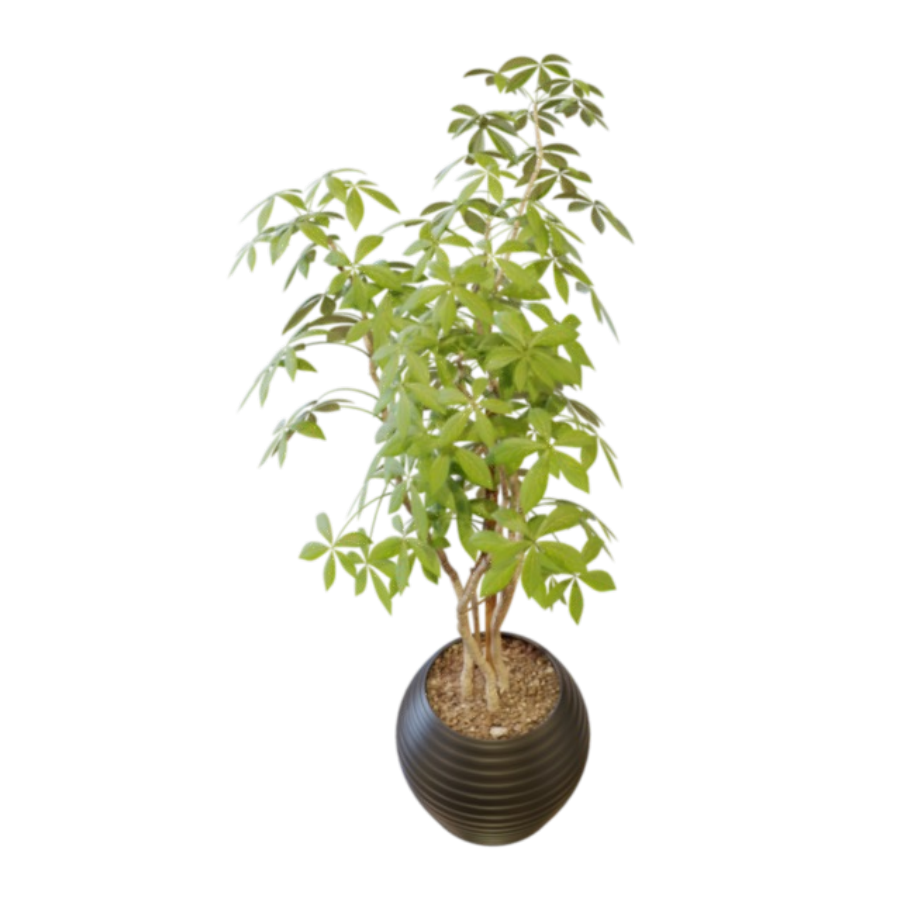} &
        \includegraphics[width=0.075\textwidth]{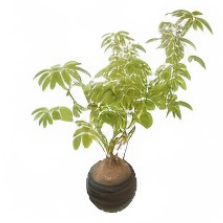} &
        \includegraphics[width=0.075\textwidth]{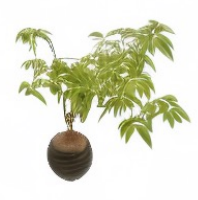} &
        \includegraphics[width=0.075\textwidth]{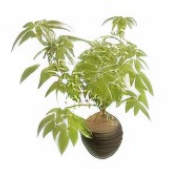} &
        \includegraphics[width=0.075\textwidth]{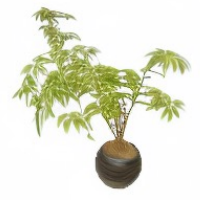} &
        \includegraphics[width=0.075\textwidth]{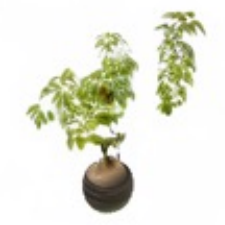} &
        \includegraphics[width=0.075\textwidth]{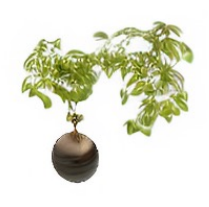} &
        \includegraphics[width=0.075\textwidth]{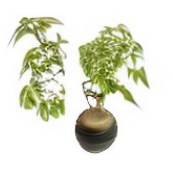} &
        \includegraphics[width=0.075\textwidth]{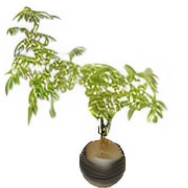} &
        \includegraphics[width=0.075\textwidth]{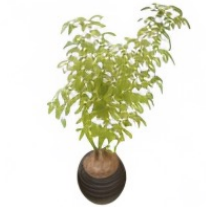} &
        \includegraphics[width=0.075\textwidth]{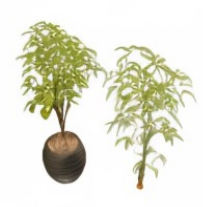} &
        \includegraphics[width=0.075\textwidth]{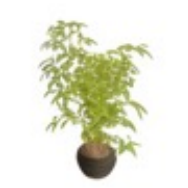} &
        \includegraphics[width=0.075\textwidth]{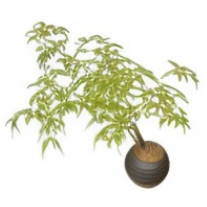} \\
        &
        \includegraphics[width=0.075\textwidth]{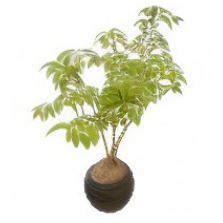} &
        \includegraphics[width=0.075\textwidth]{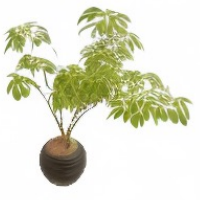} &
        \includegraphics[width=0.075\textwidth]{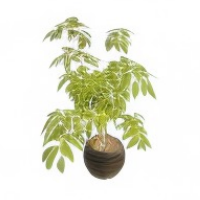} &
        \includegraphics[width=0.075\textwidth]{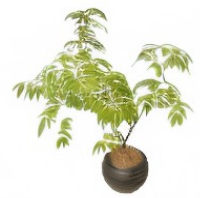} &
        \includegraphics[width=0.075\textwidth]{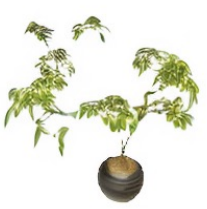} &
        \includegraphics[width=0.075\textwidth]{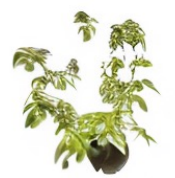} &
        \includegraphics[width=0.075\textwidth]{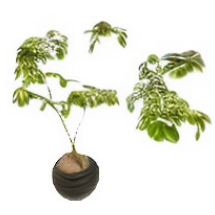} &
        \includegraphics[width=0.075\textwidth]{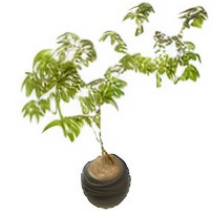} &
        \includegraphics[width=0.075\textwidth]{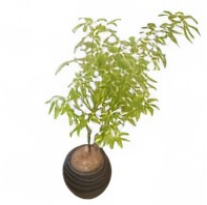} &
        \includegraphics[width=0.075\textwidth]{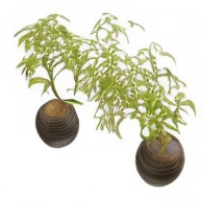} &
        \includegraphics[width=0.075\textwidth]{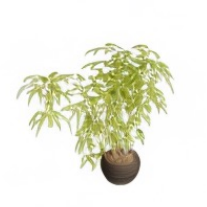} &
        \includegraphics[width=0.075\textwidth]{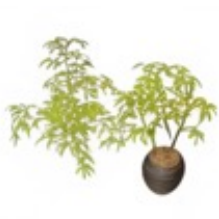} \\  
        \includegraphics[width=0.075\textwidth]{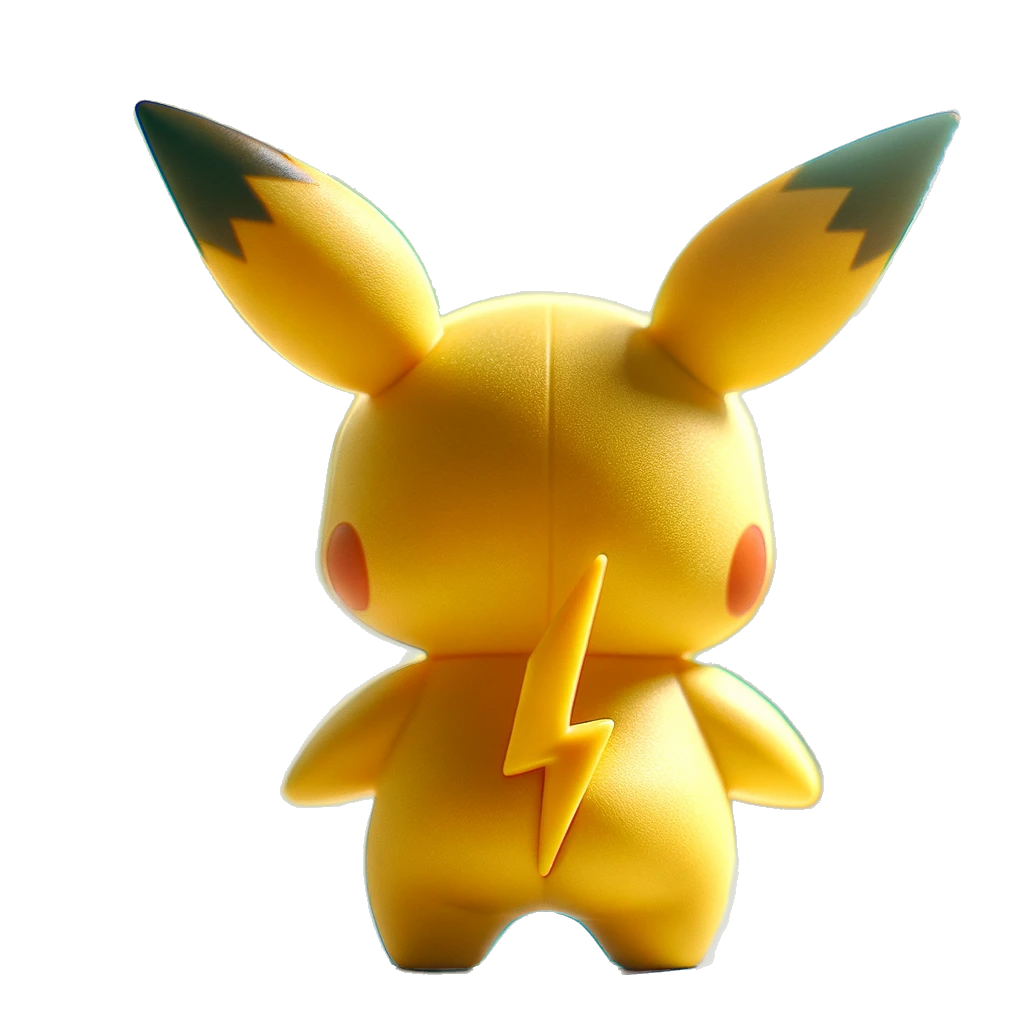} &
        \includegraphics[width=0.075\textwidth]{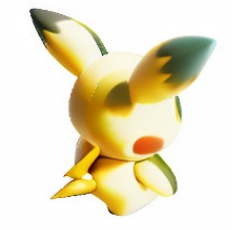} &
        \includegraphics[width=0.075\textwidth]{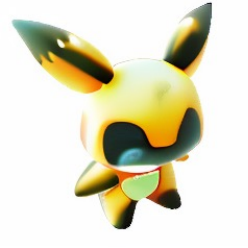} &
        \includegraphics[width=0.075\textwidth]{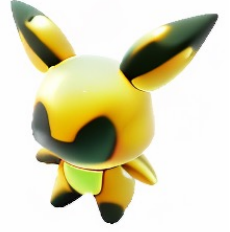} &
        \includegraphics[width=0.075\textwidth]{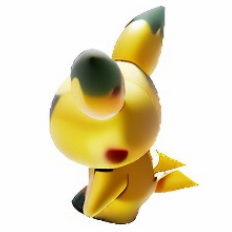} &
        \includegraphics[width=0.075\textwidth]{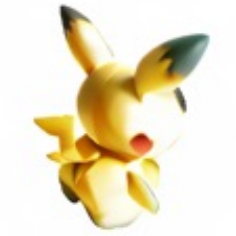} &
        \includegraphics[width=0.075\textwidth]{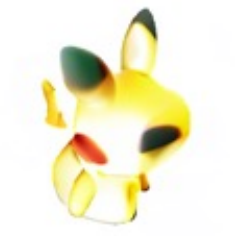} &
        \includegraphics[width=0.075\textwidth]{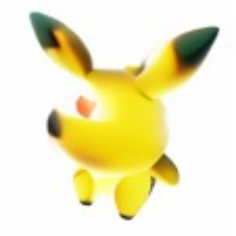} &
        \includegraphics[width=0.075\textwidth]{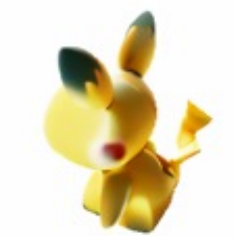} &
        \includegraphics[width=0.075\textwidth]{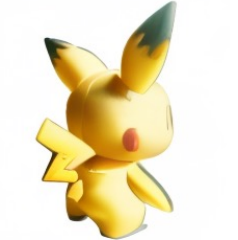} &
        \includegraphics[width=0.075\textwidth]{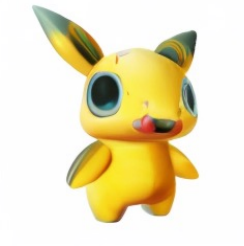} &
        \includegraphics[width=0.075\textwidth]{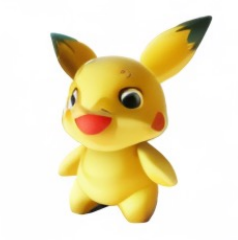} &
        \includegraphics[width=0.075\textwidth]{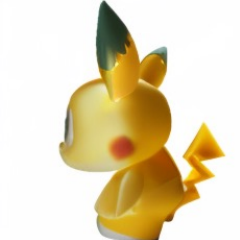} \\
        &
        \includegraphics[width=0.075\textwidth]{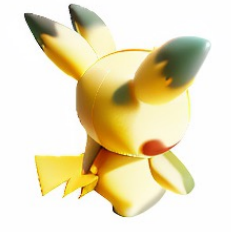} &
        \includegraphics[width=0.075\textwidth]{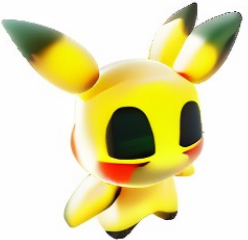} &
        \includegraphics[width=0.075\textwidth]{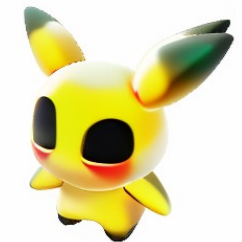} &
        \includegraphics[width=0.075\textwidth]{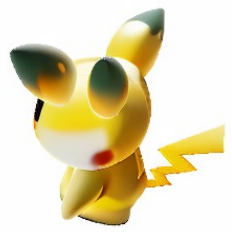} &
        \includegraphics[width=0.075\textwidth]{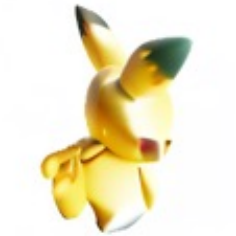} &
        \includegraphics[width=0.075\textwidth]{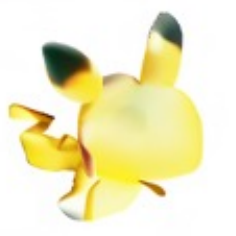} &
        \includegraphics[width=0.075\textwidth]{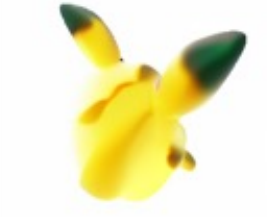} &
        \includegraphics[width=0.075\textwidth]{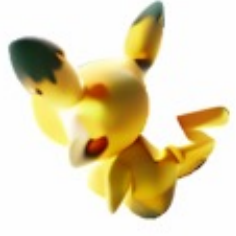} &
        \includegraphics[width=0.075\textwidth]{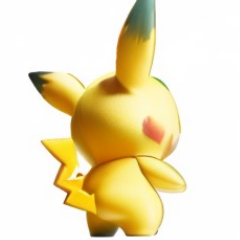} &
        \includegraphics[width=0.075\textwidth]{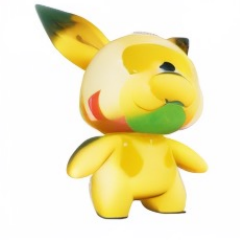} &
        \includegraphics[width=0.075\textwidth]{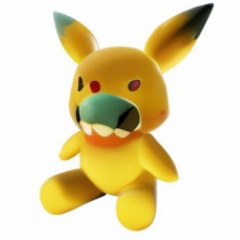} &
        \includegraphics[width=0.075\textwidth]{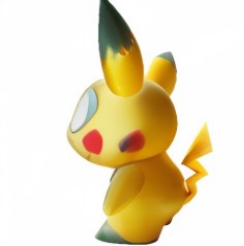} \\
        \multicolumn{1}{c}{Input} & \multicolumn{4}{c}{HarmonyView} & \multicolumn{4}{c}{SyncDreamer~\cite{liu2023syncdreamer}} & \multicolumn{4}{c}{Zero123~\cite{liu2023zero}} \\
    \end{tabular}
    }
    \vspace{\abovefigcapmargin}
    \caption{
        \textbf{Additional novel-view synthesis comparison.} HarmonyView creates diverse, coherent multi-view images for complex scenes, effortlessly generating realistic front views from rear-view input images.
    }
    \vspace{\belowfigcapmargin}
    \label{fig:appendix_nvs}
\end{figure*}

%% file: tables/statistical_analysis.tex
\begin{table*}[]
    \centering
    \resizebox{\linewidth}{!}{
    \begin{tabular}{lcccccccccccc}
       \toprule
       \multirow{2}{*}{Method}  & \multicolumn{3}{c}{PSNR$\uparrow$} & \multicolumn{3}{c}{SSIM$\uparrow$} & \multicolumn{3}{c}{LPIPS$\downarrow$} & \multicolumn{3}{c}{$E_{flow}$$\downarrow$} \\
       \arrayrulecolor{gray}\cmidrule(lr){2-4} \cmidrule(lr){5-7} \cmidrule(lr){8-10} \cmidrule(lr){11-13}
              & Best & Avg. & Var. & Best & Avg. & $\text{Var.}^*$ & Best & Avg. & $\text{Var.}^*$ & Best & Avg. & Var. \\
       \arrayrulecolor{black}\midrule
       Zero123~\cite{liu2023zero}    
       & 18.98 & 18.79 & 0.048 & 0.795 & 0.792 & 1.003 & 0.166 & 0.170 & 2.025 & 3.820 & 4.185 & 0.197 \\
       SyncDreamer~\cite{liu2023syncdreamer}   
       & 20.19 & 19.74 & 0.242 & 0.819 & 0.813 & 4.465 & 0.140 & 0.148 & 7.922 & 2.071 & 2.446 & 0.458 \\
       \rowcolor{gray!25}
       HarmonyView & \textbf{20.69} & \textbf{20.24} & \textbf{0.260} & \textbf{0.825} & \textbf{0.819} & \textbf{5.295} & \textbf{0.133} & \textbf{0.140} & \textbf{8.038} & \textbf{1.945} & \textbf{2.350} & \textbf{0.510} \\
       \arrayrulecolor{black}\bottomrule
    \end{tabular}
    }
    \vspace{\abovetabcapmargin}
    \caption{
        \textbf{Statistical analysis of novel-view synthesis on GSO~\cite{downs2022google} dataset.}
        We report PSNR, SSIM, LPIPS, and $E_{flow}$ for the best-matched instance with GT, as well as the average and variance across four instances. The variances marked as $*$ are reported with scaling by $10^{-5}$.
    }
    \vspace{\belowtabcapmargin}
    \label{tab:statistic_analysis}
\end{table*}

%% file: figs/appendix_qualitative_3d_recon.tex
\begin{figure*}
    \centering
    \setlength\tabcolsep{5pt}
    \resizebox{\linewidth}{!}{
    \renewcommand{\arraystretch}{0.5}
    \begin{tabular}{ccccccc}
        \includegraphics[width=0.13\textwidth]{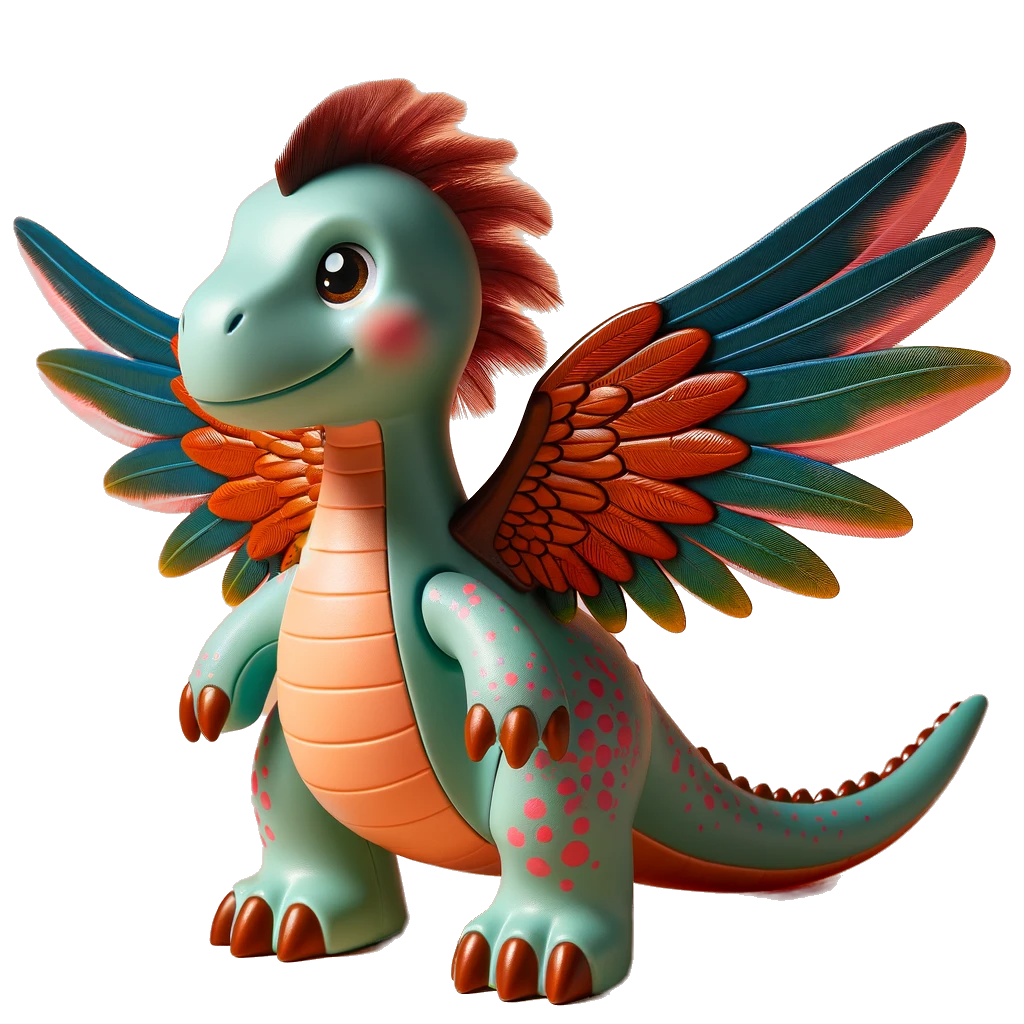} &
        \includegraphics[width=0.13\textwidth]{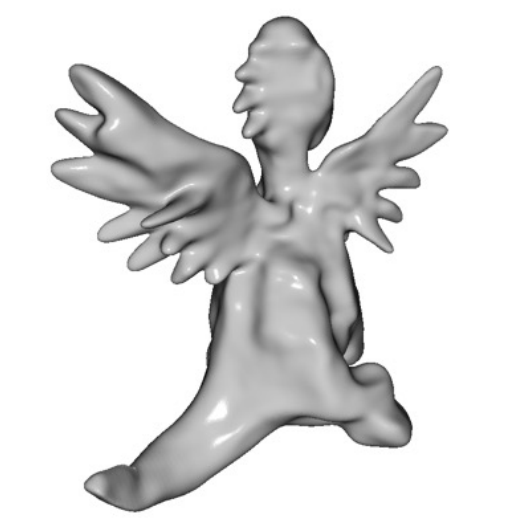} &
        \includegraphics[width=0.13\textwidth]{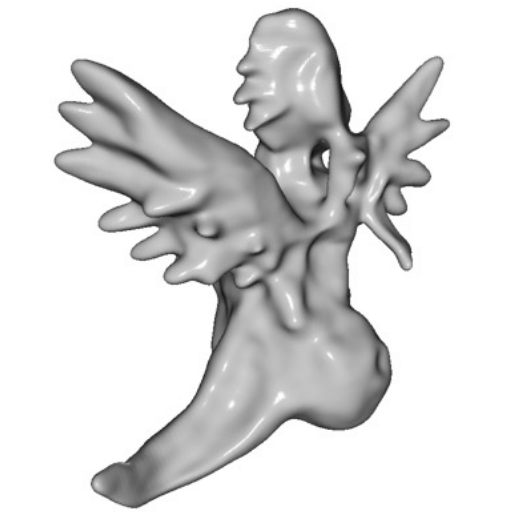} &
        \includegraphics[width=0.13\textwidth]{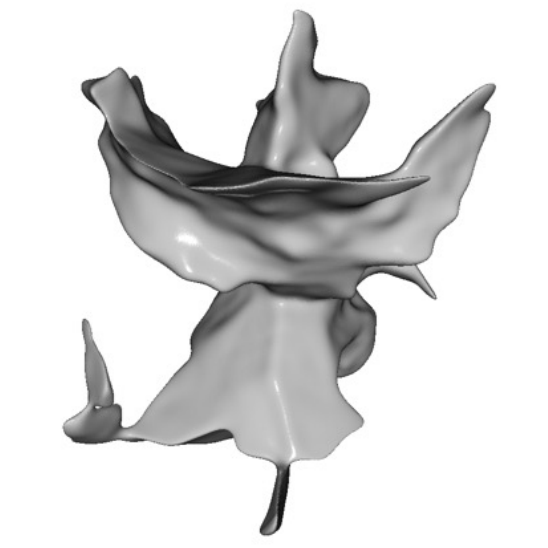} &
        \includegraphics[width=0.13\textwidth]{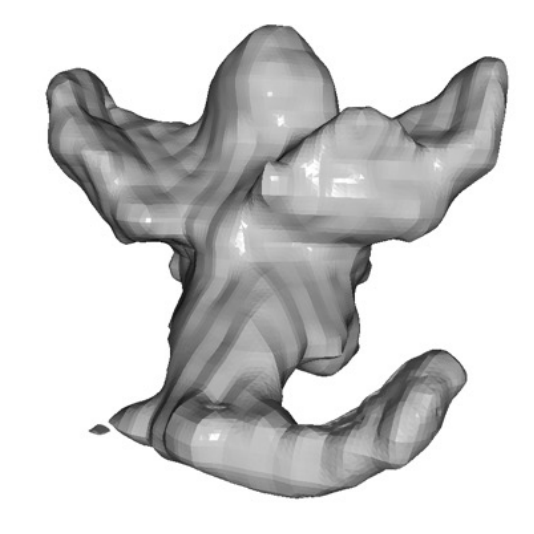} &
        \includegraphics[width=0.13\textwidth]{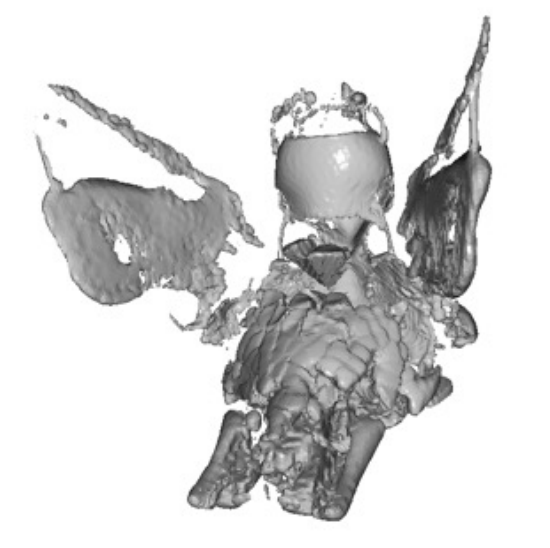} &
        \includegraphics[width=0.13\textwidth]{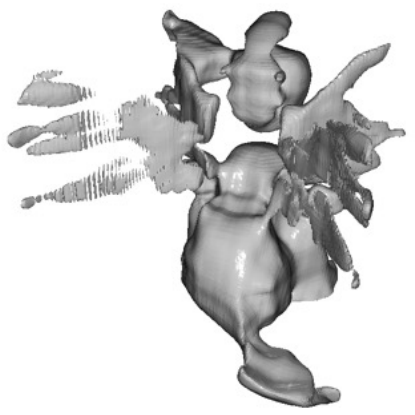} \\
        \includegraphics[width=0.13\textwidth]{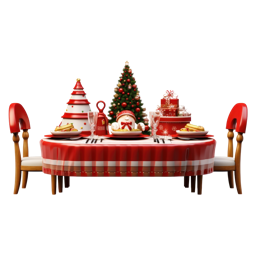} &
        \includegraphics[width=0.13\textwidth]{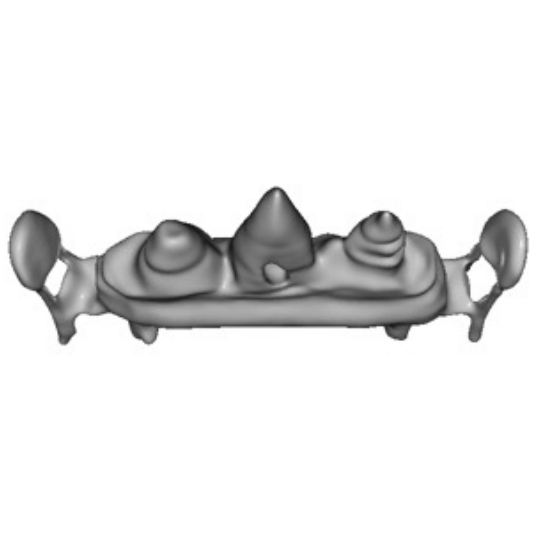} &
        \includegraphics[width=0.13\textwidth]{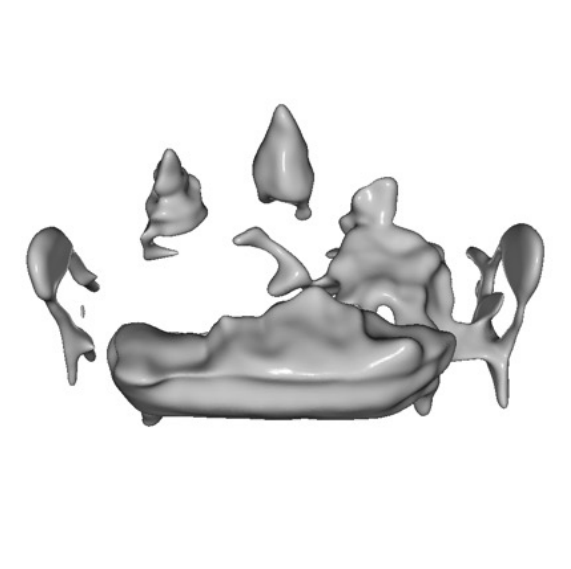} &
        \includegraphics[width=0.13\textwidth]{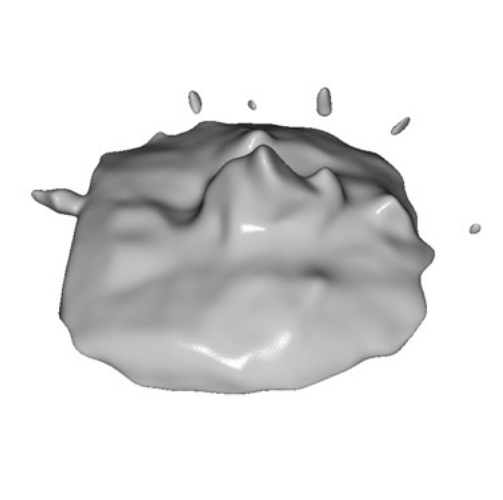} &
        \includegraphics[width=0.13\textwidth]{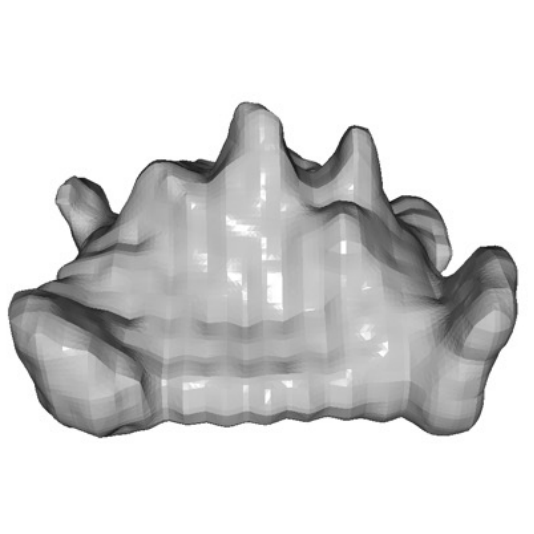} &
        \includegraphics[width=0.13\textwidth]{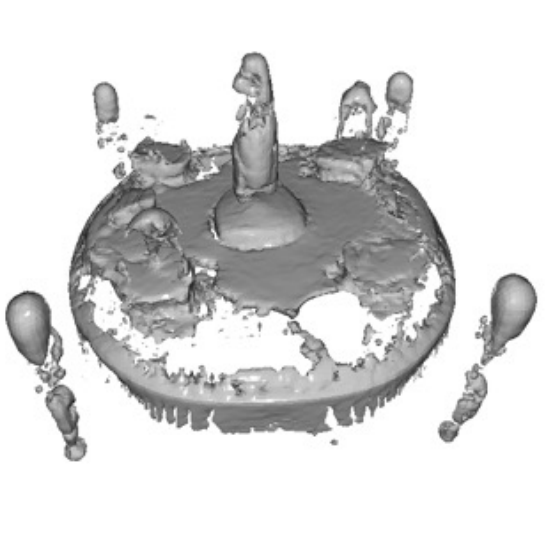} &
        \includegraphics[width=0.13\textwidth]{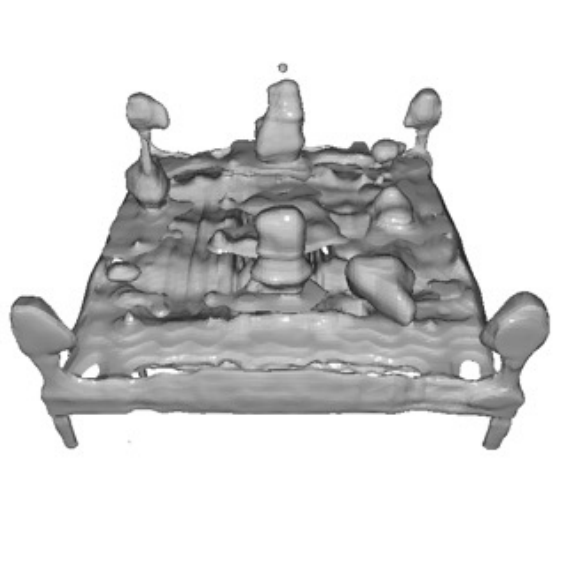} \\
        \includegraphics[width=0.13\textwidth]{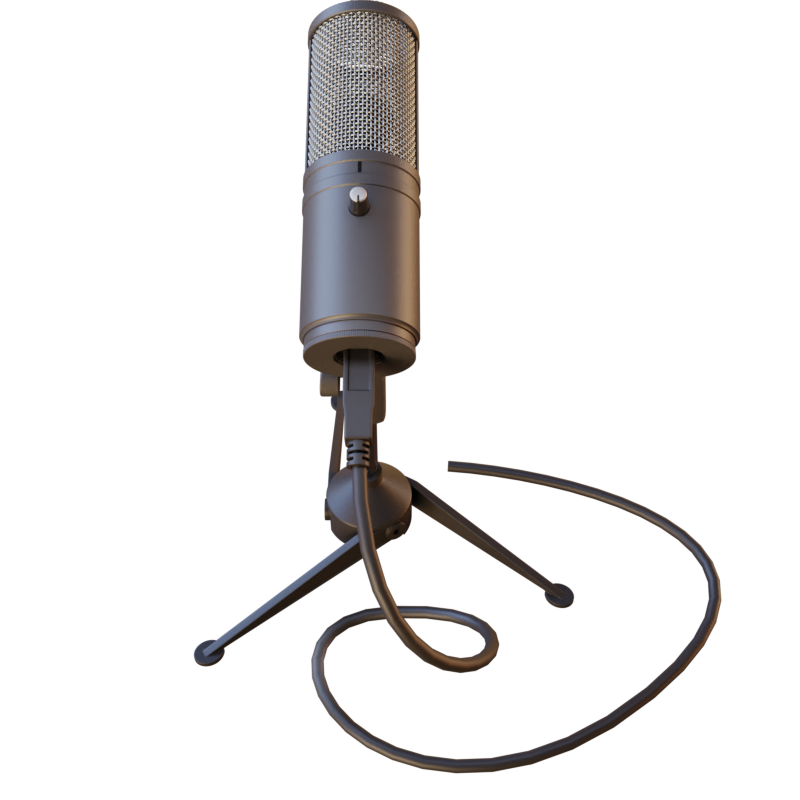} &
        \includegraphics[width=0.13\textwidth]{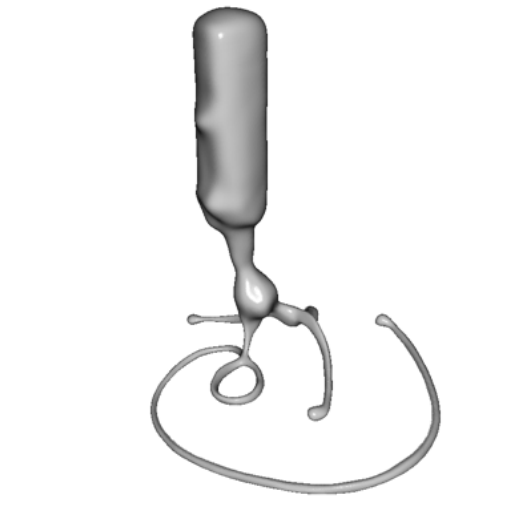} &
        \includegraphics[width=0.13\textwidth]{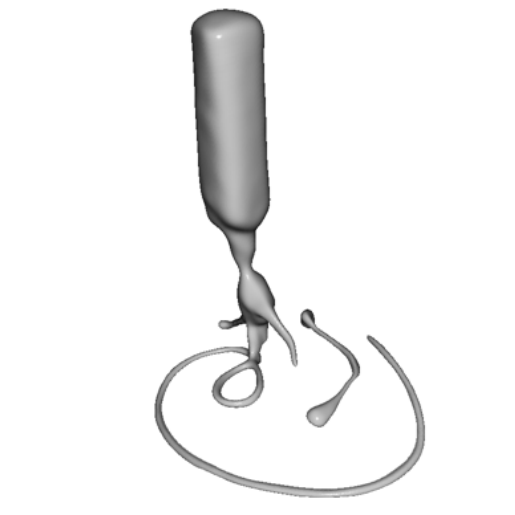} &
        \includegraphics[width=0.13\textwidth]{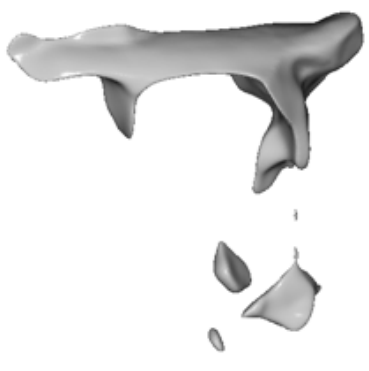} &
        \includegraphics[width=0.13\textwidth]{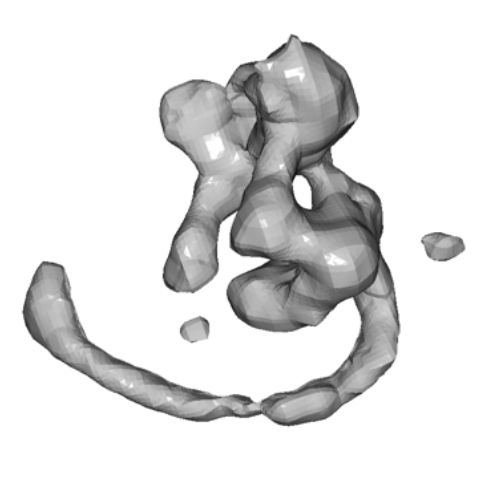} &
        \includegraphics[width=0.13\textwidth]{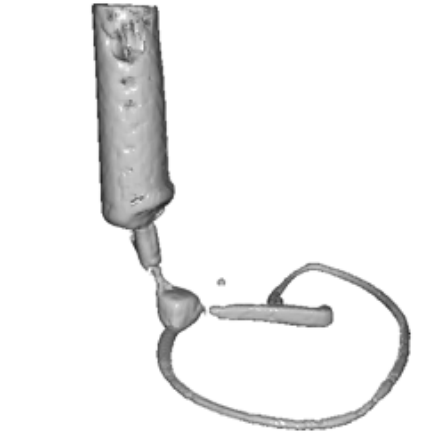} &
        \includegraphics[width=0.13\textwidth]{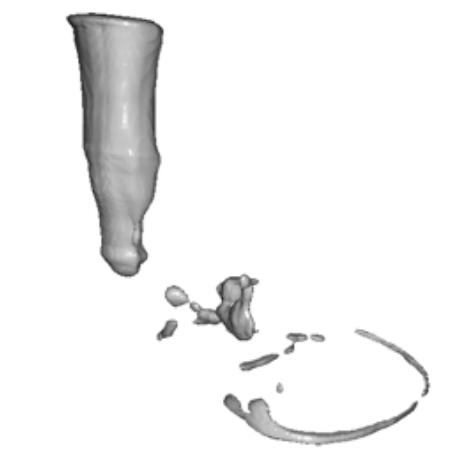} \\
        \includegraphics[width=0.13\textwidth]{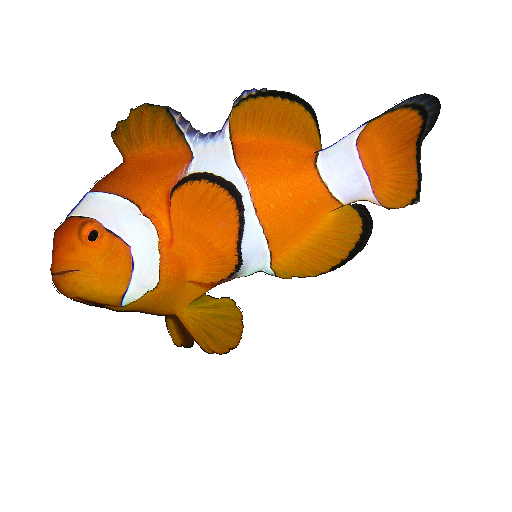} &
        \includegraphics[width=0.13\textwidth]{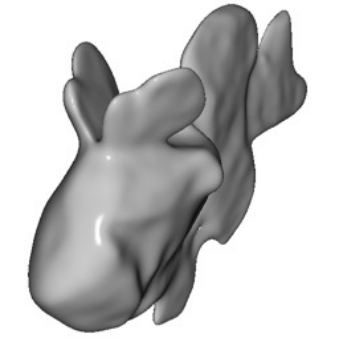} &
        \includegraphics[width=0.13\textwidth]{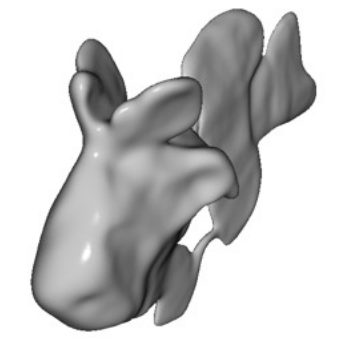} &
        \includegraphics[width=0.13\textwidth]{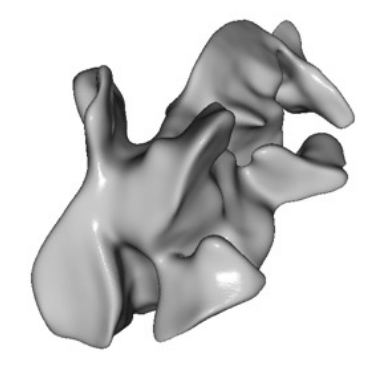} &
        \includegraphics[width=0.13\textwidth]{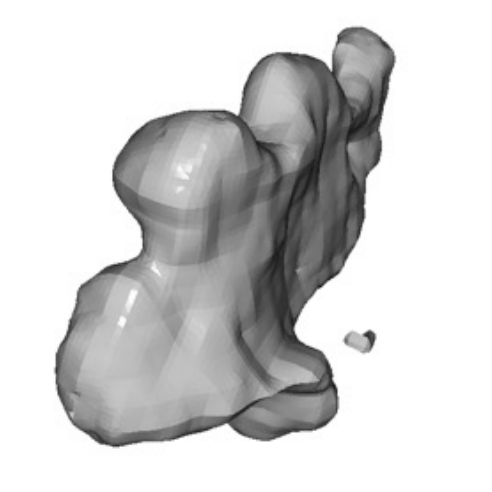} &
        \includegraphics[width=0.13\textwidth]{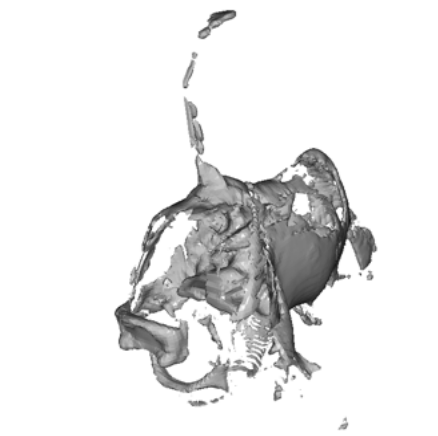} &
        \includegraphics[width=0.13\textwidth]{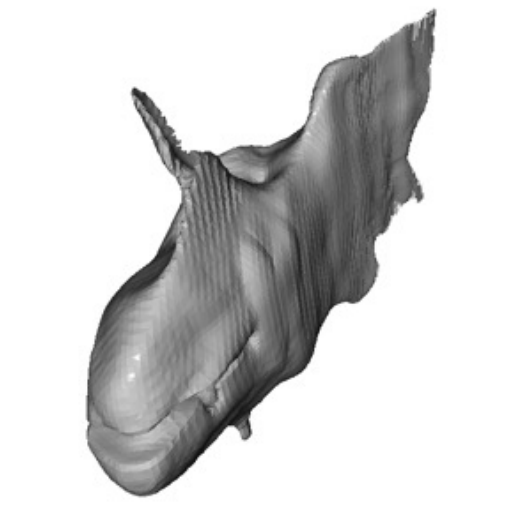} \\
        Input & HarmonyView & {\small SyncDreamer~\cite{liu2023syncdreamer}} & Zero123~\cite{liu2023zero} & One-2-3-45~\cite{liu2023one} & Point-E~\cite{nichol2022point}& Shap-E~\cite{jun2023shap} \\
    \end{tabular}
    }
    \vspace{\abovefigcapmargin}
    \caption{
        \textbf{Additional 3D reconstruction comparison.}
        HarmonyView excels at generating high-fidelity 3D meshes that achieve precise geometry with a realistic appearance while sidestepping common pitfalls for comprehensive and captivating reconstructions.
    }
    \vspace{\belowfigcapmargin}
    \label{fig:appendix_recon}
\end{figure*}

%% file: _main.bbl
\begin{thebibliography}{88}
\providecommand{\natexlab}[1]{#1}
\providecommand{\url}[1]{\texttt{#1}}
\expandafter\ifx\csname urlstyle\endcsname\relax
  \providecommand{\doi}[1]{doi: #1}\else
  \providecommand{\doi}{doi: \begingroup \urlstyle{rm}\Url}\fi

\bibitem[Chan et~al.(2023)Chan, Nagano, Chan, Bergman, Park, Levy, Aittala, De~Mello, Karras, and Wetzstein]{chan2023generative}
Eric~R Chan, Koki Nagano, Matthew~A Chan, Alexander~W Bergman, Jeong~Joon Park, Axel Levy, Miika Aittala, Shalini De~Mello, Tero Karras, and Gordon Wetzstein.
\newblock Generative novel view synthesis with 3d-aware diffusion models.
\newblock \emph{arXiv preprint arXiv:2304.02602}, 2023.

\bibitem[Chang et~al.(2015)Chang, Funkhouser, Guibas, Hanrahan, Huang, Li, Savarese, Savva, Song, Su, et~al.]{chang2015shapenet}
Angel~X Chang, Thomas Funkhouser, Leonidas Guibas, Pat Hanrahan, Qixing Huang, Zimo Li, Silvio Savarese, Manolis Savva, Shuran Song, Hao Su, et~al.
\newblock Shapenet: An information-rich 3d model repository.
\newblock \emph{arXiv preprint arXiv:1512.03012}, 2015.

\bibitem[Chen et~al.(2023)Chen, Chen, Jiao, and Jia]{chen2023fantasia3d}
Rui Chen, Yongwei Chen, Ningxin Jiao, and Kui Jia.
\newblock Fantasia3d: Disentangling geometry and appearance for high-quality text-to-3d content creation.
\newblock \emph{arXiv preprint arXiv:2303.13873}, 2023.

\bibitem[Chen(2023)]{chen2023review}
Zhiqin Chen.
\newblock A review of deep learning-powered mesh reconstruction methods.
\newblock \emph{arXiv preprint arXiv:2303.02879}, 2023.

\bibitem[Chen and Zhang(2019)]{chen2019imnet}
Zhiqin Chen and Hao Zhang.
\newblock Learning implicit fields for generative shape modeling.
\newblock In \emph{Proceedings of the IEEE/CVF Conference on Computer Vision and Pattern Recognition}, pages 5939--5948, 2019.

\bibitem[Chen and Zhang(2021)]{chen2021neural}
Zhiqin Chen and Hao Zhang.
\newblock Neural marching cubes.
\newblock \emph{ACM Transactions on Graphics (TOG)}, 40\penalty0 (6):\penalty0 1--15, 2021.

\bibitem[Chen et~al.(2020)Chen, Tagliasacchi, and Zhang]{chen2020bsp}
Zhiqin Chen, Andrea Tagliasacchi, and Hao Zhang.
\newblock Bsp-net: Generating compact meshes via binary space partitioning.
\newblock In \emph{Proceedings of the IEEE/CVF Conference on Computer Vision and Pattern Recognition}, pages 45--54, 2020.

\bibitem[Cheng et~al.(2023)Cheng, Lee, Tulyakov, Schwing, and Gui]{cheng2023sdfusion}
Yen-Chi Cheng, Hsin-Ying Lee, Sergey Tulyakov, Alexander~G Schwing, and Liang-Yan Gui.
\newblock Sdfusion: Multimodal 3d shape completion, reconstruction, and generation.
\newblock In \emph{Proceedings of the IEEE/CVF Conference on Computer Vision and Pattern Recognition}, pages 4456--4465, 2023.

\bibitem[Choy et~al.(2016)Choy, Xu, Gwak, Chen, and Savarese]{choy20163dr2d2}
Christopher~B Choy, Danfei Xu, JunYoung Gwak, Kevin Chen, and Silvio Savarese.
\newblock 3d-r2n2: A unified approach for single and multi-view 3d object reconstruction.
\newblock In \emph{Computer Vision--ECCV 2016: 14th European Conference, Amsterdam, The Netherlands, October 11-14, 2016, Proceedings, Part VIII 14}, pages 628--644. Springer, 2016.

\bibitem[Deng et~al.(2020)Deng, Genova, Yazdani, Bouaziz, Hinton, and Tagliasacchi]{deng2020cvxnet}
Boyang Deng, Kyle Genova, Soroosh Yazdani, Sofien Bouaziz, Geoffrey Hinton, and Andrea Tagliasacchi.
\newblock Cvxnet: Learnable convex decomposition.
\newblock In \emph{Proceedings of the IEEE/CVF Conference on Computer Vision and Pattern Recognition}, pages 31--44, 2020.

\bibitem[Dey and Boddeti(2022)]{dey2022generating}
Rahul Dey and Vishnu~Naresh Boddeti.
\newblock Generating diverse 3d reconstructions from a single occluded face image.
\newblock In \emph{Proceedings of the IEEE/CVF Conference on Computer Vision and Pattern Recognition}, pages 1547--1557, 2022.

\bibitem[Dhariwal and Nichol(2021)]{dhariwal2021diffusion}
Prafulla Dhariwal and Alexander Nichol.
\newblock Diffusion models beat gans on image synthesis.
\newblock \emph{Advances in neural information processing systems}, 34:\penalty0 8780--8794, 2021.

\bibitem[Downs et~al.(2022)Downs, Francis, Koenig, Kinman, Hickman, Reymann, McHugh, and Vanhoucke]{downs2022google}
Laura Downs, Anthony Francis, Nate Koenig, Brandon Kinman, Ryan Hickman, Krista Reymann, Thomas~B McHugh, and Vincent Vanhoucke.
\newblock Google scanned objects: A high-quality dataset of 3d scanned household items.
\newblock In \emph{2022 International Conference on Robotics and Automation (ICRA)}, pages 2553--2560. IEEE, 2022.

\bibitem[Fan et~al.(2017)Fan, Su, and Guibas]{fan2017point}
Haoqiang Fan, Hao Su, and Leonidas~J Guibas.
\newblock A point set generation network for 3d object reconstruction from a single image.
\newblock In \emph{Proceedings of the IEEE conference on computer vision and pattern recognition}, pages 605--613, 2017.

\bibitem[Gal et~al.(2022)Gal, Alaluf, Atzmon, Patashnik, Bermano, Chechik, and Cohen-Or]{gal2022image}
Rinon Gal, Yuval Alaluf, Yuval Atzmon, Or Patashnik, Amit~H Bermano, Gal Chechik, and Daniel Cohen-Or.
\newblock An image is worth one word: Personalizing text-to-image generation using textual inversion.
\newblock \emph{arXiv preprint arXiv:2208.01618}, 2022.

\bibitem[Genova et~al.(2020)Genova, Cole, Sud, Sarna, and Funkhouser]{genova2020local}
Kyle Genova, Forrester Cole, Avneesh Sud, Aaron Sarna, and Thomas Funkhouser.
\newblock Local deep implicit functions for 3d shape.
\newblock In \emph{Proceedings of the IEEE/CVF Conference on Computer Vision and Pattern Recognition}, pages 4857--4866, 2020.

\bibitem[Gkioxari et~al.(2019)Gkioxari, Malik, and Johnson]{gkioxari2019meshrcnn}
Georgia Gkioxari, Jitendra Malik, and Justin Johnson.
\newblock Mesh r-cnn.
\newblock In \emph{Proceedings of the IEEE/CVF international conference on computer vision}, pages 9785--9795, 2019.

\bibitem[Groueix et~al.(2018)Groueix, Fisher, Kim, Russell, and Aubry]{groueix2018papier}
Thibault Groueix, Matthew Fisher, Vladimir~G Kim, Bryan~C Russell, and Mathieu Aubry.
\newblock A papier-m{\^a}ch{\'e} approach to learning 3d surface generation.
\newblock In \emph{Proceedings of the IEEE conference on computer vision and pattern recognition}, pages 216--224, 2018.

\bibitem[Gupta et~al.(2023)Gupta, Xiong, Nie, Jones, and O{\u{g}}uz]{gupta20233dgen}
Anchit Gupta, Wenhan Xiong, Yixin Nie, Ian Jones, and Barlas O{\u{g}}uz.
\newblock 3dgen: Triplane latent diffusion for textured mesh generation.
\newblock \emph{arXiv preprint arXiv:2303.05371}, 2023.

\bibitem[Hessel et~al.(2021)Hessel, Holtzman, Forbes, Bras, and Choi]{hessel2021clipscore}
Jack Hessel, Ari Holtzman, Maxwell Forbes, Ronan~Le Bras, and Yejin Choi.
\newblock Clipscore: A reference-free evaluation metric for image captioning.
\newblock \emph{arXiv preprint arXiv:2104.08718}, 2021.

\bibitem[Ho and Salimans(2022)]{ho2022classifier}
Jonathan Ho and Tim Salimans.
\newblock Classifier-free diffusion guidance.
\newblock \emph{arXiv preprint arXiv:2207.12598}, 2022.

\bibitem[Ho et~al.(2020)Ho, Jain, and Abbeel]{ho2020denoising}
Jonathan Ho, Ajay Jain, and Pieter Abbeel.
\newblock Denoising diffusion probabilistic models.
\newblock \emph{Advances in neural information processing systems}, 33:\penalty0 6840--6851, 2020.

\bibitem[Jain et~al.(2022)Jain, Mildenhall, Barron, Abbeel, and Poole]{jain2022zero}
Ajay Jain, Ben Mildenhall, Jonathan~T Barron, Pieter Abbeel, and Ben Poole.
\newblock Zero-shot text-guided object generation with dream fields.
\newblock In \emph{Proceedings of the IEEE/CVF Conference on Computer Vision and Pattern Recognition}, pages 867--876, 2022.

\bibitem[Jampani et~al.(2021)Jampani, Chang, Sargent, Kar, Tucker, Krainin, Kaeser, Freeman, Salesin, Curless, et~al.]{jampani2021slide}
Varun Jampani, Huiwen Chang, Kyle Sargent, Abhishek Kar, Richard Tucker, Michael Krainin, Dominik Kaeser, William~T Freeman, David Salesin, Brian Curless, et~al.
\newblock Slide: Single image 3d photography with soft layering and depth-aware inpainting.
\newblock In \emph{Proceedings of the IEEE/CVF International Conference on Computer Vision}, pages 12518--12527, 2021.

\bibitem[Jiang et~al.(2023)Jiang, Tang, Chang, Song, Wang, and Cao]{jiang2023efficient}
Yifan Jiang, Hao Tang, Jen-Hao~Rick Chang, Liangchen Song, Zhangyang Wang, and Liangliang Cao.
\newblock Efficient-3dim: Learning a generalizable single-image novel-view synthesizer in one day.
\newblock \emph{arXiv preprint arXiv:2310.03015}, 2023.

\bibitem[Jun and Nichol(2023)]{jun2023shap}
Heewoo Jun and Alex Nichol.
\newblock Shap-e: Generating conditional 3d implicit functions.
\newblock \emph{arXiv preprint arXiv:2305.02463}, 2023.

\bibitem[Kant et~al.(2023)Kant, Siarohin, Vasilkovsky, Guler, Ren, Tulyakov, and Gilitschenski]{kant2023invs}
Yash Kant, Aliaksandr Siarohin, Michael Vasilkovsky, Riza~Alp Guler, Jian Ren, Sergey Tulyakov, and Igor Gilitschenski.
\newblock invs: Repurposing diffusion inpainters for novel view synthesis.
\newblock \emph{arXiv preprint arXiv:2310.16167}, 2023.

\bibitem[Liao et~al.(2018)Liao, Donne, and Geiger]{liao2018deep}
Yiyi Liao, Simon Donne, and Andreas Geiger.
\newblock Deep marching cubes: Learning explicit surface representations.
\newblock In \emph{Proceedings of the IEEE Conference on Computer Vision and Pattern Recognition}, pages 2916--2925, 2018.

\bibitem[Lin et~al.(2023{\natexlab{a}})Lin, Gao, Tang, Takikawa, Zeng, Huang, Kreis, Fidler, Liu, and Lin]{lin2023magic3d}
Chen-Hsuan Lin, Jun Gao, Luming Tang, Towaki Takikawa, Xiaohui Zeng, Xun Huang, Karsten Kreis, Sanja Fidler, Ming-Yu Liu, and Tsung-Yi Lin.
\newblock Magic3d: High-resolution text-to-3d content creation.
\newblock In \emph{Proceedings of the IEEE/CVF Conference on Computer Vision and Pattern Recognition}, pages 300--309, 2023{\natexlab{a}}.

\bibitem[Lin et~al.(2023{\natexlab{b}})Lin, Han, Gong, Xu, Zhang, and Li]{lin2023consistent123}
Yukang Lin, Haonan Han, Chaoqun Gong, Zunnan Xu, Yachao Zhang, and Xiu Li.
\newblock Consistent123: One image to highly consistent 3d asset using case-aware diffusion priors.
\newblock \emph{arXiv preprint arXiv:2309.17261}, 2023{\natexlab{b}}.

\bibitem[Liu et~al.(2023{\natexlab{a}})Liu, Xu, Jin, Chen, Xu, Su, et~al.]{liu2023one}
Minghua Liu, Chao Xu, Haian Jin, Linghao Chen, Zexiang Xu, Hao Su, et~al.
\newblock One-2-3-45: Any single image to 3d mesh in 45 seconds without per-shape optimization.
\newblock \emph{arXiv preprint arXiv:2306.16928}, 2023{\natexlab{a}}.

\bibitem[Liu et~al.(2023{\natexlab{b}})Liu, Wu, Van~Hoorick, Tokmakov, Zakharov, and Vondrick]{liu2023zero}
Ruoshi Liu, Rundi Wu, Basile Van~Hoorick, Pavel Tokmakov, Sergey Zakharov, and Carl Vondrick.
\newblock Zero-1-to-3: Zero-shot one image to 3d object.
\newblock In \emph{Proceedings of the IEEE/CVF International Conference on Computer Vision}, pages 9298--9309, 2023{\natexlab{b}}.

\bibitem[Liu et~al.(2023{\natexlab{c}})Liu, Lin, Zeng, Long, Liu, Komura, and Wang]{liu2023syncdreamer}
Yuan Liu, Cheng Lin, Zijiao Zeng, Xiaoxiao Long, Lingjie Liu, Taku Komura, and Wenping Wang.
\newblock Syncdreamer: Generating multiview-consistent images from a single-view image.
\newblock \emph{arXiv preprint arXiv:2309.03453}, 2023{\natexlab{c}}.

\bibitem[Liu et~al.(2023{\natexlab{d}})Liu, Feng, Black, Nowrouzezahrai, Paull, and Liu]{liu2023meshdiffusion}
Zhen Liu, Yao Feng, Michael~J Black, Derek Nowrouzezahrai, Liam Paull, and Weiyang Liu.
\newblock Meshdiffusion: Score-based generative 3d mesh modeling.
\newblock \emph{arXiv preprint arXiv:2303.08133}, 2023{\natexlab{d}}.

\bibitem[Long et~al.(2023)Long, Guo, Lin, Liu, Dou, Liu, Ma, Zhang, Habermann, Theobalt, et~al.]{long2023wonder3d}
Xiaoxiao Long, Yuan-Chen Guo, Cheng Lin, Yuan Liu, Zhiyang Dou, Lingjie Liu, Yuexin Ma, Song-Hai Zhang, Marc Habermann, Christian Theobalt, et~al.
\newblock Wonder3d: Single image to 3d using cross-domain diffusion.
\newblock \emph{arXiv preprint arXiv:2310.15008}, 2023.

\bibitem[Lorraine et~al.(2023)Lorraine, Xie, Zeng, Lin, Takikawa, Sharp, Lin, Liu, Fidler, and Lucas]{lorraine2023att3d}
Jonathan Lorraine, Kevin Xie, Xiaohui Zeng, Chen-Hsuan Lin, Towaki Takikawa, Nicholas Sharp, Tsung-Yi Lin, Ming-Yu Liu, Sanja Fidler, and James Lucas.
\newblock Att3d: Amortized text-to-3d object synthesis.
\newblock \emph{arXiv preprint arXiv:2306.07349}, 2023.

\bibitem[Melas-Kyriazi et~al.(2023{\natexlab{a}})Melas-Kyriazi, Laina, Rupprecht, and Vedaldi]{melas2023realfusion}
Luke Melas-Kyriazi, Iro Laina, Christian Rupprecht, and Andrea Vedaldi.
\newblock Realfusion: 360deg reconstruction of any object from a single image.
\newblock In \emph{Proceedings of the IEEE/CVF Conference on Computer Vision and Pattern Recognition}, pages 8446--8455, 2023{\natexlab{a}}.

\bibitem[Melas-Kyriazi et~al.(2023{\natexlab{b}})Melas-Kyriazi, Rupprecht, and Vedaldi]{melas2023pc2}
Luke Melas-Kyriazi, Christian Rupprecht, and Andrea Vedaldi.
\newblock Pc2: Projection-conditioned point cloud diffusion for single-image 3d reconstruction.
\newblock In \emph{Proceedings of the IEEE/CVF Conference on Computer Vision and Pattern Recognition}, pages 12923--12932, 2023{\natexlab{b}}.

\bibitem[Mildenhall et~al.(2021)Mildenhall, Srinivasan, Tancik, Barron, Ramamoorthi, and Ng]{mildenhall2021nerf}
Ben Mildenhall, Pratul~P Srinivasan, Matthew Tancik, Jonathan~T Barron, Ravi Ramamoorthi, and Ren Ng.
\newblock Nerf: Representing scenes as neural radiance fields for view synthesis.
\newblock \emph{Communications of the ACM}, 65\penalty0 (1):\penalty0 99--106, 2021.

\bibitem[Mu et~al.(2022)Mu, Wang, Wu, and Li]{mu20223d}
Fangzhou Mu, Jian Wang, Yicheng Wu, and Yin Li.
\newblock 3d photo stylization: Learning to generate stylized novel views from a single image.
\newblock In \emph{Proceedings of the IEEE/CVF Conference on Computer Vision and Pattern Recognition}, pages 16273--16282, 2022.

\bibitem[Nichol et~al.(2021)Nichol, Dhariwal, Ramesh, Shyam, Mishkin, McGrew, Sutskever, and Chen]{nichol2021glide}
Alex Nichol, Prafulla Dhariwal, Aditya Ramesh, Pranav Shyam, Pamela Mishkin, Bob McGrew, Ilya Sutskever, and Mark Chen.
\newblock Glide: Towards photorealistic image generation and editing with text-guided diffusion models.
\newblock \emph{arXiv preprint arXiv:2112.10741}, 2021.

\bibitem[Nichol et~al.(2022)Nichol, Jun, Dhariwal, Mishkin, and Chen]{nichol2022point}
Alex Nichol, Heewoo Jun, Prafulla Dhariwal, Pamela Mishkin, and Mark Chen.
\newblock Point-e: A system for generating 3d point clouds from complex prompts.
\newblock \emph{arXiv preprint arXiv:2212.08751}, 2022.

\bibitem[Park et~al.(2022)Park, Go, and Kim]{park2022bridging}
Byeongjun Park, Hyojun Go, and Changick Kim.
\newblock Bridging implicit and explicit geometric transformations for single-image view synthesis.
\newblock \emph{arXiv preprint arXiv:2209.07105}, 2022.

\bibitem[Poole et~al.(2022)Poole, Jain, Barron, and Mildenhall]{poole2022dreamfusion}
Ben Poole, Ajay Jain, Jonathan~T Barron, and Ben Mildenhall.
\newblock Dreamfusion: Text-to-3d using 2d diffusion.
\newblock \emph{arXiv preprint arXiv:2209.14988}, 2022.

\bibitem[Prince et~al.(2002)Prince, Cheok, Farbiz, Williamson, Johnson, Billinghurst, and Kato]{prince20023d}
Simon Prince, Adrian~David Cheok, Farzam Farbiz, Todd Williamson, Nikolas Johnson, Mark Billinghurst, and Hirokazu Kato.
\newblock 3d live: Real time captured content for mixed reality.
\newblock In \emph{Proceedings. International Symposium on Mixed and Augmented Reality}, pages 7--317. IEEE, 2002.

\bibitem[Qian et~al.(2023)Qian, Mai, Hamdi, Ren, Siarohin, Li, Lee, Skorokhodov, Wonka, Tulyakov, et~al.]{qian2023magic123}
Guocheng Qian, Jinjie Mai, Abdullah Hamdi, Jian Ren, Aliaksandr Siarohin, Bing Li, Hsin-Ying Lee, Ivan Skorokhodov, Peter Wonka, Sergey Tulyakov, et~al.
\newblock Magic123: One image to high-quality 3d object generation using both 2d and 3d diffusion priors.
\newblock \emph{arXiv preprint arXiv:2306.17843}, 2023.

\bibitem[Radford et~al.(2021)Radford, Kim, Hallacy, Ramesh, Goh, Agarwal, Sastry, Askell, Mishkin, Clark, et~al.]{radford2021learning}
Alec Radford, Jong~Wook Kim, Chris Hallacy, Aditya Ramesh, Gabriel Goh, Sandhini Agarwal, Girish Sastry, Amanda Askell, Pamela Mishkin, Jack Clark, et~al.
\newblock Learning transferable visual models from natural language supervision.
\newblock In \emph{International conference on machine learning}, pages 8748--8763. PMLR, 2021.

\bibitem[Ramesh et~al.(2021)Ramesh, Pavlov, Goh, Gray, Voss, Radford, Chen, and Sutskever]{ramesh2021zero}
Aditya Ramesh, Mikhail Pavlov, Gabriel Goh, Scott Gray, Chelsea Voss, Alec Radford, Mark Chen, and Ilya Sutskever.
\newblock Zero-shot text-to-image generation.
\newblock In \emph{International Conference on Machine Learning}, pages 8821--8831. PMLR, 2021.

\bibitem[Ranftl et~al.(2020)Ranftl, Lasinger, Hafner, Schindler, and Koltun]{ranftl2020towards}
Ren{\'e} Ranftl, Katrin Lasinger, David Hafner, Konrad Schindler, and Vladlen Koltun.
\newblock Towards robust monocular depth estimation: Mixing datasets for zero-shot cross-dataset transfer.
\newblock \emph{IEEE transactions on pattern analysis and machine intelligence}, 2020.

\bibitem[Rombach et~al.(2022)Rombach, Blattmann, Lorenz, Esser, and Ommer]{rombach2022high}
Robin Rombach, Andreas Blattmann, Dominik Lorenz, Patrick Esser, and Bj{\"o}rn Ommer.
\newblock High-resolution image synthesis with latent diffusion models.
\newblock In \emph{Proceedings of the IEEE/CVF Conference on Computer Vision and Pattern Recognition}, pages 10684--10695, 2022.

\bibitem[Sargent et~al.(2023)Sargent, Li, Shah, Herrmann, Yu, Zhang, Chan, Lagun, Fei-Fei, Sun, et~al.]{sargent2023zeronvs}
Kyle Sargent, Zizhang Li, Tanmay Shah, Charles Herrmann, Hong-Xing Yu, Yunzhi Zhang, Eric~Ryan Chan, Dmitry Lagun, Li Fei-Fei, Deqing Sun, et~al.
\newblock Zeronvs: Zero-shot 360-degree view synthesis from a single real image.
\newblock \emph{arXiv preprint arXiv:2310.17994}, 2023.

\bibitem[Sharma et~al.(2020)Sharma, Liu, Maji, Kalogerakis, Chaudhuri, and M{\v{e}}ch]{sharma2020parsenet}
Gopal Sharma, Difan Liu, Subhransu Maji, Evangelos Kalogerakis, Siddhartha Chaudhuri, and Radom{\'\i}r M{\v{e}}ch.
\newblock Parsenet: A parametric surface fitting network for 3d point clouds.
\newblock In \emph{Computer Vision--ECCV 2020: 16th European Conference, Glasgow, UK, August 23--28, 2020, Proceedings, Part VII 16}, pages 261--276. Springer, 2020.

\bibitem[Shi et~al.(2023{\natexlab{a}})Shi, Chen, Zhang, Liu, Xu, Wei, Chen, Zeng, and Su]{shi2023zero123++}
Ruoxi Shi, Hansheng Chen, Zhuoyang Zhang, Minghua Liu, Chao Xu, Xinyue Wei, Linghao Chen, Chong Zeng, and Hao Su.
\newblock Zero123++: a single image to consistent multi-view diffusion base model.
\newblock \emph{arXiv preprint arXiv:2310.15110}, 2023{\natexlab{a}}.

\bibitem[Shi et~al.(2023{\natexlab{b}})Shi, Wang, Cao, Tang, Qi, Yang, Huang, Liu, Zhang, and Shum]{shi2023toss}
Yukai Shi, Jianan Wang, He Cao, Boshi Tang, Xianbiao Qi, Tianyu Yang, Yukun Huang, Shilong Liu, Lei Zhang, and Heung-Yeung Shum.
\newblock Toss: High-quality text-guided novel view synthesis from a single image.
\newblock \emph{arXiv preprint arXiv:2310.10644}, 2023{\natexlab{b}}.

\bibitem[Shi et~al.(2023{\natexlab{c}})Shi, Wang, Ye, Long, Li, and Yang]{shi2023mvdream}
Yichun Shi, Peng Wang, Jianglong Ye, Mai Long, Kejie Li, and Xiao Yang.
\newblock Mvdream: Multi-view diffusion for 3d generation.
\newblock \emph{arXiv preprint arXiv:2308.16512}, 2023{\natexlab{c}}.

\bibitem[Shih et~al.(2020)Shih, Su, Kopf, and Huang]{shih20203d}
Meng-Li Shih, Shih-Yang Su, Johannes Kopf, and Jia-Bin Huang.
\newblock 3d photography using context-aware layered depth inpainting.
\newblock In \emph{Proceedings of the IEEE/CVF Conference on Computer Vision and Pattern Recognition}, pages 8028--8038, 2020.

\bibitem[Sinha et~al.(2017)Sinha, Unmesh, Huang, and Ramani]{sinha2017surfnet}
Ayan Sinha, Asim Unmesh, Qixing Huang, and Karthik Ramani.
\newblock Surfnet: Generating 3d shape surfaces using deep residual networks.
\newblock In \emph{Proceedings of the IEEE conference on computer vision and pattern recognition}, pages 6040--6049, 2017.

\bibitem[Sohl-Dickstein et~al.(2015)Sohl-Dickstein, Weiss, Maheswaranathan, and Ganguli]{sohl2015deep}
Jascha Sohl-Dickstein, Eric Weiss, Niru Maheswaranathan, and Surya Ganguli.
\newblock Deep unsupervised learning using nonequilibrium thermodynamics.
\newblock In \emph{International conference on machine learning}, pages 2256--2265. PMLR, 2015.

\bibitem[Song et~al.(2020)Song, Meng, and Ermon]{song2020denoising}
Jiaming Song, Chenlin Meng, and Stefano Ermon.
\newblock Denoising diffusion implicit models.
\newblock \emph{arXiv preprint arXiv:2010.02502}, 2020.

\bibitem[Sun et~al.(2018)Sun, Wu, Zhang, Zhang, Zhang, Xue, Tenenbaum, and Freeman]{sun2018pix3d}
Xingyuan Sun, Jiajun Wu, Xiuming Zhang, Zhoutong Zhang, Chengkai Zhang, Tianfan Xue, Joshua~B Tenenbaum, and William~T Freeman.
\newblock Pix3d: Dataset and methods for single-image 3d shape modeling.
\newblock In \emph{Proceedings of the IEEE conference on computer vision and pattern recognition}, pages 2974--2983, 2018.

\bibitem[Szymanowicz et~al.(2023)Szymanowicz, Rupprecht, and Vedaldi]{szymanowicz2023viewset}
Stanislaw Szymanowicz, Christian Rupprecht, and Andrea Vedaldi.
\newblock Viewset diffusion:(0-) image-conditioned 3d generative models from 2d data.
\newblock \emph{arXiv preprint arXiv:2306.07881}, 2023.

\bibitem[Tang et~al.(2023{\natexlab{a}})Tang, Wang, Zhang, Zhang, Yi, Ma, and Chen]{tang2023make}
Junshu Tang, Tengfei Wang, Bo Zhang, Ting Zhang, Ran Yi, Lizhuang Ma, and Dong Chen.
\newblock Make-it-3d: High-fidelity 3d creation from a single image with diffusion prior.
\newblock \emph{arXiv preprint arXiv:2303.14184}, 2023{\natexlab{a}}.

\bibitem[Tang et~al.(2023{\natexlab{b}})Tang, Zhang, Chen, Wang, and Furukawa]{tang2023mvdiffusion}
Shitao Tang, Fuyang Zhang, Jiacheng Chen, Peng Wang, and Yasutaka Furukawa.
\newblock Mvdiffusion: Enabling holistic multi-view image generation with correspondence-aware diffusion.
\newblock \emph{arXiv preprint arXiv:2307.01097}, 2023{\natexlab{b}}.

\bibitem[Teed and Deng(2020)]{teed2020raft}
Zachary Teed and Jia Deng.
\newblock Raft: Recurrent all-pairs field transforms for optical flow.
\newblock In \emph{European conference on computer vision}, pages 402--419. Springer, 2020.

\bibitem[Tulsiani et~al.(2016)Tulsiani, Kar, Carreira, and Malik]{tulsiani2016learning}
Shubham Tulsiani, Abhishek Kar, Joao Carreira, and Jitendra Malik.
\newblock Learning category-specific deformable 3d models for object reconstruction.
\newblock \emph{IEEE transactions on pattern analysis and machine intelligence}, 39\penalty0 (4):\penalty0 719--731, 2016.

\bibitem[Tulsiani et~al.(2017)Tulsiani, Zhou, Efros, and Malik]{tulsiani2017drc}
Shubham Tulsiani, Tinghui Zhou, Alexei~A Efros, and Jitendra Malik.
\newblock Multi-view supervision for single-view reconstruction via differentiable ray consistency.
\newblock In \emph{Proceedings of the IEEE conference on computer vision and pattern recognition}, pages 2626--2634, 2017.

\bibitem[Wang et~al.(2023{\natexlab{a}})Wang, Du, Li, Yeh, and Shakhnarovich]{wang2023score}
Haochen Wang, Xiaodan Du, Jiahao Li, Raymond~A Yeh, and Greg Shakhnarovich.
\newblock Score jacobian chaining: Lifting pretrained 2d diffusion models for 3d generation.
\newblock In \emph{Proceedings of the IEEE/CVF Conference on Computer Vision and Pattern Recognition}, pages 12619--12629, 2023{\natexlab{a}}.

\bibitem[Wang et~al.(2018)Wang, Zhang, Li, Fu, Liu, and Jiang]{wang2018pixel2mesh}
Nanyang Wang, Yinda Zhang, Zhuwen Li, Yanwei Fu, Wei Liu, and Yu-Gang Jiang.
\newblock Pixel2mesh: Generating 3d mesh models from single rgb images.
\newblock In \emph{Proceedings of the European conference on computer vision (ECCV)}, pages 52--67, 2018.

\bibitem[Wang et~al.(2021)Wang, Liu, Liu, Theobalt, Komura, and Wang]{wang2021neus}
Peng Wang, Lingjie Liu, Yuan Liu, Christian Theobalt, Taku Komura, and Wenping Wang.
\newblock Neus: Learning neural implicit surfaces by volume rendering for multi-view reconstruction.
\newblock \emph{arXiv preprint arXiv:2106.10689}, 2021.

\bibitem[Wang et~al.(2004)Wang, Bovik, Sheikh, and Simoncelli]{wang2004image}
Zhou Wang, Alan~C Bovik, Hamid~R Sheikh, and Eero~P Simoncelli.
\newblock Image quality assessment: from error visibility to structural similarity.
\newblock \emph{IEEE transactions on image processing}, 13\penalty0 (4):\penalty0 600--612, 2004.

\bibitem[Wang et~al.(2023{\natexlab{b}})Wang, Lu, Wang, Bao, Li, Su, and Zhu]{wang2023prolificdreamer}
Zhengyi Wang, Cheng Lu, Yikai Wang, Fan Bao, Chongxuan Li, Hang Su, and Jun Zhu.
\newblock Prolificdreamer: High-fidelity and diverse text-to-3d generation with variational score distillation.
\newblock \emph{arXiv preprint arXiv:2305.16213}, 2023{\natexlab{b}}.

\bibitem[Watson et~al.(2022)Watson, Chan, Martin-Brualla, Ho, Tagliasacchi, and Norouzi]{watson2022novel}
Daniel Watson, William Chan, Ricardo Martin-Brualla, Jonathan Ho, Andrea Tagliasacchi, and Mohammad Norouzi.
\newblock Novel view synthesis with diffusion models.
\newblock \emph{arXiv preprint arXiv:2210.04628}, 2022.

\bibitem[Weng et~al.(2023{\natexlab{a}})Weng, Yang, Wang, Li, Zhang, Chen, and Zhang]{weng2023consistent123}
Haohan Weng, Tianyu Yang, Jianan Wang, Yu Li, Tong Zhang, CL Chen, and Lei Zhang.
\newblock Consistent123: Improve consistency for one image to 3d object synthesis.
\newblock \emph{arXiv preprint arXiv:2310.08092}, 2023{\natexlab{a}}.

\bibitem[Weng et~al.(2023{\natexlab{b}})Weng, Wang, and Yeung]{weng2023zeroavatar}
Zhenzhen Weng, Zeyu Wang, and Serena Yeung.
\newblock Zeroavatar: Zero-shot 3d avatar generation from a single image.
\newblock \emph{arXiv preprint arXiv:2305.16411}, 2023{\natexlab{b}}.

\bibitem[Wibirama et~al.(2020)Wibirama, Santosa, Widyarani, Brilianto, and Hafidh]{wibirama2020physical}
Sunu Wibirama, Paulus~Insap Santosa, Putu Widyarani, Nanda Brilianto, and Wina Hafidh.
\newblock Physical discomfort and eye movements during arbitrary and optical flow-like motions in stereo 3d contents.
\newblock \emph{Virtual Reality}, 24\penalty0 (1):\penalty0 39--51, 2020.

\bibitem[Wu et~al.(2018)Wu, Zhang, Zhang, Zhang, Freeman, and Tenenbaum]{wu2018learning}
Jiajun Wu, Chengkai Zhang, Xiuming Zhang, Zhoutong Zhang, William~T Freeman, and Joshua~B Tenenbaum.
\newblock Learning shape priors for single-view 3d completion and reconstruction.
\newblock In \emph{Proceedings of the European Conference on Computer Vision (ECCV)}, pages 646--662, 2018.

\bibitem[Wu et~al.(2020{\natexlab{a}})Wu, Zhuang, Xu, Zhang, and Chen]{wu2020pq}
Rundi Wu, Yixin Zhuang, Kai Xu, Hao Zhang, and Baoquan Chen.
\newblock Pq-net: A generative part seq2seq network for 3d shapes.
\newblock In \emph{Proceedings of the IEEE/CVF Conference on Computer Vision and Pattern Recognition}, pages 829--838, 2020{\natexlab{a}}.

\bibitem[Wu et~al.(2020{\natexlab{b}})Wu, Rupprecht, and Vedaldi]{wu2020unsupervised}
Shangzhe Wu, Christian Rupprecht, and Andrea Vedaldi.
\newblock Unsupervised learning of probably symmetric deformable 3d objects from images in the wild.
\newblock In \emph{Proceedings of the IEEE/CVF conference on computer vision and pattern recognition}, pages 1--10, 2020{\natexlab{b}}.

\bibitem[Xiao et~al.(2021)Xiao, Kreis, and Vahdat]{xiao2021tackling}
Zhisheng Xiao, Karsten Kreis, and Arash Vahdat.
\newblock Tackling the generative learning trilemma with denoising diffusion gans.
\newblock \emph{arXiv preprint arXiv:2112.07804}, 2021.

\bibitem[Xu et~al.(2023)Xu, Jiang, Wang, Fan, Wang, and Wang]{xu2023neurallift}
Dejia Xu, Yifan Jiang, Peihao Wang, Zhiwen Fan, Yi Wang, and Zhangyang Wang.
\newblock Neurallift-360: Lifting an in-the-wild 2d photo to a 3d object with 360deg views.
\newblock In \emph{Proceedings of the IEEE/CVF Conference on Computer Vision and Pattern Recognition}, pages 4479--4489, 2023.

\bibitem[Yang et~al.(2023)Yang, Cheng, Duan, Ji, and Li]{yang2023consistnet}
Jiayu Yang, Ziang Cheng, Yunfei Duan, Pan Ji, and Hongdong Li.
\newblock Consistnet: Enforcing 3d consistency for multi-view images diffusion.
\newblock \emph{arXiv preprint arXiv:2310.10343}, 2023.

\bibitem[Ye et~al.(2023)Ye, Wang, Li, Shi, and Wang]{ye2023consistent}
Jianglong Ye, Peng Wang, Kejie Li, Yichun Shi, and Heng Wang.
\newblock Consistent-1-to-3: Consistent image to 3d view synthesis via geometry-aware diffusion models.
\newblock \emph{arXiv preprint arXiv:2310.03020}, 2023.

\bibitem[Yu et~al.(2019)Yu, Lin, Yang, Shen, Lu, and Huang]{yu2019free}
Jiahui Yu, Zhe Lin, Jimei Yang, Xiaohui Shen, Xin Lu, and Thomas~S Huang.
\newblock Free-form image inpainting with gated convolution.
\newblock In \emph{Proceedings of the IEEE/CVF international conference on computer vision}, pages 4471--4480, 2019.

\bibitem[Zeng et~al.(2022)Zeng, Vahdat, Williams, Gojcic, Litany, Fidler, and Kreis]{zeng2022lion}
Xiaohui Zeng, Arash Vahdat, Francis Williams, Zan Gojcic, Or Litany, Sanja Fidler, and Karsten Kreis.
\newblock Lion: Latent point diffusion models for 3d shape generation.
\newblock \emph{arXiv preprint arXiv:2210.06978}, 2022.

\bibitem[Zhang et~al.(2018{\natexlab{a}})Zhang, Isola, Efros, Shechtman, and Wang]{zhang2018unreasonable}
Richard Zhang, Phillip Isola, Alexei~A Efros, Eli Shechtman, and Oliver Wang.
\newblock The unreasonable effectiveness of deep features as a perceptual metric.
\newblock In \emph{Proceedings of the IEEE conference on computer vision and pattern recognition}, pages 586--595, 2018{\natexlab{a}}.

\bibitem[Zhang et~al.(2018{\natexlab{b}})Zhang, Zhang, Zhang, Tenenbaum, Freeman, and Wu]{zhang2018learning}
Xiuming Zhang, Zhoutong Zhang, Chengkai Zhang, Josh Tenenbaum, Bill Freeman, and Jiajun Wu.
\newblock Learning to reconstruct shapes from unseen classes.
\newblock \emph{Advances in neural information processing systems}, 31, 2018{\natexlab{b}}.

\bibitem[Zhou and Tulsiani(2023)]{zhou2023sparsefusion}
Zhizhuo Zhou and Shubham Tulsiani.
\newblock Sparsefusion: Distilling view-conditioned diffusion for 3d reconstruction.
\newblock In \emph{Proceedings of the IEEE/CVF Conference on Computer Vision and Pattern Recognition}, pages 12588--12597, 2023.

\bibitem[Zou et~al.(2023)Zou, Cheng, Cao, Huang, Shan, and Zhang]{zou2023sparse3d}
Zi-Xin Zou, Weihao Cheng, Yan-Pei Cao, Shi-Sheng Huang, Ying Shan, and Song-Hai Zhang.
\newblock Sparse3d: Distilling multiview-consistent diffusion for object reconstruction from sparse views.
\newblock \emph{arXiv preprint arXiv:2308.14078}, 2023.

\end{thebibliography}
